\documentclass{article} %
\usepackage{iclr2027_conference,times}

\usepackage{amsmath,amsfonts,bm}

\def\eqref#1{equation~\ref{#1}}

\def\1{\bm{1}}

\DeclareMathAlphabet{\mathsfit}{\encodingdefault}{\sfdefault}{m}{sl}
\SetMathAlphabet{\mathsfit}{bold}{\encodingdefault}{\sfdefault}{bx}{n}

\usepackage[colorlinks=true,linkcolor=black,citecolor=black,urlcolor=black,filecolor=black]{hyperref}
\usepackage{url}
\usepackage{xurl} %

\usepackage{amsmath}
\usepackage{amssymb}
\usepackage{amsthm}
\usepackage{graphicx}
\usepackage{xcolor}
\usepackage{caption}
\usepackage{float}
\usepackage{placeins}
\definecolor{atlassr}{HTML}{2869A0}
\definecolor{atlastitle}{HTML}{25364B}

\usepackage{algorithm}
\usepackage{algpseudocode}
\usepackage{booktabs}
\usepackage{array}      %
\usepackage{tabularx}   %
\usepackage{longtable}  %

\newcolumntype{Y}{>{\centering\arraybackslash}X}
\newcolumntype{L}{>{\raggedright\arraybackslash}X}

\newcommand{\SADLTableStyle}{%
    \small
    \setlength{\tabcolsep}{4pt}%
    \renewcommand{\arraystretch}{1.12}%
}

\newtheorem{proposition}{Proposition}

\title{\centering SADL: On the Demand-Supply Geometry\\of Representation Collapse}

\author{%
\begin{minipage}{0.97\textwidth}
\centering\normalfont
Junhan Sun$^{1}$ \quad Yigui Feng$^{2}$ \quad Mingzhe Liu$^{2}$ \quad Haoyan Xu$^{2}$ \quad Lei Ye$^{2}$\\[2pt]
Ziqi Han$^{2}$ \quad Hanru Bai$^{3}$ \quad Yanzhao Zhang$^{3}$ \quad Hao Zhao$^{2,4,\dagger}$ \quad Guofeng Zhang$^{1,5,\dagger}$\\[5pt]
{\small $^{1}$State Key Laboratory of CAD\&CG, Zhejiang University\\
$^{2}$RoboParty Lab \qquad $^{3}$Alibaba Group\\
$^{4}$AIR, Tsinghua University \qquad $^{5}$InSpatio \qquad $^{\dagger}$Corresponding authors}
\end{minipage}%
}

\iclrfinalcopy %
\hypersetup{
  pdftitle={SADL: On the Demand-Supply Geometry of Representation Collapse},
  pdfauthor={Junhan Sun, Yigui Feng, Mingzhe Liu, Haoyan Xu, Lei Ye, Ziqi Han, Hanru Bai, Yanzhao Zhang, Hao Zhao, Guofeng Zhang}
}

\begin{document}
\maketitle

\pagestyle{plain}\thispagestyle{plain}

\begin{abstract}\relax \relax \relax In \relax Joint-Embedding Predictive Architectures (JEPAs)\relax \relax \relax , a well-constrained projector output does not by itself prevent representation collapse. We find that, from initialization, the encoder representation is dominated by a component shared by different inputs. A batch-normalized projector removes this component, so regularization after the projector cannot correct it, and the differences between representations shrink until they collapse. We propose \relax \textbf{SADL} (Supply And Demand Loss), \relax \relax \relax which keeps the original training objective and constrains the encoder directly when needed\relax . \textbf{SADL} compares \relax \relax \relax the representation differences that the projector output demands with the differences the encoder can retain and pass on to the projector, and adds a constraint on the encoder only when supply falls short of demand\relax . It works with different \relax \relax \relax Demand Regularizers and introduces no \relax auxiliary networks, \relax \relax \relax additional trainable parameters, or architectural changes\relax . In single-task and jointly trained four-task LeWorldModel (LeWM) and INTACT experiments, \relax \textbf{SADL} lowers \relax the collapse rate \relax \relax \relax over ten random seeds \relax from 50\% to 0\%\relax \relax \relax . After one \relax epoch of training\relax \relax , encoder effective rank \relax \relax \relax is \relax 3.4--6.4$\times$ \relax that of the corresponding baseline without SADL, \relax and latent prediction error \relax \relax \relax falls to 1/20--1/303 of it. Without additional tuning, \textbf{SADL} stays robust as its loss weight spans orders of magnitude, and with several Demand Regularizers INTACT exceeds \relax 95\% mean search-free control success. \relax \relax \relax On \relax four image datasets, \relax \relax \relax \textbf{SADL} also consistently improves \relax accuracy and stability \relax \relax \relax across random seeds. These results show that matching the representation differences a regularizer demands with the differences the encoder can supply leads to \relax faster and more stable representation learning \relax \relax \relax in \relax control and image classification.\end{abstract}

\section{Introduction}
\label{sec:introduction}

Joint-Embedding Predictive Architectures (JEPAs) learn from images and video by predicting latent targets instead of reconstructing every pixel \citep{assran2023ijepa,bardes2024vjepa}. \relax \relax \relax An encoder $E_\theta$ maps inputs to a representation $H$, which a projector $P_\phi$ maps to an endpoint \relax $Z$\relax ; existing Demand Regularizers typically constrain the geometry at $Z$ \relax \citep{zbontar2021barlow,bardes2022vicreg,balestriero2025lejepa}. Yet an apparently regular \relax \relax \relax $Z$ \relax does not, on its own, certify that \relax \relax $H$ retains sample variation, or that such variation survives the projection from $H$ to $Z$. \relax \relax \relax Predictability alone does not ensure a useful representation: suppressing distinctions between samples can also make targets easier to predict, and in a \relax world model, losing these distinctions can turn different states or action consequences into nearly indistinguishable control inputs.

\relax \relax \relax We find that this gap is present from initialization. In LeWM and INTACT, the encoder representation \relax $H$ \relax \relax \relax is dominated by a component shared by all inputs: sample variation carries about $0.01\%$ of its energy at initialization and only $3$--$5\%$ after one epoch of training with a Demand Regularizer (Section~\ref{sec:where}). Two mechanisms keep it there. First, the batch normalization (BN) in the projector subtracts the batch mean, so the shared direction lies in the null space of the projector Jacobian, and the gradients of the Demand Regularizer and the task loss both vanish along it. Second, \relax the \relax \relax \relax remaining variation grows mainly through the task gradient and concentrates in a few directions, leaving an \relax effective rank of \relax about $15$ in \relax $H$\relax \relax \relax ; the endpoint gradient reaches these directions but does little to expand them. The projector can remix and amplify existing variation to meet the Demand Regularizer's requirements, but cannot recover what $H$ has lost, so $Z$ inherits the degradation in milder form (effective rank about $20$)\relax .

\relax \relax \relax We find that the problem is not which Demand Regularizer is universally better, but \relax \emph{whether the \relax \relax demand \relax capacity of the endpoint \relax matches the \relax \relax supply \relax capacity of the encoder representation\relax }.
The answer changes over training and across \relax \relax \relax Demand Regularizers.
Supply \relax has two distinct limitations: the encoder may fail to excite sample-dependent directions, or the projector may fail to transmit them.
We therefore introduce \textbf{SADL} (\textbf{S}upply \textbf{A}nd \textbf{D}emand \textbf{L}oss), \relax \relax \relax which adds a Supply Regularizer on $H$ to the Demand Regularizer on $Z$\relax .
\textbf{SADL} reads \relax \relax \relax the demand capacity $q$ \relax from both the \relax \relax \relax Demand Regularizer \relax and the structure already attained at $Z$, measures \relax \relax \relax the supply capacity $k$ as \relax the weaker of \relax \relax \relax the sample variation (excitation) in $H$ and its transmission to $Z$\relax , and corrects $H$ only while \relax \relax \relax $k<q$.
On the shared direction, $\mathcal S$ supplies the nonzero gradient that the other losses lack; on the remaining directions, it lets the weaker ones grow faster. Within a few hundred steps, sample variation exceeds $80\%$ of the energy of $H$ and settles near $99\%$, and the effective ranks of $H$ and \relax $Z$ \relax \relax \relax both rise to about $53$\relax . The selected \relax \relax \relax Demand Regularizer \relax remains unchanged; \textbf{SADL} adds no auxiliary networks, learnable gates, or separate regularization schedule.

\begin{figure}[htbp]
\centering
\relax %
\relax \relax \includegraphics[width=\linewidth,height=0.29\textheight,keepaspectratio]{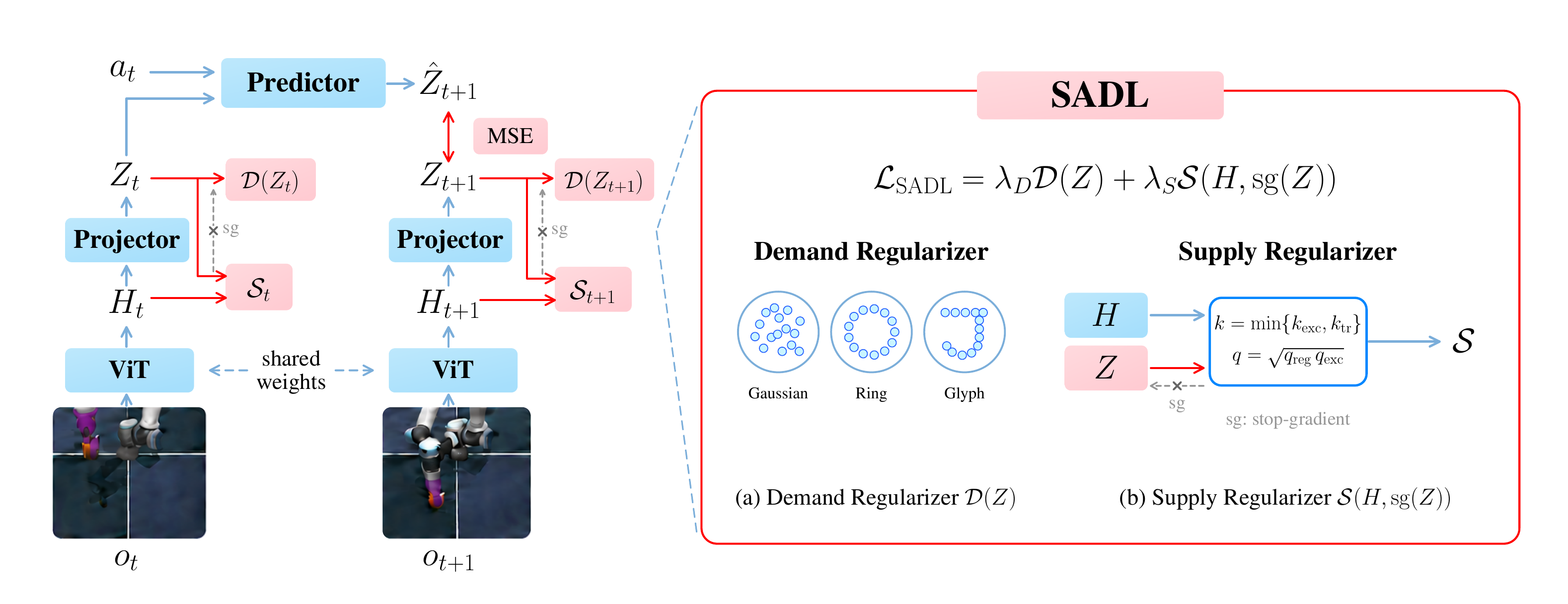}
\relax \caption{Overview of \textbf{SADL}. The \relax \relax \relax Demand Regularizer \relax $\mathcal D(Z)$ and \relax \relax \relax $q_{\rm exc}=F_{\rm exc}(\operatorname{sg}(Z))$ \relax jointly determine \relax the \relax demand \relax capacity \relax $q$; \relax \relax excitation and \relax \relax transmission determine \relax \relax \relax the supply \relax capacity $k$. Their mismatch is penalized by the one-sided \relax \relax \relax Supply Regularizer \relax $\mathcal S$ alongside the task \relax \relax \relax loss\relax . The figure displays the scalar dependence on $q$ and $k$; Section~\relax \relax \relax \ref{sec:ahead_sadl} \relax gives its gradient-routing arguments.}
\label{fig:sadl_overview}
\end{figure}

We evaluate \textbf{SADL} on LeWorldModel (LeWM) and INTACT with multiple \relax \relax \relax Demand Regularizers \relax in single-task and jointly trained four-task control, and examine its generalization to image classification. Across more than ten tested \relax \relax \relax Demand Regularizers\relax , one epoch of joint four-task INTACT training yields at least $92\%$ mean search-free success under the reported evaluation protocols (Appendix Figure~\ref{fig:geometry_gallery}). By epoch three, the INTACT curves reach approximately $96\%$ mean \relax \relax \relax official \relax Direct success (\relax Appendix \relax Figure~\ref{fig:geometry_sr}). Across image self-supervised learning datasets, \textbf{SADL} also serves as a near-zero-overhead plug-in that accelerates convergence while yielding more stable and higher-quality representations.

Our contributions are:
\begin{itemize}
    \item We \relax \relax \relax find that the encoder representation of JEPA-style world models is dominated by a shared component from initialization, that a BN projector leaves this component unconstrained by the downstream task and the Demand Regularizer, and that the resulting degradation of $H$ carries over to $Z$. We cast it as \relax a mismatch between \relax \relax \relax the demand capacity of $Z$ and the supply capacity of $H$\relax .
    \item We introduce \textbf{SADL}, a low-overhead, adaptive, one-sided \relax \relax \relax Supply Regularizer \relax that attaches to any \relax \relax \relax Demand Regularizer \relax satisfying the stated conditions and targets an observed supply deficit rather than imposing a fixed upstream distribution\relax , and remains stable over orders of magnitude of its weight without extra tuning\relax .
    \item Across more than ten \relax \relax \relax Demand Regularizers, control \relax and \relax image tasks, \textbf{SADL} prevents collapse, speeds up convergence, and gives \relax more stable representations\relax ; ablations of the attention backend, the supply terms, the gate, and the demand show that each component is needed\relax .
\end{itemize}

\section{Related Work}
\label{sec:related_work}

\paragraph{Predictive representations and control world models.}
Latent-dynamics models learn compact states for planning or policy improvement \citep{hafner2019planet,hafner2023dreamerv3,hansen2024tdmpc2}; iCEM and self-predictive representations refine planning and predictive training \citep{pinneri2020icem,schwarzer2021spr}. I-JEPA and V-JEPA predict in learned image/video feature spaces \citep{assran2023ijepa,bardes2024vjepa}, with later video models extending toward planning and dense features \citep{assran2025vjepa2,murlabadia2026vjepa21}. Other JEPA-style world models explore compact dynamics, object-level latent masking, and value-guided planning \citep{zhou2024dinowm,nam2026causaljepa,destrade2026valuejepa}. LeWM and INTACT provide the planning and search-free control interfaces studied here \citep{maes2026leworldmodel,sun2026intact}; PhyLatent distinguishes statistical non-collapse from dynamics-relevant information \citep{zeng2026phylatent}. \textbf{SADL} focuses on \relax \relax \relax whether the supply capacity of the encoder representation supports a chosen Demand Regularizer\relax .

\paragraph{Self-supervised objectives and collapse.}
Contrastive image learning uses pairwise or assignment relations \citep{chen2020simclr,he2020moco,caron2020swav}, whereas BYOL, SimSiam, and DINO learn without explicit negative pairs through architectural or optimization asymmetries \citep{grill2020byol,chen2021simsiam,caron2021dino}. Masked prediction and self-distillation provide other routes \citep{he2022mae,zhou2022ibot,assran2022msn,baevski2022data2vec,oquab2023dinov2}. These methods differ in their prediction targets and collapse-avoidance mechanisms; \textbf{SADL} instead studies \relax \relax \relax whether the supply capacity of $H$ meets the demand capacity of $Z$\relax .

\paragraph{Endpoint geometry and statistical regularization.}
Alignment and uniformity, whitening, Barlow Twins, VICReg, and VICRegL constrain embedding geometry through distributional or moment statistics \citep{wang2020alignment,ermolov2021wmse,zbontar2021barlow,bardes2022vicreg,bardes2022vicregl,balestriero2025lejepa,zhao2026subjepa}. VISReg's shape term uses sliced $W_2^2$ matching, while QQWorld's quantile objective approximates one-dimensional $W_2^2$ \citep{wu2026visreg,yu2026qqworld}. In the balanced setting, Earth Mover's Distance corresponds to $W_1$ for normalized discrete measures with a metric ground cost \citep{peyre2019computational,rubner2000emd}. \relax \relax \relax Maximal coding rate reduction (MCR$^2$) pursues representational diversity through a log-determinant expansion term \citep{yu2020mcr2}; its efficient variational form \citep{baek2022mcr2variational}, the total coding rate of EMP-SSL \citep{tong2023empssl}, and CorInfoMax \citep{ozsoy2022corinfomax} further develop log-determinant objectives for self-supervised learning. \textbf{SADL} uses the normalized form of the MCR$^2$ expansion term as its excitation function $F_{\rm exc}$; the precise relation and the design differences are stated in Section~\ref{sec:ablations}. These objectives \relax constrain endpoint geometry; \textbf{SADL} retains the selected loss \relax \relax \relax as its Demand Regularizer and compares the resulting demand capacity with the excitation and transmission of $H$\relax .

\paragraph{Projection interfaces and intermediate regularization.}
DirectCLR and Guillotine Regularization show that projection and readout choices affect dimensional collapse and downstream utility \citep{jing2022dimensional,bordes2023guillotine}. Under specific projector conditions, endpoint variance--covariance control can induce pre-projection independence; VCReg and Seq-VCR also constrain intermediate representations \citep{mialon2022pairwise,zhu2023vcreg,arefin2024seqvcr}. Analyses of non-contrastive Siamese learning examine architectural and optimization routes around collapse \citep{zhang2022simsiamcollapse,li2022siamesecollapse}, while the contrastive--non-contrastive duality concerns relations between loss families under stated assumptions \citep{garrido2023duality}. Unlike fixed intermediate constraints, \textbf{SADL} conditions its correction on \relax \relax \relax the demand capacity read \relax from the selected \relax \relax \relax Demand Regularizer\relax .

\section{Method}
\label{sec:theory}

\begin{figure}[t]
\centering
\relax %
\relax \relax \includegraphics[width=\linewidth,height=0.29\textheight,keepaspectratio]{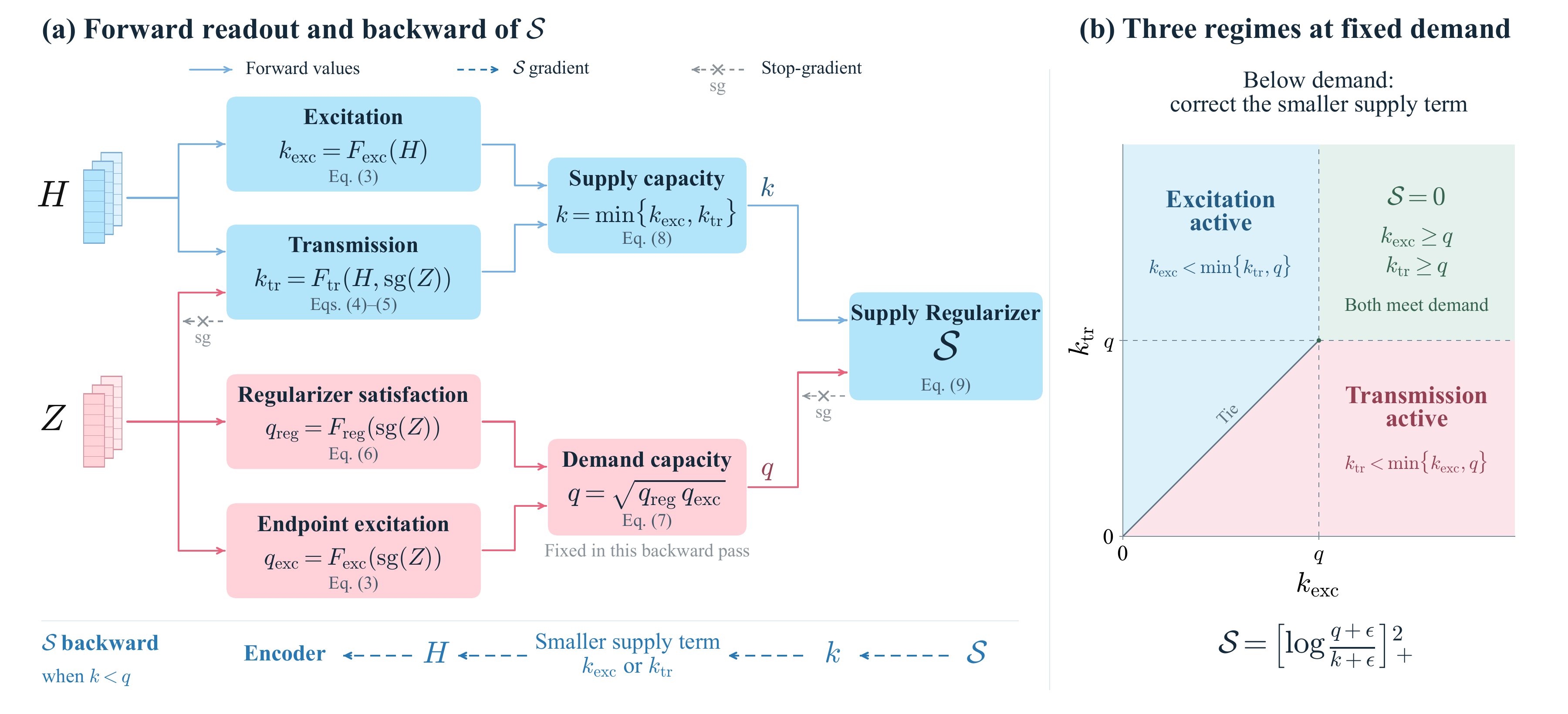}
\relax \caption{Demand--supply mechanism. \relax \relax \relax $q_{\rm reg}$ and $q_{\rm exc}$, computed from the detached \relax endpoint\relax \relax \relax , \relax determine \relax the \relax demand \relax capacity \relax $q$, while the weaker of \relax \relax excitation \relax $k_{\rm exc}$ \relax and \relax \relax transmission \relax $k_{\rm tr}$ \relax determines \relax the \relax supply \relax capacity \relax $k$. The \relax \relax \relax Supply Regularizer \relax acts only for $k<q$ and updates \relax $H$ so that even \relax the \relax \relax \relax smaller supply term meets the demand\relax ; ordinary task and \relax \relax \relax Demand Regularizer \relax gradients remain in place.}
\label{fig:sadl_mechanism}
\end{figure}

For $N$ aligned samples, an encoder \relax \relax \relax $E_\theta$ and a projector $P_\phi$
produce $H=E_\theta(X)\in\mathbb{R}^{N\times d_H}$
and $Z=P_\phi(H)\in\mathbb{R}^{N\times d_Z}$; we call $Z$ the endpoint\relax .
We retain the task loss $\mathcal{L}_{\mathrm{task}}$ and \relax \relax \relax a
Demand Regularizer \relax $\mathcal D(Z)\geq0$ \relax on the endpoint\relax , where lower values indicate
greater \relax \relax satisfaction and $0<\mathcal D(0)<\infty$.
\relax \relax \textbf{SADL} reads a detached \relax \relax \relax demand capacity \relax $q$, measures the
\relax \relax \relax supply capacity \relax $k$ \relax of $H$\relax , and corrects $H$ \relax through a Supply Regularizer \relax only when $k<q$.
Figure~\ref{fig:sadl_mechanism} summarizes the demand and supply \relax \relax \relax terms \relax and their gradient routes.
\subsection{\relax \relax \relax Collapse Mechanism\relax }
\label{sec:ahead_projection_path}

\relax \relax \relax Let $J_P(H)$ denote the Jacobian of the \relax batch-level projector \relax \relax \relax $Z=P_\phi(H)$.
Backpropagation gives
$\nabla_H\mathcal D(Z)=J_P(H)^\top\nabla_Z\mathcal D(Z)$\relax .
Thus, for any perturbation $\Delta H$,
\begin{equation}
\left|\left\langle\nabla_H\mathcal D(Z),\Delta H\right\rangle_F\right|
\leq
\|\nabla_Z\mathcal D(Z)\|_F
\left\|J\relax \relax \relax _P\relax (H)[\Delta H]\right\|_F.
\label{eq:ahead_projection_response}
\end{equation}
The corrective signal along \relax \relax \relax a direction of $H$ \relax is limited
by the projector's local response and vanishes in \relax \relax \relax the null space of $J_P$.
A BN projector gives a concrete instance. In training mode,
adding the same vector $c$ to every row of $H$ leaves $Z$ unchanged, so
$\Delta H=\mathbf 1c^\top$ lies in the null space of $J_P$ and the right-hand side
of Equation~\ref{eq:ahead_projection_response} is zero: neither the Demand
Regularizer nor any loss computed from $Z$ acts on the shared component of $H$
(Section~\ref{sec:where}). The share of this component in $H$ is therefore unconstrained,
the differences between samples are squeezed, and the representation degrades.
\relax This motivates a direct constraint on $H$,
coordinated with the structure formed at $Z$.
We characterize \relax \relax \relax the supply capacity of $H$ with two measurement functions:
excitation $F_{\rm exc}$\relax , which measures sample-dependent variation, and
\relax \relax \relax transmission $F_{\rm tr}$\relax , which measures its linear association with $Z$.
\relax \relax \relax Likewise, we express the demand capacity of $Z$ with $F_{\rm exc}$ and $F_{\rm reg}$:
the former measures the variation among endpoint samples, and the latter
how the endpoint is distributed under the Demand Regularizer\relax .

\subsection{\relax \relax \relax Three Measurement Functions\relax }
\relax %
\relax \relax \label{sec:ahead_functions}

\relax 

For \relax any matrix \relax $U\in\mathbb{R}^{N\times d_U}$ \relax of $N$ samples \relax and $\epsilon>0$, \relax \relax \relax let
\relax \begin{equation}
\relax %
\relax \relax \mu_U=\tfrac1N U^\top\mathbf{1},\quad
\bar{U}=U-\mathbf{1}\mu_U^\top,\quad
\Sigma_U=\tfrac1N\bar{U}^\top\bar{U},\quad
e_U=\max\!\left\{\tfrac{\|U\|_F^2}{Nd_U},\epsilon\right\}.
\relax \label{eq:ahead_feature_statistics}
\end{equation}
\relax \relax \relax Here $\mu_U$ is the sample mean, $\bar U$ the centered features, $\Sigma_U$ the sample covariance, and $e_U$ the uncentered mean energy per coordinate, floored at $\epsilon$. These shared statistics underpin the three functions defined below.

\paragraph{Excitation $F_{\rm exc}(U)$.}
The goal is a single score that detects both a dominant shared component and spectral concentration, since either form of degradation limits sample variation; this lets the correction excite \relax $H$ \relax \relax \relax along the directions that are actually missing. Normalizing the log-determinant by the \emph{uncentered} energy $e_U$ (rather than the centered $\operatorname{tr}(\Sigma_U)$) makes the score sensitive to a large batch-mean relative to sample variation. Formally,
\relax \begin{equation}
\relax \relax \relax F\relax _{\mathrm{exc}}(U)
=\relax \relax \relax \frac{\log\det(I+\gamma\Sigma_U/e_U)}{m_U\log(1+\gamma d_U/m_U)}\relax ,\qquad \relax m_U=\min(d_U,N-1),\quad\relax \gamma>0.
\label{eq:ahead_excitation}
\end{equation}
This score lies in $[0,1]$ \relax \relax \relax and equals one exactly when the batch is as isotropic as its rank $m_U$ permits (Appendix~\ref{app:ahead_theory}). It is orthogonally invariant and \relax scale invariant whenever the energy floor \relax %
\relax \relax is inactive. \relax With \relax \relax \relax $f_U=\operatorname{tr}(\Sigma_U)/(d_Ue_U)$ the fraction of energy in sample variation, $F_{\mathrm{exc}}(U)\leq\log(1+\gamma d_Uf_U/m_U)/\log(1+\gamma d_U/m_U)$, with equality when all nonzero eigenvalues of $\Sigma_U$ are equal. A dominant shared component lowers this bound; concentration in few directions keeps $F_{\mathrm{exc}}$ below it. One score therefore covers both failure modes.

\paragraph{Transmission $F_{\rm tr}(H,Z)$.}
Excitation alone does not check whether the sample variation in $H$ actually reaches $Z$. The projector could, in principle, discard it. We measure alignment through regularized canonical correlation \citep{hotelling1936relations}. Let $\Sigma_{HZ}=N^{-1}\bar{H}^\top\bar{Z}$ and define
\relax \begin{equation}
\relax \relax \relax A_H\relax =\relax %
\relax \relax \Sigma_H+\rho\tau_H I\relax ,\relax \relax \relax \quad
A_Z=\Sigma_Z+\rho\tau_Z I\relax ,\relax %
\relax \relax \quad
\tau_U=\max\!\left\{\tfrac{\operatorname{tr}(\Sigma_U)}{d_U}\relax ,\relax \relax \relax \epsilon\right\},\quad
K\relax =\relax %
\relax \relax A_H^{-1/2}\Sigma_{HZ}A_Z^{-1/2},
\label{eq:ahead_whitening}
\relax \end{equation}
Here \relax %
\relax \relax $\tau_U$ is the mean per-coordinate variance of $U$, so the ridge $\rho\tau_U I$ with coefficient $\rho>0$ is relative to the scale of $U$\relax ; \relax %
\relax \relax $A_H$ and $A_Z$ are the regularized covariances, and $K$ is the whitened cross-covariance. We then set
\begin{equation}
F_{\mathrm{tr}}(H,Z)
=\frac{\log\det(I+\gamma K^\top K)}{m\log(1+\gamma)}
=\frac{1}{m}\sum_{i=1}^{m}\frac{\log(1+\gamma s_i^2)}{\log(1+\gamma)},
\qquad m=\min(d_H,d_Z,N-1),
\label{eq:ahead_transmission}
\end{equation}\relax 
where \relax \relax \relax $s_1\geq\cdots\geq s_m$ are the singular values of $K$ (the regularized canonical correlations). Normalizing over all $m$ interface modes distinguishes broad alignment from correlation in only a few directions. The \relax relative ridge scale \relax $\tau_U$ \relax on the $H$ side \relax \relax \relax retains its gradient, which preserves forward and backward invariance of $k_{\rm tr}=F_{\rm tr}(H,\operatorname{sg}(Z))$ \relax to global rescaling of $H$ when\relax \relax{} the floor is inactive. We \relax %
\relax \relax need this invariance so that $k_{\rm tr}$ measures how the variation of $H$ is transmitted rather than its magnitude; otherwise the Supply Regularizer could raise $k_{\rm tr}$ merely by rescaling $H$. This is a \relax second-order association proxy\relax ; it \relax does not capture \relax \relax \relax nonlinear dependence, and both its value and gradient vanish \relax at zero cross-covariance\relax \relax \relax . In practice, transmission acts as a safeguard: it contributes no gradient while transmission keeps up with excitation and takes over when it falls behind (Section~\ref{sec:ablations}).

\paragraph{Regularizer readout $F_{\rm reg}(Z)$.}
To read demand from both the endpoint and the chosen Demand Regularizer, we need a score of how well that regularizer is currently satisfied that is normalized across the scales of different Demand Regularizers. We define
\begin{equation}
F_{\mathrm{reg}}(Z)=\operatorname{clip}\!\left(1-\frac{\mathcal D(Z)}{\mathcal D(0)},0,1\right).
\label{eq:ahead_demand_readouts}
\end{equation}
$F_{\rm reg}$ is zero when the Demand Regularizer is at its worst feasible value and rises toward one as it is satisfied. The normalization by $\mathcal D(0)$ (the loss on a shape-matched zero input) makes the score invariant to common positive scaling of $\mathcal D(Z)$ \relax and \relax \relax \relax $\mathcal D(0)$, but it depends on the prescribed loss zero point.

\relax 

\relax \relax \relax \subsection{Demand and Supply Capacity}
\label{sec:ahead_capacity}

The demand capacity combines two terms from the detached endpoint:
$q_{\mathrm{reg}}=F_{\mathrm{reg}}(\operatorname{sg}(Z))$ and
$q_{\mathrm{exc}}=F_{\mathrm{exc}}(\operatorname{sg}(Z))$.
Here $q_{\mathrm{reg}}$ increases as the Demand Regularizer is satisfied,
while $q_{\mathrm{exc}}$ measures the current sample variation at the endpoint.
The demand capacity is the geometric mean of the two terms,
\begin{equation}
q=\sqrt{q_{\mathrm{reg}}\,q_{\mathrm{exc}}},
\label{eq:ahead_smoothed_demand}
\end{equation}
so that $q$ is high only when both terms are high.
Appendix~\ref{app:ahead_theory} treats the general interpolation
$q_{\mathrm{reg}}^{1-\alpha}q_{\mathrm{exc}}^{\alpha}$ and its $\epsilon$-smoothed implementation;
we use $\alpha=1/2$ throughout.
Thus $q$ follows the endpoint during training without a learned coefficient, so the demand adapts by itself; a constant demand matches it only when tuned for each setting (Section~\ref{sec:ablations}; Table~\ref{tab:reb_constq}).

The supply capacity is the weaker of the two supply terms:
\relax \begin{equation}
k
=\min\{\relax \relax \relax k\relax _{\mathrm{exc}}\relax ,k_{\mathrm{tr}}\},
\qquad k_{\mathrm{exc}}=F_{\mathrm{exc}}\relax (H),
\relax \relax \relax \qquad k\relax _{\mathrm{tr}}\relax =F_{\mathrm{tr}}\relax (H,\operatorname{sg}(Z))\relax \relax .
\label{eq:ahead_supply}
\end{equation}
Hence $k\geq q$ requires both \relax \relax \relax terms \relax to reach $q$\relax \relax \relax , so that even the smaller
supply term meets the demand\relax .

\subsection{\relax \relax \relax SADL\relax }
\relax %
\relax \relax \label{sec:ahead_sadl}

\relax 

We penalize insufficient \relax \relax \relax supply capacity \relax through the one-sided
\relax \relax \relax Supply Regularizer
\relax \begin{equation}
\mathcal S\bigl(H,\operatorname{sg}(Z)\relax %
\relax \bigr)=\left[\log\frac{q+\epsilon}{k+\epsilon}\right]_+^2,
\qquad[x]_+=\max(x,0).
\label{eq:ahead_bridge}
\end{equation}
\relax \relax \relax It \relax covers both excitation and \relax %
\relax \relax transmission and is \relax zero by construction
whenever \relax \relax \relax the supply capacity meets the demand capacity\relax .

The complete objective \relax \relax \relax is
\relax \begin{equation}
\relax %
\relax \relax \mathcal{L}
=\mathcal{L}_{\mathrm{task}}
+\lambda_D\mathcal D(Z)
+\lambda_S\mathcal S\bigl(H,\operatorname{sg}(Z)\bigr).
\relax \label{eq:ahead_objective}
\end{equation}
Here $\lambda_D$ weights the \relax \relax \relax Demand Regularizer and \relax $\lambda_S$ weights the \relax \relax \relax Supply Regularizer;
\textbf{SADL} denotes the pair\relax . Setting $\lambda_S=0$ \relax \relax \relax recovers the $\mathcal D$-only baseline.
The detached demand capacity and endpoint \relax direct the correction
toward \relax \relax \relax $H$\relax ; the task \relax \relax \relax loss and the Demand Regularizer \relax retain their
normal gradients.
\relax \relax \relax Because the Supply Regularizer \relax remains active whenever a nonzero demand \relax \relax \relax capacity exceeds
the supply capacity, \textbf{it can also repair a near-collapsed checkpoint}
(\relax Appendix~\ref{sec:app_rescue}\relax )\relax .

\relax %
\relax \relax The resulting encoder gradient, $-aJ_{E_\theta}^{\top}\nabla_H k$ with \relax $a$ \relax \relax \relax set by the deficit, bypasses \relax the projector Jacobian\relax %
\relax \relax ; when excitation is the smaller supply term it \relax strengthens weak but nonzero spectral directions \relax \relax \relax while preserving total feature energy (Appendix~\ref{app:ahead_gradients}, Proposition~\ref{prop:spectral}).

\relax 

Sequence features are aligned in batch--time order. Centering,
covariance, whitening, and log-determinant calculations use FP32
within each task's local minibatch before shared-parameter
gradients are averaged across tasks. Excitation energy and
covariance-relative whitening scales retain their gradients,
while \relax \relax \relax $\mathcal D(0)$ \relax is cached from a shape-matched zero input.
The architecture and endpoint computation paths remain unchanged.

\section{Experiments}
\label{sec:experiments}

\relax \relax \relax Experiments are organized in six parts: (1) setup, (2) collapse and repair, (3) internal ablations, (4) convergence and \relax representation quality, \relax \relax \relax (5) general adaptation across Demand Regularizers, and (6) generalization to image tasks. Extended \relax benchmarks and per-task results appear in the appendix.

\subsection{Experimental setup}

We evaluate INTACT and LeWorldModel (LeWM) \citep{maes2026leworldmodel,sun2026intact} on PushT \citep{chi2023diffusionpolicy}, Cube, Reacher, and TwoRoom, both separately and under joint four-task training, with three \relax \relax \relax Demand Regularizers: VISReg \citep{wu2026visreg}, SIGReg \citep{balestriero2025lejepa}\relax , \relax \relax and QQWorld \relax \citep{yu2026qqworld}\relax . In the primary comparison, joint models share a ViT-Tiny encoder, projector, and BatchNorm; the \relax \relax independent Flash collapse diagnostic \relax \relax \relax uses the same shared BatchNorm (Appendix~\ref{sec:app_collapse})\relax . Unless noted otherwise, \textbf{SADL} uses $\alpha=0.5$ and $\lambda_S=0.01$, and control models use BF16. INTACT uses search-free Direct control, whereas LeWM uses Pure-CEM. All control scores below are four-task macro success on the \relax \relax \relax official \relax benchmark. The primary performance comparison uses three training seeds per configuration and evaluates 300 episodes per task at each checkpoint; we report the mean and sample standard deviation across training seeds. The independent collapse diagnostic uses ten training seeds, with its protocol specified in the table caption. \relax Throughout, we classify each training seed at the end of training by the fraction $f_H$ of the energy of $H$ carried by sample variation on the fixed probe: \emph{collapsed} if $f_H<0.05$, \emph{degraded} if $0.05\leq f_H<0.5$, \emph{over-regularized} if $f_H\geq0.5$ but official SR falls below $30\%$, and \emph{normal} otherwise. Tables report these counts as N/O/D/C.

\relax 

Table~\ref{tab:joint} \relax \relax holds the encoder and predictor \relax \relax \relax attention backends, the Demand Regularizer, and its \relax coefficient fixed between baseline and +\textbf{SADL}\relax ; the coefficients are those tuned for the $\mathcal D$-only baseline without \textbf{SADL} \relax (SIGReg $0.03$, VISReg $1.5$, QQWorld $3.5$). Table~\ref{tab:single} retains the original single-task protocol; its backend details, additional benchmarks, and multi-epoch results are provided in Appendix~\ref{app:ahead_control}.

\subsection{\relax \relax \relax Collapse \relax and \relax \relax \relax repair\relax }
\relax \label{sec:where}

\relax 

We \relax \relax \relax compare $\mathcal D$-only \relax and \relax \relax \relax +\textbf{SADL} training by the official success rate, the seed outcomes, and fixed-probe statistics of $H$ and $Z$ (Table~\ref{tab:where}): the fraction of energy in sample variation $f=\operatorname{tr}(\Sigma)/(d\,e)$, the effective rank, the mean sample cosine, and the prediction loss. The results answer the two points of the introduction. \emph{Shared component in the null space.} At initialization, sample variation carries about $0.01\%$ of the energy of $H$, and with a Demand Regularizer alone it is still $3$\relax --\relax \relax \relax $5\%$ after one epoch, while $Z$ already carries $30$\relax --\relax \relax \relax $42\%$. Along the shared direction, the projector's Jacobian--vector product is of order $10^{-13}$, and the gradient energy there is about $10^{-10}$ for the Demand Regularizer and $10^{-9}$ for the task loss, but $42$\relax --\relax \relax \relax $65\%$ for $\mathcal S$. \emph{Variation in few directions.} The remaining variation stays concentrated: $\operatorname{erank}(H)$ is $15$--$17$ and the mean sample cosine of $H$ is about $0.96$. Judged at the end of training (epoch 2), every $\mathcal D$-only seed is degraded, and with Flash part of them collapse: all ten without a Demand Regularizer and five of ten with SIGReg.

Every \textbf{SADL} seed is normal, and \textbf{SADL} roughly doubles LeWM success; within a quarter of an epoch sample variation exceeds $95\%$ of the energy of $H$, and after one epoch the effective ranks of $H$ \relax and \relax \relax \relax $Z$ reach about $53$. Part of this correction goes through the bias $\beta$ of the encoder's final LayerNorm, which is added identically to every sample and so moves only the shared component: $\beta$ grows to cancel the shared mean, and the norm of the batch mean of $H$ falls from $13$\relax --\relax \relax \relax $14.7$ to about $1.4$. Because the Supply Regularizer stays active whenever demand exceeds supply, \textbf{SADL} also repairs near-collapsed checkpoints (Appendix~\ref{sec:app_rescue}). The single-task $\mathcal D$-only baseline of Section~\ref{sec:convergence} uses MATH because $\mathcal D$-only training with Flash is collapse-prone; with matched backends and the same number of updates, +\textbf{SADL} reaches $96.3\%$ with both on INTACT at epoch 2, and LeWM rises from $46.0\%$ to $70.4\%$ under MATH and from $31.4\%$ to $69.1\%$ under Flash (Appendix Table~\ref{tab:reb_backend}). The gain of \textbf{SADL} is therefore not a backend effect.

\relax 

\relax %
\relax \relax \begin{table}[H]
\relax \centering
\caption{\relax \relax \relax Collapse and repair in four-task SIGReg training, at epoch 1 unless noted. SR is official four-task \relax success \relax \relax (\relax \relax \%; mean $\pm$ SD over \relax \relax seeds\relax ; LeWM Pure-CEM, INTACT Direct\relax )\relax . Probe columns are fixed-probe medians over tasks \relax and \relax \relax \relax training seeds (LeWM ten seeds, INTACT three seeds)\relax . \relax \relax \relax N\relax /\relax \relax \relax O\relax /\relax \relax \relax D/C counts training seeds that end normal, over-regularized, degraded, or collapsed (Section~\ref{sec:experiments})\relax ; \relax \relax \relax outcomes are judged at the end of training. $f$ \relax is \relax \relax \relax the fraction of energy in sample variation\relax , \relax \relax \relax $\overline{\cos}$ the \relax mean off-diagonal sample cosine, and \relax \relax \relax $L_{\rm pred}$ the latent prediction loss\relax . \relax \relax \relax Ep\relax .\relax \,0.25 is step 1,423 of 5,695 updates. $^\ddagger$With $f<10^{-6}$ the centered covariance is at the numerical floor, so this effective rank reflects numerical noise rather than sample diversity.\relax }
\relax %
\relax \relax \label{tab:where}
\relax \scriptsize
\setlength{\tabcolsep}{1.2pt}
\begin{tabular*}{\linewidth}{@{\extracolsep{\fill}}llcccccccccc@{}}
\toprule
Model & Setting & SR & N/O/D/C & $f_H$ & $f_Z$ & $\operatorname{erank}(H)$ & $\operatorname{erank}(Z)$ & $\overline{\cos}_H$ & $\overline{\cos}_Z$ & $L_{\rm pred}$ \\
\midrule
LeWM & initialization & $21.5\pm0.0$ & -- & $.0001$ & $.0001$ & $33.2$ & $37.1$ & $.999$ & $.999$ & $.078$ \\
 & MATH, $\mathcal D$-only & $32.0\pm1.6$ & $0/0/10/0$ & $.039$ & $.343$ & $17.3$ & $21.4$ & $.961$ & $.662$ & $.744$ \\
 & Flash, no $\mathcal D$ & $24.1\pm1.2$ & $0/0/0/10$ & $.0000$ & $.0000$ & $69.4^\ddagger$ & $65.3^\ddagger$ & $1.000$ & $1.000$ & $6.377$ \\
 & Flash, $\mathcal D$-only & $31.9\pm5.7$ & $0/0/5/5$ & $.033$ & $.355$ & $14.5$ & $18.5$ & $.967$ & $.660$ & $.995$ \\
 & Flash, +\textbf{SADL} & $\boldsymbol{65.0\pm3.5}$ & $\mathbf{10/0/0/0}$ & $\mathbf{.986}$ & $\mathbf{.990}$ & $\mathbf{52.7}$ & $\mathbf{53.3}$ & $\mathbf{.012}$ & $\mathbf{.008}$ & $\mathbf{.017}$ \\
\midrule
INTACT & initialization & $19.7\pm2.6$ & -- & $.0001$ & $.0001$ & $21.5$ & $23.0$ & $.999$ & $.999$ & $.078$ \\
 & MATH, $\mathcal D$-only & $40.5\pm8.5$ & $0/0/3/0$ & $.033$ & $.304$ & $17.0$ & $19.7$ & $.967$ & $.703$ & $1.172$ \\
 & Flash, $\mathcal D$-only & $45.8\pm2.6$ & $0/0/2/1$ & $.046$ & $.415$ & $14.8$ & $19.9$ & $.954$ & $.593$ & $1.007$ \\
 & MATH, +\textbf{SADL}, ep.\,0.25 & $82.6\pm2.1$ & -- & $.956$ & $.982$ & $28.7$ & $29.3$ & $.043$ & $.016$ & $.043$ \\
 & MATH, +\textbf{SADL} & $94.9\pm0.4$ & $\mathbf{3/0/0/0}$ & $\mathbf{.983}$ & $\mathbf{.990}$ & $\mathbf{52.4}$ & $\mathbf{52.9}$ & $\mathbf{.015}$ & $\mathbf{.008}$ & $\mathbf{.019}$ \\
 & Flash, +\textbf{SADL} & $\boldsymbol{95.4\pm0.2}$ & $\mathbf{3/0/0/0}$ & $\mathbf{.983}$ & $\mathbf{.990}$ & $\mathbf{52.5}$ & $\mathbf{52.8}$ & $\mathbf{.016}$ & $\mathbf{.008}$ & $\mathbf{.017}$ \\
\bottomrule
\end{tabular*}
\end{table}

\relax %
\relax \relax \subsection{Internal ablations}
\label{sec:ablations}

We ablate the Supply Regularizer on $H$ and the adaptive demand\relax . Table~\relax \relax \relax \ref{tab:robust} lists one ablation cell per row, grouped by the component it tests. All cells are four-task INTACT with SIGReg; SR is reported at epoch 1, and seed outcomes use the N\relax /\relax \relax \relax O\relax /\relax \relax \relax D/C classification of Section~\ref{sec:experiments}.

\paragraph{Supply Regularizer on $H$.} \emph{Supply term.} With tr-only, two of three seeds collapse. At the default $\lambda_S=0.01$, the correction is about $2\%$ of the task gradient and $k_{\rm tr}$ is never the smaller term, so exc-only \relax and \relax \relax \relax \textbf{SADL} optimize the same objective. For larger $\lambda_S$, excitation catches up and \textbf{SADL} increasingly selects $k_{\rm tr}$ (up to $63\%$ of TwoRoom training steps at $\lambda_S=0.3$); it then holds transmission at the level of excitation, whereas exc-only lets it fall lower (Appendix~\ref{app:rebuttal}; controlled cases in Appendix Figure~\ref{fig:app_e_roles}). Over $\lambda_S\in[0.3,1]$, \textbf{SADL} is higher than exc-only in both its best weight and its worst seed. \emph{Gate and potential.} The one-sided linear form is lower on both models ($95.31\%$ on INTACT; $58.4$ versus $63.0\%$ on LeWM). Without the direct gradient of the Demand Regularizer ($\mathcal S$-only), two of three seeds degrade and success stays below $50\%$ on INTACT and $23\%$ on LeWM. \emph{Regularizers on $H$ without gate or demand.} Under the same tuning budget, regularizing $H$ directly reaches the same best success rate, but only \textbf{SADL} stays stable over two orders of magnitude of its weight: the always-on coding rate, which is the same excitation score without gate and demand, degrades or collapses at small weights and is over-regularized at large ones. The MCR$^2$ expansion term \citep{yu2020mcr2,baek2022mcr2variational} is the closest relative of $F_{\rm exc}$; it computes the rate on uncentered, unit-normalized features, so the shared component and the sample differences enter the same log-determinant and are not separated (a constant representation still has a positive rate, whereas $F_{\rm exc}$ is zero). It therefore does not read how much of the energy of $H$ is sample variation, and five of its seeds are over-regularized\relax .

\begin{table}[H]
\centering
\caption{\relax \relax \relax Internal ablations, \relax four-task \relax \relax \relax INTACT Direct with SIGReg, official SR \relax (\relax \relax \relax \%\relax ) \relax \relax \relax at epoch 1. Every row uses three training seeds per $\lambda$ value; Weights is the number of $\lambda$ values tried\relax . \relax \relax \relax N\relax /\relax \relax \relax O\relax /\relax \relax \relax D/C counts seeds over all $\lambda$ values\relax ; \relax \relax \relax Best SR gives \relax the \relax \relax \relax best $\lambda$ in parentheses; Worst seed is the lowest single-seed SR over all $\lambda$ values, with its $\lambda$ in parentheses\relax . \relax \relax \relax \textbf{SADL} uses $k=\min\{k_{\rm exc},k_{\rm tr}\}$\relax , \relax \relax \relax exc-only $k=k_{\rm exc}$, and tr-only $k=k_{\rm tr}$. At $\lambda_S=.01$, $k_{\rm tr}$ \relax is \relax \relax \relax never the smaller term, so exc-only coincides with \textbf{SADL} \relax and \relax \relax \relax the two \relax are \relax \relax \relax compared for $\lambda_S\geq0.3$, where $k_{\rm tr}$ is selected\relax . \relax Each \textbf{SADL} row comes from the same run batch as the rows it is compared with. Every weight is listed in Appendix Table~\ref{tab:reb_direct}.\relax }
\relax %
\relax \relax \label{tab:robust}
\relax \scriptsize
\setlength{\tabcolsep}{3pt}
\renewcommand{\arraystretch}{0.95}
\begin{tabular*}{\linewidth}{@{\extracolsep{\fill}}llcccc@{}}
\toprule
Ablation cell & Weight range & Weights & N/O/D/C & Best SR & Worst seed \\
\midrule
\multicolumn{6}{@{}l}{\textit{Supply term}} \\
\textbf{SADL} & $.01$ & 1 & $\mathbf{3/0/0/0}$ & $\mathbf{95.58}$ & $\mathbf{95.42}$ \\
tr-only & $.01$ & 1 & $1/0/0/2$ & $47.94$ & $23.33$ \\
\textbf{SADL} & $.3$--$1$ & 2 & $\mathbf{6/0/0/0}$ & $\mathbf{93.50}$ ($1$) & $\mathbf{92.50}$ ($.3$) \\
exc-only & $.3$--$1$ & 2 & $6/0/0/0$ & $93.08$ ($1$) & $91.17$ ($.3$) \\
\midrule
\multicolumn{6}{@{}l}{\textit{Gate and potential}} \\
\textbf{SADL} & $.01$ & 1 & $\mathbf{3/0/0/0}$ & $\mathbf{95.58}$ & $\mathbf{95.42}$ \\
Two-sided $u^2$ (no gate) & $.01$ & 1 & $3/0/0/0$ & $95.64$ & $88.42$ \\
One-sided linear $[u]_+$ & $.01$ & 1 & $3/0/0/0$ & $95.31$ & $94.75$ \\
$\mathcal S$-only (no direct gradient of $\mathcal D$) & $.01$ & 1 & $1/0/2/0$ & $45.69$ & $33.33$ \\
\midrule
\multicolumn{6}{@{}l}{\textit{Regularizers on $H$ without gate or demand}} \\
\textbf{SADL} & $.01$--$1$ & 5 & $\mathbf{15/0/0/0}$ & $\mathbf{95.42}$ ($.01$) & $\mathbf{91.17}$ ($1$) \\
Always-on coding rate & $.03$--$3$ & 5 & $11/2/1/1$ & $95.28$ ($.1$) & $25.00$ ($.03$) \\
MCR$^2$ expansion & $.03$--$10$ & 8 & $18/5/1/0$ & $94.75$ ($.3$) & $13.33$ ($.1$) \\
VCReg & $.01$--$1$ & 5 & $15/0/0/0$ & $94.75$ ($.1$) & $87.42$ ($.01$) \\
\bottomrule
\end{tabular*}
\end{table}

\paragraph{Adaptive demand.} A constant demand matches the adaptive $q$ only when tuned for each setting: the best constant is $0.7$ for SIGReg and $0.9$ for VISReg in control, and $0.9$, $0.68$, and $0.5$ on Imagenette, ImageNet-LT, and CIFAR-10, while the adaptive $q$ stays within seed variation of the best constant in all five (Table~\ref{tab:reb_constq}). \textbf{A constant demand must therefore be re-tuned for each setting, whereas the adaptive $q$ needs no tuning.}

\begin{table}[H]
\centering
\caption{Constant demand $q$ versus adaptive demand. Control: four-task INTACT, official SR (\%). Images: Imagenette online probe at epoch 800 (3 seeds), ImageNet-LT best online accuracy within 100 epochs (3 seeds), CIFAR-10 validation accuracy at epoch 100 (5 seeds).}
\label{tab:reb_constq}
\scriptsize
\setlength{\tabcolsep}{2pt}
\renewcommand{\arraystretch}{0.95}
\begin{tabular*}{\linewidth}{@{\extracolsep{\fill}}lccccccc@{}}
\toprule
 & \multicolumn{2}{c}{SIGReg control} & \multicolumn{2}{c}{VISReg control} & \multicolumn{3}{c}{Image accuracy} \\
\cmidrule(lr){2-3}\cmidrule(lr){4-5}\cmidrule(l){6-8}
Demand & Ep.\,1 & Ep.\,2 & Ep.\,1 & Ep.\,2 & Imagenette & ImageNet-LT & CIFAR-10 \\
\midrule
baseline & -- & -- & -- & -- & $90.19\pm.26$ & $22.51\pm.46$ & $78.38\pm.16$ \\
$q=0.3$ & $94.1\pm.8$ & $95.5\pm.8$ & $94.4\pm.3$ & $96.5\pm.2$ & $90.32\pm.03$ & $22.20\pm.15$ & $78.91\pm.35$ \\
$q=0.5$ & $95.0\pm.2$ & $96.1\pm.2$ & $95.3\pm.3$ & $96.8\pm.2$ & $90.49\pm.23$ & $22.60\pm.19$ & $\boldsymbol{79.06\pm.26}$ \\
$q=0.68$ & $94.9\pm.2$ & $96.3\pm.6$ & $95.4\pm.7$ & $97.1\pm.2$ & $90.61\pm.43$ & $\boldsymbol{23.00\pm.06}$ & $78.77\pm.42$ \\
$q=0.7$ & $95.4\pm.2$ & $96.8\pm.0$ & $95.8\pm.7$ & $97.1\pm.3$ & $90.57\pm.23$ & $22.74\pm.09$ & $78.90\pm.55$ \\
$q=0.9$ & $95.1\pm.5$ & $96.3\pm.2$ & $95.5\pm.4$ & $97.1\pm.2$ & $90.66\pm.46$ & $22.68\pm.08$ & $78.75\pm.38$ \\
adaptive & $\boldsymbol{94.9\pm.8}$ & $\boldsymbol{96.5\pm.2}$ & $\boldsymbol{95.8\pm.6}$ & $\boldsymbol{97.1\pm.6}$ & $\boldsymbol{91.04\pm.22}$ & $\boldsymbol{22.92\pm.07}$ & $\boldsymbol{78.75\pm.47}$ \\
\bottomrule
\end{tabular*}
\end{table}

\relax %
\relax \begin{table}[!t]
\centering
\caption{Epoch 1 official macro success rate (SR, \%; mean $\pm$ SD over three seeds) and matched-stage fixed probes, for single-task training and joint four-task training. $\mathrm{SR}$/$\mathrm{SR}_{\rm SADL}$ are without/with \textbf{SADL}; erank is effective rank, cos is mean off-diagonal sample cosine, and subscripts $H/Z$ indicate the representation. Probe entries show baseline $\to$ +\textbf{SADL} (three-seed means).}
\label{tab:single}\label{tab:joint}
\relax \scriptsize
\setlength{\tabcolsep}{1.1pt}
\renewcommand{\arraystretch}{1.0}
\begin{tabular*}{\linewidth}{@{\extracolsep{\fill}}llccr@{${}\to{}$}lr@{${}\to{}$}lr@{${}\to{}$}lr@{${}\to{}$}lr@{${}\to{}$}l@{}}
\toprule
Model & $\mathcal D$ & $\mathrm{SR}$ & $\mathrm{SR}_{\rm SADL}$ & \multicolumn{2}{c}{$\mathrm{erank}_H$} & \multicolumn{2}{c}{$\mathrm{cos}_H$} & \multicolumn{2}{c}{$\mathrm{erank}_Z$} & \multicolumn{2}{c}{$\mathrm{cos}_Z$} & \multicolumn{2}{c}{$L_{\rm pred}$} \\
\midrule
\multicolumn{14}{@{}l}{\textbf{Single-task}} \\
LeWM & SIGReg & $45.25\pm6.50$ & $\mathbf{72.19\pm0.97}$ & $15.08$ & $\mathbf{51.90}$ & $.963$ & $\mathbf{.013}$ & $22.97$ & $\mathbf{46.32}$ & $.421$ & $\mathbf{.007}$ & $3.247$ & $\mathbf{.021}$ \\
 & VISReg & $55.67\pm5.46$ & $\mathbf{76.31\pm1.23}$ & $13.30$ & $\mathbf{45.60}$ & $.943$ & $\mathbf{.013}$ & $21.36$ & $\mathbf{34.59}$ & $.214$ & $\mathbf{.007}$ & $.586$ & $\mathbf{.011}$ \\
 & QQWorld & $51.19\pm6.98$ & $\mathbf{75.28\pm1.73}$ & $15.33$ & $\mathbf{51.49}$ & $.956$ & $\mathbf{.013}$ & $26.25$ & $\mathbf{47.54}$ & $.287$ & $\mathbf{.007}$ & $.869$ & $\mathbf{.018}$ \\
\midrule
INTACT & SIGReg & $58.64\pm8.14$ & $\mathbf{95.33\pm0.44}$ & $15.28$ & $\mathbf{51.96}$ & $.976$ & $\mathbf{.011}$ & $25.86$ & $\mathbf{47.95}$ & $.363$ & $\mathbf{.007}$ & $3.160$ & $\mathbf{.021}$ \\
 & VISReg & $71.97\pm12.79$ & $\mathbf{95.75\pm0.08}$ & $13.74$ & $\mathbf{46.07}$ & $.963$ & $\mathbf{.014}$ & $23.77$ & $\mathbf{37.39}$ & $.264$ & $\mathbf{.007}$ & $.620$ & $\mathbf{.012}$ \\
 & QQWorld & $61.56\pm8.61$ & $\mathbf{95.47\pm0.21}$ & $15.16$ & $\mathbf{51.26}$ & $.970$ & $\mathbf{.013}$ & $27.35$ & $\mathbf{49.88}$ & $.350$ & $\mathbf{.006}$ & $2.279$ & $\mathbf{.019}$ \\
\midrule
\multicolumn{14}{@{}l}{\textbf{Joint four-task}} \\
LeWM & SIGReg & $37.2\pm8.3$ & $\mathbf{63.0\pm1.3}$ & $15.7$ & $\mathbf{54.0}$ & $.937$ & $\mathbf{.013}$ & $24.4$ & $\mathbf{50.9}$ & $.419$ & $\mathbf{.008}$ & $.651$ & $\mathbf{.018}$ \\
 & VISReg & $47.7\pm5.1$ & $\mathbf{71.6\pm3.0}$ & $13.7$ & $\mathbf{48.0}$ & $.853$ & $\mathbf{.014}$ & $22.7$ & $\mathbf{37.7}$ & $.376$ & $\mathbf{.009}$ & $.267$ & $\mathbf{.014}$ \\
 & QQWorld & $32.9\pm8.3$ & $\mathbf{68.4\pm4.2}$ & $15.8$ & $\mathbf{54.5}$ & $.944$ & $\mathbf{.013}$ & $21.7$ & $\mathbf{52.7}$ & $.619$ & $\mathbf{.008}$ & $1.643$ & $\mathbf{.017}$ \\
\midrule
INTACT & SIGReg & $28.2\pm6.4$ & $\mathbf{95.6\pm.2}$ & $8.3$ & $\mathbf{53.1}$ & $.994$ & $\mathbf{.014}$ & $9.7$ & $\mathbf{53.1}$ & $.744$ & $\mathbf{.008}$ & $5.300$ & $\mathbf{.017}$ \\
 & VISReg & $54.4\pm2.9$ & $\mathbf{95.4\pm.6}$ & $13.9$ & $\mathbf{48.2}$ & $.912$ & $\mathbf{.016}$ & $21.8$ & $\mathbf{39.8}$ & $.492$ & $\mathbf{.010}$ & $.526$ & $\mathbf{.016}$ \\
 & QQWorld & $38.6\pm12.8$ & $\mathbf{95.6\pm.3}$ & $12.9$ & $\mathbf{53.6}$ & $.952$ & $\mathbf{.014}$ & $18.2$ & $\mathbf{54.4}$ & $.661$ & $\mathbf{.008}$ & $2.044$ & $\mathbf{.019}$ \\
\bottomrule
\end{tabular*}
\relax %
\relax \relax \end{table}
\begin{table}[!t]
\centering
\relax \caption{\relax \relax \relax Epoch 1 official \relax joint control \relax with eight Demand Regularizers \relax and LeJEPA-related diagnostics. Entries are three-seed means (Macro: mean $\pm$ SD); LeWM uses Pure-CEM 300$\times$30, INTACT Direct, and $\lambda_S=.01$. $\widehat J_{\rm proj}$ and $b_{1024}$ are projected Fisher and finite-sample kNN-bias proxies; full definitions, uncertainty, and evaluation protocol appear in Appendix~\ref{app:ahead_geometry}. \relax Target: G, Gaussian; NG, non-Gaussian or density-free.\relax }\label{tab:app_f_eight_official}
\scriptsize\setlength{\tabcolsep}{1.6pt}\renewcommand{\arraystretch}{0.88}
\begin{tabular*}{\linewidth}{@{\extracolsep{\fill}}lllcccccccc@{}}
\toprule
Model & Target & $\mathcal D$ & $\lambda_D$ & PushT & Cube & Reacher & TwoRoom & Macro & $\widehat J_{\rm proj}$ & $b_{1024}$ \\
\midrule
LeWM & G & SIGReg & 0.0114 & 56.00 & 67.11 & 82.33 & 80.78 & $71.56\pm1.27$ & 1.289 & .1772 \\
 &  & VISReg & 1.5000 & 47.33 & 68.44 & 81.00 & 89.56 & $71.58\pm4.67$ & 1.184 & .1717 \\
 &  & QQWorld & 1.2729 & 60.89 & 68.00 & 80.33 & 88.22 & $74.36\pm2.90$ & 1.179 & .1699 \\
 &  & Energy-Gaussian & 2.3383 & 49.44 & 61.56 & 79.44 & 90.56 & $70.25\pm2.35$ & 1.161 & .1712 \\
\cmidrule(l){2-11}
 & NG & VIC moments & 1.9130 & 35.56 & 62.00 & 79.67 & 92.89 & $67.53\pm6.09$ & 1.308 & .2228 \\
 &  & Sphere & 1.1318 & 73.11 & 59.56 & 81.89 & 86.78 & $75.33\pm1.36$ & 1.189 & .1660 \\
 &  & Student-$t_5$ & 1.2104 & 30.11 & 69.56 & 70.44 & 81.11 & $62.81\pm3.99$ & 1.591 & .2034 \\
 &  & GMM-Ring & 2.3345 & 51.00 & 63.33 & 79.78 & 88.89 & $70.75\pm1.23$ & 1.232 & .1756 \\
\midrule
INTACT & G & SIGReg & 0.0117 & 85.44 & 100.00 & 96.89 & 100.00 & $95.58\pm0.00$ & 1.287 & .1762 \\
 &  & VISReg & 1.5000 & 83.00 & 99.89 & 97.44 & 100.00 & $95.08\pm0.17$ & 1.250 & .1716 \\
 &  & QQWorld & 1.2746 & 85.89 & 100.00 & 97.44 & 99.89 & $95.81\pm0.13$ & 1.213 & .1745 \\
 &  & Energy-Gaussian & 2.2074 & 86.33 & 100.00 & 95.11 & 99.89 & $95.33\pm0.63$ & 1.259 & .1683 \\
\cmidrule(l){2-11}
 & NG & VIC moments & 1.7339 & 83.78 & 100.00 & 96.33 & 100.00 & $95.03\pm0.99$ & 1.330 & .2287 \\
 &  & Sphere & 1.0725 & 86.33 & 100.00 & 95.89 & 99.56 & $95.44\pm0.82$ & 1.251 & .1590 \\
 &  & Student-$t_5$ & 1.0662 & 76.67 & 99.78 & 93.00 & 99.89 & $92.33\pm1.04$ & 1.660 & .2052 \\
 &  & GMM-Ring & 2.1996 & 84.89 & 100.00 & 95.67 & 99.89 & $95.11\pm0.35$ & 1.220 & .1661 \\
\bottomrule
\end{tabular*}
\end{table}

\subsection{\relax \relax \relax Convergence \relax and \relax \relax \relax representation quality\relax }
\relax \label{sec:convergence}

\relax 

\relax \relax \relax Having established that SADL fixes the shared-component pathology, we quantify the downstream control and representation gains. We test whether the Supply Regularizer makes control-relevant representations available early in training. Each of PushT, Cube, Reacher, and TwoRoom is trained separately, and official success is macro-averaged across the four tasks. After one epoch, across all six pairs, \textbf{SADL} adds $20.6$--$36.7$ points of SR, raises $H/Z$ effective rank by $3.35$--$3.44\times$/$1.57$--$2.02\times$, and lowers prediction loss by $48$--$155\times$ \relax (Table~\relax \relax \relax \ref{tab:single}).

\relax 

\relax \relax \relax Joint \relax four-task \relax \relax \relax training gives the same picture (Table~\ref{tab:joint}): after one epoch, \relax \textbf{SADL} \relax \relax \relax raises official SR by $23.9$--$67.4$ points, effective rank at $H$ by $3.4$--$6.4\times$, and lowers prediction loss by $20$--$303\times$. The single-task baseline uses the MATH backend; matching the backend does not change the result (Section~\ref{sec:where})\relax .

\relax %
\relax \relax 

\relax

\subsection{\relax \relax \relax General adaptation \relax across \relax \relax \relax Demand Regularizers\relax }

We test \textbf{SADL} with eight \relax \relax \relax Demand Regularizers \relax rather than assuming a single Gaussian target. After one epoch, all eight give INTACT over $92\%$ search-free Direct success (seven reach $95.03\%$--$95.81\%$), while LeWM+\textbf{SADL} Pure-CEM reaches $62.81\%$--$75.33\%$ (Table~\ref{tab:app_f_eight_official}). Against matched \relax \relax \relax $\mathcal D$-only \relax LeWM at $23.81\%$--$47.69\%$, \textbf{SADL} gains $23.89$--$46.94$ points (Appendix Table~\ref{tab:app_g_lewm_e1_baseline}). \relax Appendix Figure~\ref{fig:geometry_sr} shows the training curves for four of them. \relax Student-$t_5$, GMM-Ring, Sphere shell, and VIC moments further cover non-Gaussian or density-unspecified constraints. \textbf{SADL therefore adapts across \relax \relax \relax Demand Regularizers \relax without implying that all geometries are equivalent.}

\relax %
\relax \relax 

\relax

\relax %
\relax \relax \subsection{Generalization to image tasks}

\relax 

\begin{figure}[H]
\relax \centering
\relax %
\relax \relax \includegraphics[width=\linewidth]{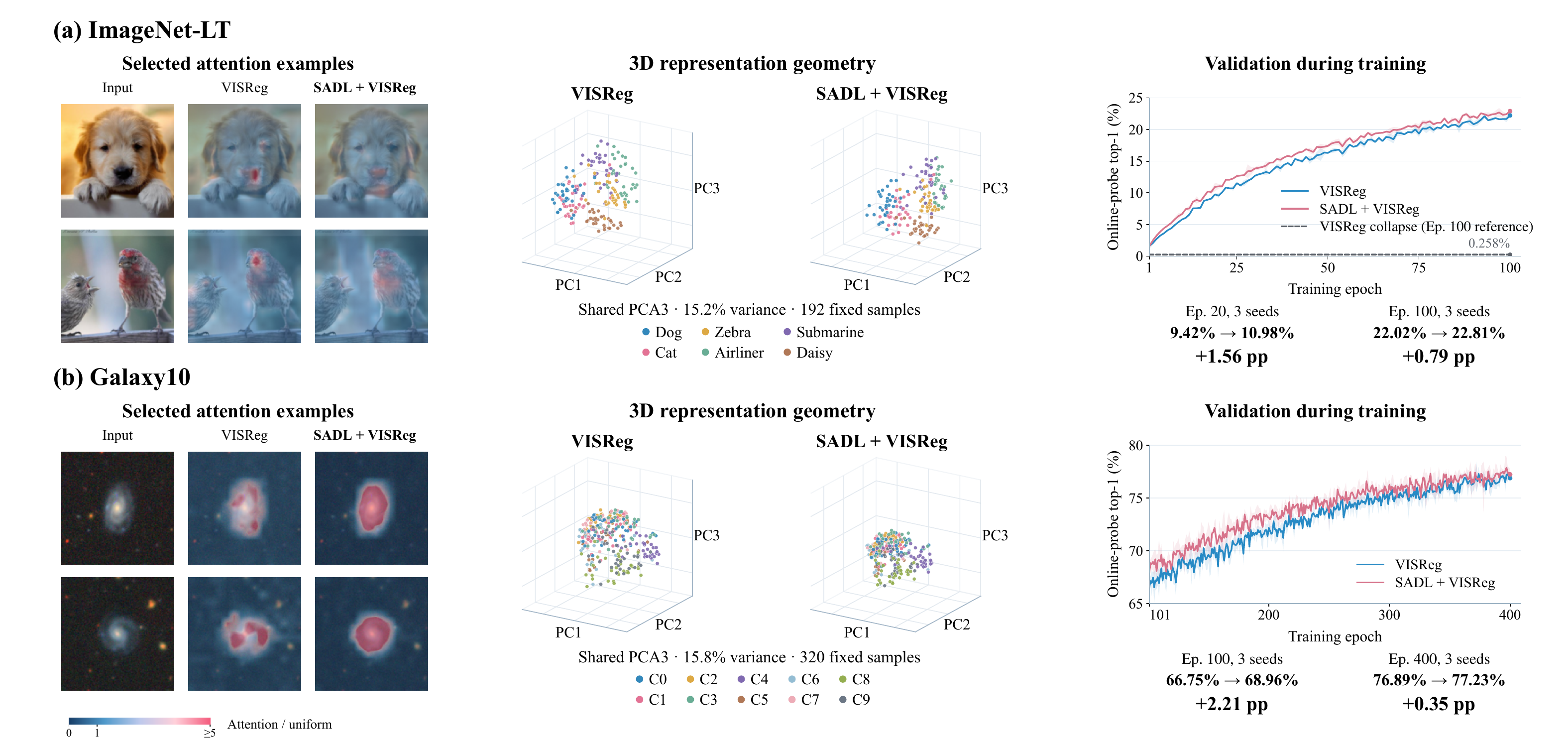}
\relax \caption{ImageNet-LT (top) and Galaxy10 (bottom): representative attention maps, shared three-dimensional feature geometry, and paired online-validation curves. The baseline remains vulnerable to \relax \relax \relax representation \relax degradation; \textbf{SADL} converges faster and more stably.}
\label{fig:image_learning}
\end{figure}

\relax %
\relax Beyond control, \textbf{SADL} improves ImageNet-LT \citep{liu2019imagenetlt}, Galaxy10 \citep{leung2024galaxy10decals}, CIFAR-10, and Imagenette under matched budgets and \relax \relax \relax Demand Regularizers \relax (Figure~\ref{fig:image_learning}; Appendix Figure~\ref{fig:image_training_advantage}, Table~\ref{tab:app_visual_full}). It adds no trainable parameters; implementation runtime and memory are reported in Appendix~\ref{app:ahead_cost}. Early paired gains span 1.56--8.91 points; \relax \relax \relax \textbf{SADL} reaches the final baseline accuracy faster on Imagenette and ImageNet-LT across replication seeds (Appendix~\ref{app:rebuttal}\relax ).

\relax %
\relax \relax 

\relax %
\relax \FloatBarrier
\section{Conclusion}
\label{sec:conclusion}
We introduce \textbf{SADL}, an adaptive one-sided \relax \relax \relax Supply Regularizer that matches the geometric-mean demand capacity to the minimum supply \relax capacity. Across JEPA-style world models \relax \relax \relax with multiple Gaussian and non-Gaussian Demand Regularizers and in image experiments\relax , it accelerates learning and yields more stable, useful representations. Ablations distinguish excitation-driven expansion from correction of transmission \relax \relax \relax deficits; the one-sided gate keeps \textbf{SADL} stable across its weight, and the adaptive demand removes per-setting tuning\relax . Together, these results support demand--supply matching as a low-overhead\relax \relax \relax , general principle for stable representation learning: it coordinates the endpoint $Z$ constrained by a Demand Regularizer with the encoder representation $H$ that must supply it, prevents collapse, and can repair near-collapsed checkpoints\relax .

\paragraph{\relax \relax \relax Limitations\relax .}
The \relax \relax \relax null-space argument of Section~\ref{sec:ahead_projection_path} is specific to BN projectors; projectors with other normalizations have different invariances, which we have not analyzed. $k_{\rm tr}$ measures second-order association only. Our models are ViT-Tiny and ViT-S, and on a small dataset whose baseline is already healthy (CIFAR-10) the gain is within seed noise.

\paragraph{Future work.}
\relax A natural next step is to generalize \relax \relax \relax supply \relax capacity to nonlinear, multimodal, and temporally structured transmission operators\relax \relax \relax , and \relax to develop online diagnostics that anticipate emerging collapse modes and trigger state-adaptive repair\relax \relax .

\clearpage

\raggedbottom\setlength{\bibsep}{6pt plus 0pt minus 0pt}
\bibliography{main,refs_verified}

\bibliographystyle{iclr2027_conference}

\clearpage
\appendix
\relax \makeatletter\setlength{\@fptop}{0pt}\setlength{\@fpsep}{16pt plus 2pt}\setlength{\@fpbot}{0pt plus 1fil}\makeatother
\setlength{\intextsep}{9pt plus 2pt minus 2pt}\setlength{\textfloatsep}{12pt plus 2pt minus 4pt}\setlength{\floatsep}{10pt plus 2pt minus 2pt}
\renewcommand{\floatpagefraction}{0.92}\renewcommand{\textfraction}{0.07}
\setcounter{topnumber}{3}\setcounter{totalnumber}{5}
\relax \raggedbottom

\section*{Appendix}

This appendix gives the theoretical derivations, numerical implementation,
training and evaluation protocols, collapse diagnosis, complete control
results, mechanism ablations, alternative \relax \relax \relax Demand Regularizers\relax , image
experiments, and computational costs. Unless stated otherwise, control
comparisons use three training seeds; each task is evaluated with three
evaluation seeds and 100 episodes per seed. Other sample counts are stated
with their respective results.

\relax %
\relax \begin{table}[H]
\centering
\captionsetup{width=\linewidth,justification=raggedright,singlelinecheck=false}\caption{Notation and experimental labels used throughout the appendix.}
\label{tab:app_notation}
\begingroup\small\setlength{\tabcolsep}{4pt}
\begin{tabularx}{\linewidth}{@{}p{0.29\linewidth}X@{}}
\toprule
Symbol & Meaning \\
\midrule
$X$; $E_\theta,P_\phi$ & Input batch of $N$ aligned samples; encoder and projector. \\
$H=E_\theta(X)$, $Z=P_\phi(H)$ & Encoder representation and endpoint (the projector output), $H\in\mathbb R^{N\times d_H}$, $Z\in\mathbb R^{N\times d_Z}$; $h_i,z_i$ are their sample rows. \\
$J_P$, $J_{E_\theta}$ & Jacobians of the projector and of the encoder. \\
$\hat Z$ & Endpoint predicted by the forward model. \\
$\mu_U,\bar{U},\Sigma_U,e_U$ & Mean, centered features, centered covariance, and uncentered mean energy $\max\{\|U\|_F^2/(Nd_U),\epsilon\}$ of $U$. \\
$f_U$ & Fraction of energy carried by sample variation, $\operatorname{tr}(\Sigma_U)/(d_Ue_U)$. \\
$\mathcal L_{\rm task}$ & Task loss. \\
$\mathcal D(Z)$, $\mathcal D(0)$ & Demand Regularizer (any existing Demand Regularizer, unchanged by SADL) and its value at the all-zero endpoint. \\
$\mathcal S(H,\operatorname{sg}(Z))$ & Supply Regularizer $[\log((q+\epsilon)/(k+\epsilon))]_+^2$. \\
$\lambda_D,\lambda_S$ & Weights of the Demand and Supply Regularizers; $\mathcal L=\mathcal L_{\rm task}+\lambda_D\mathcal D+\lambda_S\mathcal S$. \\
$F_{\rm exc},F_{\rm tr},F_{\rm reg}$ & Excitation, transmission, and regularizer functions, each with values in $[0,1]$. \\
$k_{\rm exc},k_{\rm tr},k$ & Supply terms $F_{\rm exc}(H)$, $F_{\rm tr}(H,\operatorname{sg}(Z))$, and supply capacity $k=\min\{k_{\rm exc},k_{\rm tr}\}$. \\
$q_{\rm reg},q_{\rm exc},q$ & Demand terms $F_{\rm reg}(\operatorname{sg}(Z))$, $F_{\rm exc}(\operatorname{sg}(Z))$, and demand capacity $q=\sqrt{q_{\rm reg}q_{\rm exc}}$. \\
$\alpha$, $q_\alpha^\star$, $\widetilde q_{\alpha,\epsilon}$ & Demand interpolation coefficient (default $1/2$), ideal interpolation $q_{\rm reg}^{1-\alpha}q_{\rm exc}^\alpha$, and its $\epsilon$-smoothed implementation. \\
$\tau_U,\rho,A_U,K,s_i$ & Relative ridge scale and coefficient, regularized covariances, whitened cross-covariance, and its singular values (regularized canonical correlations). \\
$\gamma$, $m_U$, $m$ & SNR coefficient; normalization dimensions $m_U=\min(d_U,N-1)$ and $m=\min(d_H,d_Z,N-1)$. \\
$\nu_i$ & Relative eigenvalues $\lambda_i(\Sigma_U)/e_U$. \\
$a$ & Correction strength in $\nabla_H(\lambda_S\mathcal S)=-a\nabla_Hk$. \\
$\operatorname{sg},\epsilon$ & Stop-gradient and numerical floor. \\
$\operatorname{erank},\cos,\eta_1,\bar\kappa$ & Effective rank, mean off-diagonal sample cosine, leading-eigenvalue share, and mean absolute standardized excess kurtosis. \\
$\widehat J_{\rm proj}$, $b_{1024}$ & Smoothed projected Fisher proxy and finite-sample kNN-bias proxy (Appendix~\ref{app:ahead_geometry}). \\
Ep.\,1/2/5/10, official & Epoch-1/2/5/10 checkpoints and the control-evaluation protocol defined below. \\
$\mathcal D$-only; SADL; exc-only; tr-only & $\lambda_S=0$; the full method; $k$ replaced by $k_{\rm exc}$; $k$ replaced by $k_{\rm tr}$. \\
MATH, Flash & Explicit versus fused memory-efficient encoder-attention backends, not distinct regularization objectives. \\
Direct; Pure-CEM & Search-free INTACT control; LeWM control with 300 candidates and 30 iterations. \\
$T_H$, $R_H$ & Planning horizon and execution horizon (actions executed before replanning), both measured in steps. \\
N/O/D/C & Normal, over-regularized, degraded, and collapsed; outcomes are judged at the end of training (Ep.\,2) using $f_H$ and official SR. \\
\bottomrule
\end{tabularx}
\endgroup
\end{table}
\setlength{\intextsep}{6pt plus 1pt minus 1pt}\setlength{\textfloatsep}{9pt plus 2pt minus 2pt}\setlength{\floatsep}{7pt plus 1pt minus 1pt}
\captionsetup[table]{font=small,skip=5pt}

\relax \clearpage
\section{Theoretical Derivations and Implementation Details}
\label{app:ahead_theory}

This appendix provides the proofs, gradient expressions, and
numerical details for Section~\ref{sec:theory}.

\relax \subsection{Notation and Basic Setup}
\label{app:ahead_notation}

Let
\relax \relax \relax $H\in\mathbb{R}^{N\times d_H}$ and
$Z\in\mathbb{R}^{N\times d_Z}$
\relax contain encoder and projected representations of the same samples
in the same row order.

\relax %

\relax 

For any representation matrix
$U\in\mathbb{R}^{N\times d_U}$, write
\[
\mu_U=\frac{1}{N}U^\top\mathbf{1},
\qquad
\relax \bar{U}=U-\mathbf{1}\mu_U^\top,
\qquad
\relax \Sigma_U=\frac{1}{N}\relax \relax \relax \bar{U}\relax ^\top\relax \relax \relax \bar{U}\relax ,
\]
\[
e_U^{\mathrm{raw}}
=
\frac{\|U\|_F^2}{Nd_U},
\qquad
e_U=\max\{e_U^{\mathrm{raw}},\epsilon\},
\]
where $\epsilon>0$ is a numerical floor.
The total energy decomposes as
\[
d_Ue_U^{\mathrm{raw}}
=
\operatorname{tr}(\Sigma_U)+\|\mu_U\|_2^2.
\]
Consequently, the centered-energy fraction
\[
\relax \relax \relax f\relax _U
=
\frac{\operatorname{tr}(\Sigma_U)}{d_Ue_U}
\]
satisfies \relax \relax \relax $0\leq f_U\leq1$\relax .
When the energy floor is inactive,
\[
\relax \relax \relax f\relax _U
=
1-\frac{\|\mu_U\|_2^2}{d_Ue_U}.
\]

All results below concern finite-batch empirical statistics.
They require neither Gaussian representations nor independence
between samples within the batch.

\subsection{Properties of \relax the \relax Excitation \relax \relax \relax Function\relax }
\label{app:ahead_excitation}

Recall that
\[
\relax \relax \relax F\relax _{\mathrm{exc}}(U)
=
\relax \relax \relax \frac{
\log\det\left(I+\gamma\Sigma_U/e_U\right)
}{
m_U\log(1+\gamma d_U/m_U)
}\relax ,
\qquad \relax m_U=\min(d_U,N-1),\quad \relax \gamma>0.
\]
Let \relax \relax \relax $\nu_i=\lambda_i(\Sigma_U)/e_U$.
Then
\relax \[
\relax \relax \relax F\relax _{\mathrm{exc}}(U)
=
\relax \relax \relax \frac{
\sum_{i=1}^{d_U}\log(1+\gamma \nu_i)
}{
m_U\log(1+\gamma d_U/m_U)
}\relax ,
\qquad
\frac{1}{d_U}\sum_{i=1}^{d_U}\relax \relax \relax \nu\relax _i=\relax \relax \relax f\relax _U.
\]

\paragraph{Range and equality conditions.}
Each logarithmic term is nonnegative because \relax \relax \relax $\nu_i\geq0$.
Since $\operatorname{rank}(\Sigma_U)\leq\min(d_U,N-1)=m_U$, at most
$m_U$ relative eigenvalues are nonzero, and their sum is $d_Uf_U$.
\relax By the strict concavity of \relax \relax \relax $\log(1+\gamma\nu)$ over these terms,
\relax \[
\relax %
\relax \relax \begin{aligned}
\sum_{i=1}^{d_U}\log(1+\gamma \nu_i)
&\leq
m_U\log\left(
1+\frac{\gamma d_Uf_U}{m_U}
\right)\\
&\leq
m_U\log\left(1+\frac{\gamma d_U}{m_U}\right).
\end{aligned}\relax 
\]
Therefore,
\[
0\leq \relax \relax \relax F\relax _{\mathrm{exc}}(U)\leq1.
\]

The lower bound is attained if and only if $\Sigma_U=0$.
Equality at the upper bound requires \relax \relax \relax $f_U=1$ and $m_U$ equal nonzero
relative eigenvalues, equivalently
\relax \[
\mu_U=0,
\qquad
\relax \operatorname{rank}(\relax \Sigma_U\relax )\relax =\relax m_U,
\qquad
\lambda_i(\Sigma_U)=\frac{d_U}{m_U}\relax e_U\relax \relax \relax \ \ (\lambda_i>0)\relax ,
\qquad
e_U^{\mathrm{raw}}=e_U.
\]
\relax \relax \relax The upper bound is thus attainable for every batch size\relax .

\paragraph{Coordinate and scale properties.}
For any orthogonal matrix $Q$, the transformation $U\mapsto UQ$
preserves total energy and orthogonally conjugates the covariance.
Hence
\[
\relax \relax \relax F\relax _{\mathrm{exc}}(UQ)=\relax \relax \relax F\relax _{\mathrm{exc}}(U).
\]

Now consider $U\mapsto cUQ$, where $c\neq0$.
If the energy floor is inactive before and after the transformation,
then
\[
\Sigma_{cUQ}=c^2Q^\top\Sigma_UQ,
\qquad
e_{cUQ}=c^2e_U.
\]
The normalized covariance changes only by orthogonal conjugation,
which gives
\[
\relax \relax \relax F\relax _{\mathrm{exc}}(cUQ)=\relax \relax \relax F\relax _{\mathrm{exc}}(U).
\]

The normalization retains sensitivity to the common feature mean.
At fixed centered covariance, increasing the mean energy increases
$e_U$ and decreases the relative covariance spectrum.

\paragraph{Rank--energy bound.}
\relax \relax \relax Assume $\operatorname{rank}(\Sigma_U)>0$.
Only $\operatorname{rank}(\Sigma_U)$ \relax relative eigenvalues are nonzero, and their sum is
\relax \relax \relax $d_Uf_U$\relax .
Applying Jensen's inequality to these nonzero terms gives
\[
\sum_{i=1}\relax \relax \relax ^{\operatorname{rank}(\Sigma_U)}\relax \log(1+\gamma \relax \relax \relax \nu\relax _i)
\leq
\relax \relax \relax \operatorname{rank}(\Sigma\relax _U\relax )\relax \log\left(
1+\relax \relax \relax \frac{\gamma d_Uf_U}{\operatorname{rank}(\Sigma_U)}
\relax \right).
\]
Thus,
\begin{equation}
\relax \relax \relax F\relax _{\mathrm{exc}}(U)
\leq
\relax \relax \relax \frac{\operatorname{rank}(\Sigma_U)}{m_U}
\frac{
\log\left(1+\gamma d_Uf_U/\operatorname{rank}(\Sigma_U)\right)
}{
\log(1+\gamma d_U/m_U)
}\relax .
\label{eq:app_ahead_excitation_rank_bound}
\end{equation}
Equality holds if and only if all positive covariance eigenvalues
are equal.
If \relax \relax \relax $\operatorname{rank}(\Sigma_U)=0$\relax , the excitation \relax \relax is zero.

This bound captures both the number of available representation
directions and the fraction of total energy devoted to
sample-dependent variation.
At fixed rank and centered energy, a uniform nonzero spectrum
maximizes excitation\relax \relax .

\subsection{Transmission \relax \relax \relax Function \relax and the \relax \relax \relax Supply-Capacity \relax Bound}
\label{app:ahead_transmission}

Define
\[
\Sigma_{HZ}=\frac{1}{N}\relax \relax \relax \bar{H}\relax ^\top\relax \relax \relax \bar{Z}\relax ,
\]
\[
\tau_H
=
\max\left\{
\relax \relax \relax \frac{\operatorname{tr}(\Sigma_H)}{d_H}\relax ,\epsilon
\right\},
\qquad
\tau_Z
=
\max\left\{
\relax \relax \relax \frac{\operatorname{tr}(\Sigma_Z)}{d_Z}\relax ,\epsilon
\right\},
\]
and
\[
A_H=\Sigma_H+\rho\tau_HI,
\qquad
A_Z=\Sigma_Z+\rho\tau_ZI,
\qquad \rho>0.
\]
The regularized whitened cross-covariance is
\[
K=A_H^{-1/2}\Sigma_{HZ}A_Z^{-1/2}.
\]

Let \relax \relax \relax $m=\min(d_H,d_Z,N-1)$, and let
$s_1,\ldots,s_{\min(d_H,d_Z)}$ \relax denote the singular values of $K$.
The transmission \relax \relax can be written as
\[
\relax \relax \relax k\relax _{\mathrm{tr}}
\relax %
\relax =
\relax \relax \relax \frac{
\sum_{i}\log(1+\gamma s_i^2)
}{
m\log(1+\gamma)
}\relax .
\]

\paragraph{Singular-value bound.}
The joint empirical covariance is positive semidefinite:
\[
\begin{pmatrix}
\Sigma_H & \Sigma_{HZ}\\
\Sigma_{HZ}^\top & \Sigma_Z
\end{pmatrix}
=
\frac{1}{N}
\relax %
\relax \relax \begin{pmatrix}
\bar{H}^\top\\
\bar{Z}^\top
\end{pmatrix}\relax 
\relax %
\relax \relax \begin{pmatrix}
\bar{H} & \bar{Z}
\end{pmatrix}\relax 
\succeq0.
\]
Adding the two positive ridge terms yields
\[
\begin{pmatrix}
A_H & \Sigma_{HZ}\\
\Sigma_{HZ}^\top & A_Z
\end{pmatrix}
\succ0.
\]
Its Schur complement satisfies
\[
A_Z-\Sigma_{HZ}^\top A_H^{-1}\Sigma_{HZ}\succ0.
\]
Congruence by $A_Z^{-1/2}$ then gives
\[
I-K^\top K\succ0.
\]
Therefore, \relax \relax \relax $0\leq s_i<1$ \relax for every $i$.

\relax \relax \relax Note that $\operatorname{rank}(\Sigma_{HZ})\leq m$\relax .
Both whitening matrices are invertible, so $K$ has exactly \relax \relax \relax $\operatorname{rank}(\Sigma_{HZ})$
\relax nonzero singular values.
It follows that
\begin{equation}
0\leq \relax \relax \relax k\relax _{\mathrm{tr}}
\relax %
\relax \leq\relax \relax \relax \frac{\operatorname{rank}(\Sigma_{HZ})}{m}\relax \leq1.
\label{eq:app_ahead_transmission_rank_bound}
\end{equation}
For $\rho>0$ and \relax \relax \relax $\operatorname{rank}(\Sigma_{HZ})>0$\relax , the middle inequality is strict.
Transmission capacity is zero if and only if the cross-covariance
is zero.

The denominator counts all \relax \relax \relax interface modes that the batch can support\relax .
Strong transmission within a small active subspace therefore
retains a capacity deficit associated with the missing directions.

\paragraph{Finite ridge and invariance.}
Transmission capacity is invariant to translations and independent
orthogonal transformations at the two endpoints.
When the corresponding scale floors are inactive, it is also
invariant to independent nonzero global rescalings.

A finite ridge contracts the whitened correlation strengths.
For example, if $H=Z$, the covariance is isotropic, and the scale
floor is inactive, then
\[
K=\frac{1}{1+\rho}I
\]
and
\[
\relax \relax \relax k\relax _{\mathrm{tr}}
=
\frac{
\log\left(1+\gamma/(1+\rho)^2\right)
}{
\log(1+\gamma)
}
<1.
\]
The normalization thus refers to ideal transmission in the
vanishing-ridge limit.

The ridge is relative to the current covariance scale and remains
on the automatic-differentiation path. In particular, under
$H\mapsto aH$ with an inactive floor,
$\Sigma_H\mapsto a^2\Sigma_H$,
$\tau_H\mapsto a^2\tau_H$, and
$\Sigma_{HZ}\mapsto a\Sigma_{HZ}$, so the channel score and its
radial directional derivative are invariant. Detaching $\tau_H$
would leave the forward value unchanged at the current point but
would break this backward invariance, creating a spurious radial
gradient whenever transmission is the \relax \relax \relax smaller supply term\relax .

\relax \relax \relax $F_{\mathrm{tr}}$ \relax measures second-order \relax \relax transmission.
Zero cross-covariance does not exclude nonlinear dependence
between the representations.

\begin{proof}[Supply-bound derivation]
By definition,
\[
k(H,\operatorname{sg}(Z))
=
\min\left\{
\relax \relax \relax k\relax _{\mathrm{exc}}\relax \relax ,\relax \relax \relax k\relax _{\mathrm{tr}}
\relax %
\relax \right\}.
\]
Combining
Equation~\ref{eq:app_ahead_excitation_rank_bound}
with
Equation~\ref{eq:app_ahead_transmission_rank_bound}
gives
\[
k(H,\operatorname{sg}(Z))
\leq
\min\left\{
\relax \relax \relax \frac{\operatorname{rank}(\Sigma_H)}{m_H}
\frac{
\log\left(1+\gamma d_Hf_H/\operatorname{rank}(\Sigma_H)\right)
}{
\log(1+\gamma d_H/m_H)
}\relax ,
\relax \relax \relax \frac{\operatorname{rank}(\Sigma_{HZ})}{m}
\relax \right\}.
\]
If \relax \relax \relax $\operatorname{rank}(\Sigma_H)=0$, then $k_{\mathrm{exc}}=0$ \relax and hence $k(H,\operatorname{sg}(Z))=0$.

The minimum corresponds to simultaneous support by both
\relax \relax \relax supply terms:
\relax \[
k\geq q
\quad\Longleftrightarrow\quad
\relax \relax \relax k\relax _{\mathrm{exc}}\relax \relax \geq q
\ \text{and}\
\relax \relax \relax k\relax _{\mathrm{tr}}\relax %
\relax \geq q.
\]
This is the joint support criterion used in the construction.
\end{proof}

\subsection{Demand Calibration and Log-Space Interpolation}
\label{app:ahead_demand}

Assume that $\mathcal{D}(Z)\geq0$ and that the zero-representation
reference is finite and positive:
\[
\mathcal D\relax \relax \relax (0_{N\times d_Z})\relax >0.
\]

\paragraph{Reference calibration.}
Consider an affine calibration of \relax \relax \relax the Demand Regularizer value satisfying
\relax \[
\mathcal{D}(Z)=0\mapsto q\relax \relax \relax _{\mathrm{reg}}\relax =1,
\qquad
\mathcal{D}(Z)=\mathcal D\relax \relax \relax (0)\relax \mapsto q\relax \relax \relax _{\mathrm{reg}}\relax =0.
\]
These conditions uniquely determine the affine map
\relax \relax \relax $1-\mathcal{D}(Z)/\mathcal D(0)$\relax .
Restricting its range to the unit interval gives
\[
q\relax \relax \relax _{\mathrm{reg}}
\relax =
\operatorname{clip}\left(
1-\relax \relax \relax \frac{\mathcal{D}(Z)}{\mathcal D(0)}\relax ,0,1
\right).
\]

For any constant $c>0$,
\[
1-\relax \relax \relax \frac{c\mathcal{D}(Z)}{c\mathcal D(0)}
\relax =
1-\relax \relax \relax \frac{\mathcal{D}(Z)}{\mathcal D(0)}\relax .
\]
The calibrated demand is therefore invariant to the overall
units of \relax \relax \relax $\mathcal D$\relax .

The second demand \relax \relax \relax term is
\relax \[
q\relax \relax \relax _{\mathrm{exc}}\relax =\relax \relax \relax F\relax _{\mathrm{exc}}(\operatorname{sg}(Z)).
\]
It uses the relative covariance structure already formed at the
endpoint to \relax \relax \relax complement $q_{\mathrm{reg}}$\relax .
Excitation and transmission have different spectral
normalizations, so their values need not obey a fixed ordering.

\paragraph{Log-space barycenter.}
For \relax \relax \relax $q_{\mathrm{reg}},q_{\mathrm{exc}}>0$, consider
\relax \[
F_\alpha(q)
=
(1-\alpha)(\log q-\log q\relax \relax \relax _{\mathrm{reg}}\relax )^2
+
\alpha(\log q-\log q\relax \relax \relax _{\mathrm{exc}}\relax )^2,
\qquad
\alpha\in[0,1].
\]
Let $y=\log q$, \relax \relax \relax $a=\log q_{\mathrm{reg}}$, and $b=\log q_{\mathrm{exc}}$\relax .
Completing the square gives
\[
F_\alpha(q)
=
\left(
y-[(1-\alpha)a+\alpha b]
\right)^2
+
\alpha(1-\alpha)(a-b)^2.
\]
The second term is independent of $q$.
The unique minimizer therefore satisfies
\[
\log q_\alpha^\star
=
(1-\alpha)\log q\relax \relax \relax _{\mathrm{reg}}\relax +\alpha\log q\relax \relax \relax _{\mathrm{exc}}\relax ,
\]
or equivalently,
\[
q_\alpha^\star=q\relax \relax \relax _{\mathrm{reg}}\relax ^{1-\alpha}q\relax \relax \relax _{\mathrm{exc}}\relax ^\alpha.
\]

Under the metric
$\mathrm{d}\ell^2=(\mathrm{d}q/q)^2$
on the positive half-line, this interpolation is a geodesic
between \relax \relax \relax $q_{\mathrm{reg}}$ and $q_{\mathrm{exc}}$\relax .
It also satisfies
\[
\min(q\relax \relax \relax _{\mathrm{reg}}\relax ,q\relax \relax \relax _{\mathrm{exc}}\relax )
\leq q_\alpha^\star
\leq\max(q\relax \relax \relax _{\mathrm{reg}}\relax ,q\relax \relax \relax _{\mathrm{exc}}\relax ).
\]
The \relax \relax \relax two terms $q_{\mathrm{reg}}$ and $q_{\mathrm{exc}}$ \relax are recovered at
$\alpha=0$ and $\alpha=1$, respectively.

\paragraph{Minimax interpretation of the symmetric choice.}
For any $y\in\mathbb{R}$,
\[
|a-b|
\leq |a-y|+|y-b|
\leq2\max\{|a-y|,|y-b|\}.
\]
The largest deviation is thus at least $|a-b|/2$,
with equality only at $y=(a+b)/2$.
Consequently,
\[
\relax \relax \relax \sqrt{q_{\mathrm{reg}}q_{\mathrm{exc}}}
\relax =
\underset{q>0}{\operatorname{argmin}}\;
\max\left\{
\left|\log\relax \relax \relax \frac{q}{q_{\mathrm{reg}}}\relax \right|,
\left|\log\relax \relax \relax \frac{q}{q_{\mathrm{exc}}}\relax \right|
\right\}.
\]
The choice $\alpha=\tfrac12$ is therefore the symmetric center of
the two \relax \relax \relax demand terms \relax in log space.
Adaptation follows from changes in \relax \relax \relax $\mathcal D(Z)$ and in the
endpoint \relax geometry, without learning the interpolation weight.

\paragraph{Zero-demand boundaries.}
For fixed $0<\alpha<1$, the geometric demand has a continuous
extension to the nonnegative quadrant:
\[
q_\alpha^\star=0
\qquad\text{whenever}\qquad
q\relax \relax \relax _{\mathrm{reg}}\relax q\relax \relax \relax _{\mathrm{exc}}\relax =0.
\]
The two boundary methods are defined separately by
\relax \relax \relax $q_0=q_{\mathrm{reg}}$ and $q_1=q_{\mathrm{exc}}$\relax .

For numerical evaluation at the boundary, the implementation uses
Equation~\ref{eq:ahead_smoothed_demand}. Thus
$\widetilde q_{\alpha,\epsilon}>0$ for $\epsilon>0$, while
$\widetilde q_{\alpha,\epsilon}\to q_\alpha^\star$ as
$\epsilon\to0$ wherever the ideal demand is positive. This
distinguishes the geometric object from its finite-precision
realization. For the symmetric default, $\epsilon\leq q\leq1+\epsilon$: \relax \relax \relax $q_{\mathrm{reg}}=q_{\mathrm{exc}}=1$ \relax gives $q=1+\epsilon$, \relax \relax \relax $q_{\mathrm{reg}}=0$ and $q_{\mathrm{exc}}=c>0$ \relax give $q=\sqrt{\epsilon(c+\epsilon)}=O(\sqrt{\epsilon})$, and \relax \relax \relax $q_{\mathrm{reg}}=q_{\mathrm{exc}}=0$ \relax gives $q=\epsilon$. At this last corner, $k=0$ yields $\mathcal S=(\log 2)^2$; an exactly constant $H$ still has zero first-order capacity gradient, as shown below.

\subsection{Geometric Derivation of the One-Sided \relax \relax \relax Supply Regularizer\relax }
\label{app:ahead_bridge}

Fix a demand $q\geq0$ and write
\[
c_1=\relax \relax \relax k\relax _{\mathrm{exc}}\relax \relax ,
\qquad
c_2=\relax \relax \relax k\relax _{\mathrm{tr}}\relax \relax ,
\relax %
\relax \qquad
k=\min(c_1,c_2).
\]
Introduce log-supply coordinates
\[
u_i=\log(c_i+\epsilon),
\qquad
b_q=\log(q+\epsilon).
\]
The two support conditions define the region
\[
\mathcal{C}_q=[b_q,\infty)^2.
\]

For any $v\in\mathcal{C}_q$,
\[
\|u-v\|_\infty
\geq
\max_{i=1,2}[b_q-u_i]_+.
\]
The choice $v_i=\max(u_i,b_q)$ attains this lower bound.
Therefore,
\[
\begin{aligned}
\operatorname{dist}_{\infty}^{\,2}(u,\mathcal{C}_q)
&=
\max_{i=1,2}[b_q-u_i]_+^2\\
&=
[b_q-\min(u_1,u_2)]_+^2\\
&=
\left[
\log\frac{q+\epsilon}{k+\epsilon}
\right]_+^2.
\end{aligned}
\]
This proves the distance representation in
Equation~\ref{eq:ahead_bridge}.

\relax \relax \relax Hence, the Supply Regularizer \relax equals the largest squared log-violation of the two
supply constraints and satisfies
\[
\mathcal S=0
\quad\Longleftrightarrow\quad
c_1\geq q
\ \text{and}\
c_2\geq q.
\]
It exerts no additional geometric pressure once both support
conditions hold.
Otherwise, its correction is determined by the \relax \relax \relax smaller supply term\relax .

The distance is defined in capacity coordinates.
The network parameterization determines how these coordinates
change under parameter updates.

\subsection{Gradient Paths and Boundary Behavior}
\label{app:ahead_gradients}

\paragraph{Response to insufficient supply.}
For fixed demand, let
\[
\Delta(k)
=
\left[
\log\frac{q+\epsilon}{k+\epsilon}
\right]_+.
\]
Then
\[
\frac{\partial\mathcal S}{\partial k}
=
-\frac{2\Delta(k)}{k+\epsilon}.
\]
Within the active region $k<q$,
\[
\frac{\partial^2\mathcal S}{\partial k^2}
=
\frac{
2\left(
1+\log\frac{q+\epsilon}{k+\epsilon}
\right)
}{
(k+\epsilon)^2
}
>0.
\]
At fixed demand, the magnitude of the negative scalar derivative
therefore increases as supply decreases.
The first derivative is zero for $k\geq q$ and is continuous at
$k=q$.

When the two capacities differ,
\[
\nabla_H\mathcal S
=
-\frac{2\Delta(k)}{k+\epsilon}
\relax %
\relax \relax \begin{cases}
\nabla_H k_{\mathrm{exc}},
&
k_{\mathrm{exc}}<k_{\mathrm{tr}},\\[3pt]
\nabla_H k_{\mathrm{tr}},
&
k_{\mathrm{tr}}<k_{\mathrm{exc}}.
\end{cases}\relax 
\]
At a tie, a generalized gradient is selected from the convex hull
of the active branch gradients.

\paragraph{Detached references and endpoint roles.}
The \relax \relax \relax Supply Regularizer \relax uses a stopped endpoint \relax \relax \relax and a stopped demand capacity:
\relax \[
\mathcal S\bigl(H,\operatorname{sg}(Z)\relax %
\relax \bigr)=\left[\log\frac{q+\epsilon}{k+\epsilon}\right]_+^2.
\]
The endpoint tensor and demand are fixed references during the
current backward pass. The covariance-relative ridge scale derived
from $H$ remains differentiable; the $Z$ side receives no gradient
because the endpoint tensor has already been stopped.

With separate encoder and projector parameters $\theta$ and $\phi$,
the \relax \relax \relax Supply Regularizer\relax 's automatic-differentiation gradients satisfy
\[
\nabla_\phi^{\mathrm{AD}}\mathcal S=0,
\qquad
\nabla_\theta^{\mathrm{AD}}\mathcal S
=
J\relax \relax \relax _{E_\theta}\relax ^{\top}
\nabla_H\mathcal S.
\]
The same loss decomposition as Equation~\ref{eq:ahead_objective} remains valid in automatic differentiation:
\[
\mathcal{L}\relax \relax =\mathcal{L}_{\mathrm{task}}+\relax \relax \lambda_D\mathcal{D}(Z)+\lambda_S\mathcal{S}\bigl(H,\operatorname{sg}(Z)\relax %
\relax \bigr)\relax \relax \relax .
\relax \]
The task objective and \relax \relax \relax Demand Regularizer \relax retain their normal
gradients.
Stopping the references directs the \relax \relax \relax Supply Regularizer toward the current
supply capacity\relax , rather than directly reducing demand to eliminate
the mismatch.

\paragraph{Feature gradient of excitation\relax \relax .}
Assume that the energy floor is inactive, with
$e_H^{\mathrm{raw}}>\epsilon$.
Define
\[
M=I+\gamma\Sigma_H/e_H,
\qquad
a_H
=
\relax \relax \relax \frac{1}{d_H}
\relax \operatorname{tr}\left(
\frac{\Sigma_H}{e_H}M^{-1}
\right).
\]
Differentiating excitation \relax \relax \relax gives
\relax \[
\mathrm{d}\relax \relax \relax k\relax _{\mathrm{exc}}
=
\relax \relax \relax \frac{\gamma}{m_H\log(1+\gamma d_H/m_H)}
\relax \left[
\frac{
\operatorname{tr}(M^{-1}\mathrm{d}\Sigma_H)
}{e_H}
-
\frac{
\operatorname{tr}(M^{-1}\Sigma_H)
}{e_H^2}
\,\mathrm{d}e_H
\right].
\]
Substituting the covariance and energy differentials yields
\begin{equation}
\nabla_H \relax \relax \relax k\relax _{\mathrm{exc}}
=
\relax \relax \relax \frac{
2\gamma
}{
Ne_Hm_H\log(1+\gamma d_H/m_H)
}
\relax \left(
\relax \relax \relax \bar{H}M\relax ^{-1}-a_HH
\right).
\label{eq:app_ahead_excitation_gradient}
\end{equation}

Along a relative covariance eigenmode \relax \relax \relax $\nu_i$\relax , the spectral response
of $M^{-1}$ is
\[
\relax \relax \relax \frac{1}{1+\gamma \nu_i}\relax .
\]
Weaker directions therefore receive larger spectral response
weights.
The second term compensates for changes along the global scaling
direction, giving
\[
\left\langle
H,\nabla_H \relax \relax \relax k\relax _{\mathrm{exc}}
\right\rangle_F=0.
\]
This identity describes the local scale property of excitation
capacity itself in representation space.

\relax \paragraph{Gradient action.}

Let $\nabla_H k$ denote the supply gradient
under the stated stop-gradient conventions. Where the selected
term is differentiable,
\begin{equation}
\nabla_H\bigl(\lambda_S\mathcal S\bigr)=-a\,\nabla_H k,
\qquad
a=\frac{2\lambda_S}{k+\epsilon}
\left[\log\frac{q+\epsilon}{k+\epsilon}\right]_+.
\label{eq:ahead_adaptive_gradient}
\end{equation}
The deficit sets the correction strength $a$, and the smaller supply
term selects its direction: excitation or transmission. Once $k\geq q$, $a=0$.
The encoder gradient is
$-aJ_{E_\theta}^{\top}\nabla_H k$, bypassing the projector Jacobian
while retaining dependence on the encoder Jacobian and the supply
gradient. The Supply Regularizer has no direct gradient on
projector parameters.

\paragraph{Spectral equalization.}
To characterize the excitation term, treat $H$ as a directly
optimized variable and define $\nu_i=\lambda_i(\Sigma_H)/e_H$.
\begin{proposition}[Spectral equalization under excitation limitation]
\label{prop:spectral}
Assume $\lambda_S>0$, an inactive energy floor, and
$0<k_{\mathrm{exc}}<\min\{q,k_{\mathrm{tr}}\}$
throughout the interval considered. The correction-only feature
flow $\dot H=a\nabla_H k_{\mathrm{exc}}$, with $a>0$, preserves
total feature energy, makes the common-mean norm non-increasing,
and satisfies
\begin{equation}
\frac{\mathrm d}{\mathrm dt}\log\frac{\nu_i}{\nu_j}
=\frac{4a\gamma}{Ne_H\,m_H\log(1+\gamma d_H/m_H)}
\left(\frac{1}{1+\gamma\nu_i}-\frac{1}{1+\gamma\nu_j}\right)
\leq0
\label{eq:ahead_excitation_dynamics}
\end{equation}
for positive relative eigenvalues $\nu_i\geq\nu_j>0$.
\end{proposition}
Thus, weaker positive directions receive larger relative growth
rates, making the positive-spectrum condition number
non-increasing while preserving total energy. Exact zero modes
remain zero. When excitation is the smaller supply term, \textbf{SADL} preferentially strengthens weak but nonzero spectral directions, counteracting incipient collapse before sample variation is completely lost.

\relax \paragraph{Zero-demand limit and exact degeneracy.}
The ideal and implemented demands are, respectively,
\[
q^\star=q\relax \relax \relax _{\mathrm{reg}}\relax ^{1-\alpha}q\relax \relax \relax _{\mathrm{exc}}\relax ^\alpha,
\qquad
q=\exp\!\left((1-\alpha)\log(q\relax \relax \relax _{\mathrm{reg}}\relax +\epsilon)
+\alpha\log(q\relax \relax \relax _{\mathrm{exc}}\relax +\epsilon)\right),
\]
and the \relax \relax \relax Supply Regularizer \relax uses
\[
\mathcal S
=
[\log(q+\epsilon)-\log(k+\epsilon)]_+^2.
\]
If either reference is zero, then $q^\star=0$ for an interior
interpolation weight, but the implemented $q$ remains at an
$\epsilon$-controlled positive scale. Exact \relax \relax \relax Supply Regularizer \relax deactivation is
therefore determined by $k\geq q$ in the smoothed implementation,
not by substituting the ideal boundary value into the code.

An exactly constant encoder representation has a separate
boundary behavior.
If $H=\mathbf{1}h_0^\top$, then
\[
\relax \relax \relax \bar{H}\relax =0,\qquad
\Sigma_H=0,\qquad
\Sigma_{HZ}=0.
\]
Both \relax \relax \relax supply terms \relax have zero first-order gradients with respect
to $H$ at this point.
The scalar supply response and the resulting feature gradient
must therefore be distinguished.
The original task objective and \relax \relax \relax Demand Regularizer \relax remain in
the complete objective, while the \relax \relax \relax Supply Regularizer \relax responds to the nonzero
demand and structure formed during training.

\relax %
\relax \section{Numerical Implementation}
\label{app:ahead_implementation}

\paragraph{Sample alignment and statistical precision.}
Sequence representations are flattened in the same batch--time
order before computing cross-covariance.
If the encoder \relax \relax \relax representation has shape
$B\times S\times d_H$ \relax and the projected output has shape
\relax \relax \relax $S\times B\times d_Z$\relax , the first two projected dimensions are
exchanged before both tensors are flattened into $N=BS$ samples.

Centering, covariance estimation, whitening, and log-determinant
operations within the \relax \relax \relax Supply Regularizer \relax are performed in FP32 with autocast
disabled.
\relax \relax \relax Demand Regularizers \relax retain their respective computation paths.

\paragraph{Cholesky whitening.}
The implementation uses Cholesky factorizations and triangular
solves instead of explicitly forming inverse square roots.
Let
\[
A=L_HL_H^\top,
\qquad
B=L_ZL_Z^\top,
\]
and compute
\[
\widetilde K
=
L_H^{-1}\Sigma_{HZ}L_Z^{-\top}.
\]
The matrices $\widetilde K$ and $K$ differ only by orthogonal
transformations on the left and right, and therefore have the
same singular values.

Using
\[
\det\left(I+\gamma\widetilde K^\top\widetilde K\right)
=
\det\left(I+\gamma\widetilde K\widetilde K^\top\right),
\]
the log-determinant can be computed from the smaller Gram matrix,
whose dimension is \relax \relax \relax $\min(d_H,d_Z)$\relax .

The ridge scales are computed from differentiable mean covariance
energies:
\[
\tau_U
=
\max\left\{
\frac{\operatorname{tr}(\Sigma_U)}{d_U}
, 
\epsilon
\right\}.
\]
Both this relative ridge scale and the total energy in excitation
capacity retain their gradients, subject to their numerical floors.

\paragraph{Gradient-semantic audit.}
The \relax \relax \relax implementation used for the reported experiments \relax keeps the relative ridge scale
differentiable. \relax \relax \relax An earlier control implementation \relax detached this scalar,
which preserves the forward score but can create a radial gradient
when transmission is the \relax \relax \relax smaller supply term.
A matched epoch 1 replay \relax produced identical values for all 204
common diagnostics through step 5,650 of 5,695, excluding wall time.
Thus the audited headline configurations are unchanged, while the
released implementation follows the scale-invariant gradient
semantics stated above.

\paragraph{Reference energy and smoothing.}
The reference energy \relax \relax \relax $\mathcal D(0)$ \relax is evaluated on an all-zero tensor with
the same shape as the endpoint representation and cached for the
current configuration.
The evaluation does not update model parameters.
Its random-number state is saved and restored so that reference
calibration does not alter the subsequent training random sequence.

The energy floors and logarithmic smoothing serve different
numerical roles.
The former define stable capacity normalizations, while the latter
keep the comparison finite at zero supply.
Equation~\ref{eq:ahead_smoothed_demand} also places $\epsilon$
inside the implemented interpolation. This is a numerical
regularization of the ideal demand, not a change to the ideal
log-geodesic definition.

\begin{algorithm}[t]
    \caption{Single-batch computation of SADL}
    \label{alg:ahead}
    \begin{algorithmic}[1]
        \Require Aligned representations $H,Z$;
        task loss $\mathcal{L}_{\mathrm{task}}$;
        \relax \relax \relax Demand Regularizer \relax $\mathcal{D}$
        \Require Reference \relax \relax \relax $\mathcal D(0)$\relax ;
        weights $\lambda_D,\lambda_S$;
        smoothing $\epsilon$
        \Ensure Total \relax \relax \relax loss $\mathcal{L}$

\relax 

        \State Evaluate $\mathcal{D}(Z)$ with its endpoint gradient retained

        \State
        \relax \relax \relax $q_{\mathrm{reg}}\gets
        \operatorname{clip}(1-\mathcal D(Z)/\mathcal D(0),0,1)$

\relax 

        \State
        \relax \relax \relax $q_{\mathrm{exc}}\gets
        F_{\mathrm{exc}}(\operatorname{sg}(Z))$

\relax 

        \State
        \relax \relax \relax $q\gets
        \operatorname{sg}\!\left(
        \exp\{[\log(q_{\mathrm{reg}}+\epsilon)+\log(q_{\mathrm{exc}}+\epsilon)]/2\}
        \right)$

\relax 

        \State \relax \relax \relax $k_{\mathrm{exc}}\gets F_{\mathrm{exc}}(H)$, $k_{\mathrm{tr}}\gets F_{\mathrm{tr}}(H,\operatorname{sg}(Z))$

\relax 

        \State
        \relax $k\gets
        \min\left\{
        k_{\mathrm{exc}},
        k_{\mathrm{tr}}
        \right\}$

        \State
        \relax $\mathcal S\gets
        [\log(q+\epsilon)-\log(k+\epsilon)]_+^2$

        \State \relax %
\relax \Return \relax \relax \relax $\mathcal{L}_{\mathrm{task}}+\lambda_D\mathcal D(Z)+\lambda_S\mathcal S$

\relax 

        \Statex
        \textit{Capacity statistics and matrix factorizations
        use FP32 with autocast disabled.}
    \end{algorithmic}
\end{algorithm}

\paragraph{Configuration in the robotic control experiments.}
Table~\ref{tab:app_ahead_hyperparameters} lists the \relax \relax \relax Supply Regularizer \relax settings
used in the robotic control experiments.
All three \relax \relax \relax Demand Regularizers \relax share the same SADL parameters.

\begin{table}[t]
    \centering
    \caption{
        SADL configuration in the robotic control experiments.
        Endpoint weights multiply the corresponding raw
        regularizer implementations.
    }
    \label{tab:app_ahead_hyperparameters}
    \small
    \begin{tabular*}{\linewidth}{@{\extracolsep{\fill}}lc@{}}
        \toprule
        Parameter & Value \\
        \midrule
        Demand interpolation weight $\alpha$ & $0.5$ \\
        Capacity parameter $\gamma$ & $9$ \\
        Ridge coefficient $\rho$ & $10^{-3}$ \\
        Numerical floor and log smoothing $\epsilon$ & $10^{-6}$ \\
        Supply Regularizer weight $\lambda_S$ & $0.01$ \\
        SIGReg $\lambda_D$ & $0.03$ \\
        VISReg $\lambda_D$ & $1.5$ \\
        QQWorld $\lambda_D$ & $3.5$ \\
        \bottomrule
    \end{tabular*}
\end{table}

\relax %
\relax \section{Experimental Setup and Statistical Protocols}
\label{app:ahead_setup}

\subsection{World-Model Training}
\label{app:ahead_training}

Single-task experiments train one model per task.
Joint experiments share the encoder and projector while
retaining task-specific prediction and control modules.
BatchNorm running buffers and shared-parameter gradients
are averaged across task workers.

Capacity statistics are computed from each task's local batch
before gradient aggregation.
The \relax \relax \relax Supply Regularizer \relax therefore acts on within-task statistics,
rather than a covariance pooled across tasks.
\relax Across 25,396 audited capacity records, excitation was the bottleneck
in every case.

\relax 

\relax %
\relax \relax \begin{table}[!htbp]
\relax \centering
\caption{Core world-model training configuration.}
\label{tab:app_ahead_configuration}
\SADLTableStyle
\begin{tabularx}{\linewidth}{@{}lL@{}}
\toprule
Configuration & Setting \\
\midrule
Tasks & PushT, Cube, Reacher, TwoRoom \\
Training seeds & 0, 42, 3072 \\
Encoder & ViT-Tiny, trained from scratch \\
Encoder representation dimension & 192 \\
Input image size & $224\times224$ \\
Patch size & $14\times14$ \\
Projector & MLP with BatchNorm; hidden dimension 2,048 \\
Projected dimension & 192 \\
Predictor & Six transformer blocks \\
Training clip length & Eight frames \\
Conditioning history & Three steps \\
Numerical precision & BF16 mixed precision \\
Batch size per task & 256 clips \\
Joint batch size & 1,024 clips across four workers \\
Optimizer & AdamW \\
Learning rate & $5\times10^{-4}$ \\
Weight decay & $10^{-3}$ \\
\relax \relax \relax $\mathcal D$-only \relax encoder backend & MATH \\
SADL and structural-ablation encoder backend & Flash \\
Predictor backend & MATH \\
\bottomrule
\end{tabularx}
\end{table}

INTACT retains its forward-prediction, inverse-action,
and goal-displacement objectives with weights
$1.0$, $0.1$, and $0.05$, respectively.
LeWM uses forward prediction without the INTACT
inverse-action and goal-action objectives.
The shared \relax \relax \relax Supply Regularizer \relax parameters are given in
Table~\ref{tab:app_ahead_hyperparameters}.

Joint \relax \relax \relax $\mathcal D$-only \relax runs extend to ten epochs.
\relax \relax \relax SADL and exc-only \relax runs extend to five epochs.
The \relax \relax \relax tr-only, two-sided\relax , one-sided-linear,
and \relax \relax \relax $\mathcal S$-only \relax comparisons use two-epoch runs.
Single-task comparisons use the fixed first-epoch checkpoint.

\subsection{Control Interfaces and Success Criteria}
\label{app:ahead_evaluation}

\paragraph{Control execution.}
Let $T_H$ denote the planning horizon and $R_H$ the number of actions executed before replanning.
INTACT uses Direct control with $T_H=R_H=5$.
Each replanning step executes five distinct predicted actions.
The initial context contains the actual actions preceding
the evaluation start, with zero padding only before a true
episode boundary.
Subsequent context entries contain executed actions.

LeWM uses Pure-CEM with 300 candidate sequences per iteration,
30 iterations, and 30 retained candidates.
Both horizons are five, and no actor warm start is used.
\relax The initialization rows in Table~\ref{tab:where} use these same
Direct and Pure-CEM interfaces. LeWM gives the same initialization SR
for its three seeds because the nearly constant latent makes the
Pure-CEM cost essentially flat.

\relax 

\paragraph{CLEAR-LeWM.}
Moderate and Strict evaluation use a goal offset of 25
primitive steps and an evaluation budget of 50 primitive actions.
Sampling is episode-balanced over the \texttt{all} split,
and initially successful pairs are excluded.
These evaluations are not held-out-episode tests.
TwoRoom additionally uses cross-room pairs, clean source windows,
and swept-collision handling.

\begin{table}[t]
\centering
\caption{
Task-level CLEAR-LeWM success predicates.
``Hold'' gives the number of consecutive primitive steps
satisfying the predicate.
Thresholds use the respective environment coordinate systems.
}
\label{tab:app_success_criteria}
\SADLTableStyle
\begin{tabularx}{\linewidth}{@{}lLL@{}}
\toprule
Task & Moderate & Strict \\
\midrule
PushT
& Joint agent-and-block position error $<20$ native pixel units;
block angle error $<20^\circ$; hold one step.
& Block position error $<10$; block angle error $<10^\circ$;
hold three steps. \\
Cube
& Position error $\leq0.04$ m; no orientation constraint;
hold one step.
& Position error $\leq0.03$ m and cube-symmetry-aware orientation
error $\leq15^\circ$; hold three steps. \\
Reacher
& Maximum joint error $<0.05$ rad; periodic shoulder and
unwrapped elbow error; hold one step.
& End-effector distance $\leq0.01$ m; hold two steps. \\
TwoRoom
& Goal distance $<16$ native units; hold one step.
& Goal distance $<8$, valid collision-free route,
doorway crossing, and correct goal room; hold one step. \\
\bottomrule
\end{tabularx}
\end{table}

\paragraph{Official.}
Official retains the upstream evaluator's row-uniform
sampling, initially successful pairs, and first-hit success semantics.
Its base thresholds are $20$ position units and $20^\circ$
for PushT, $0.04$ m without an orientation condition for Cube,
$0.05$ rad using raw joint errors for Reacher, and
$16$ native units for TwoRoom.
The configured goal offset and evaluation budget are
25 and 50, respectively.
The evaluator retains its historical action-loop compatibility
convention.
Official and CLEAR results are reported separately.

\subsection{Aggregation and Pairing}
\label{app:ahead_aggregation}

For each training seed and task, evaluation uses seeds
$0$, $1$, and $42$, with 100 episodes per evaluation seed.
Let $s_{k,t,e}$ denote the corresponding success percentage.
We compute
\[
s_{k,t}=\frac{1}{3}\sum_{e=1}^{3}s_{k,t,e},
\qquad
m_k=\frac{1}{4}\sum_{t=1}^{4}s_{k,t},
\]
and report
\[
\bar m=\frac{1}{3}\sum_{k=1}^{3}m_k,
\qquad
\operatorname{SD}(m)
=
\sqrt{\frac{1}{2}\sum_{k=1}^{3}(m_k-\bar m)^2}.
\]
Task-level standard deviations use the same training-seed
aggregation.
Evaluation episodes are not treated as independent
training replicates.

For paired comparisons, differences are computed before aggregation:
\[
\Delta_k=m_k^A-m_k^B.
\]
We report the mean and sample standard deviation of $\Delta_k$.
Rounding is applied only after aggregation.

A complete four-task result contains all 36
task--training-seed--evaluation-seed records.
Checkpoint identity, training epoch, control interface,
and success criterion are checked before aggregation.

\relax %
\relax \section{Collapse Phenomenology and Path Diagnosis}
\label{sec:app_collapse}

\relax This appendix follows the same sequence as Sections~\ref{sec:ahead_projection_path} and~\ref{sec:where}. At initialization, sample variation carries about $0.01\%$ of the energy of the encoder representation $H$ in both LeWM and INTACT (Table~\ref{tab:where}). A BatchNorm projector removes the batch mean and scale of $H$, so the Demand Regularizer and the task loss, which reach $H$ only through the projector, have essentially no gradient along this shared component (Equation~\ref{eq:ahead_projection_response}). With $\mathcal D$-only training the encoder therefore stays dominated by it: with MATH every training seed ends \emph{degraded} ($0.05\leq f_H<0.5$), and with Flash part of the seeds \emph{collapse} ($f_H<0.05$), five of ten LeWM seeds with SIGReg and all ten without a Demand Regularizer (Table~\ref{tab:app_per_seed_outcomes}). Over training, the encoder drops below a fraction of its peak before or together with $Z$, never after it (Table~\ref{tab:app_collapse_timing}). With \textbf{SADL} every training seed ends \emph{normal}, and continuing a near-collapsed checkpoint with \textbf{SADL} can repair it (Appendix~\ref{sec:app_rescue}).

\relax We proceed from the observed failure through controlled component and
constraint-placement tests to an operational definition and a \relax \relax \relax supply capacity
\relax explanation. Fixed probes identify where variation is lost; a paired ten-seed
study estimates \relax \relax \relax paired rank-drop \relax events.

\subsection{Observed failure: \relax \relax \relax Demand Regularizers \relax do not certify the \relax \relax \relax encoder representation\relax }

In one jointly trained four-task INTACT--VISReg run, \relax \relax \relax official \relax Direct success
decreases from $55.67\%$ at \relax \relax \relax epoch 1 \relax to $24.75\%$ at \relax \relax \relax epoch 5. \relax At the same checkpoints,
\relax \relax \relax $\operatorname{erank}(H)$ \relax decreases from $13.74$ to $6.64$, while the mean
inter-sample cosine $c_H$ increases from $.910$ to $.997$. The control failure
coincides with increasingly indistinguishable encoder \relax \relax \relax representations\relax ; endpoint
regularization loss alone does not establish representation utility.

The layer-wise audit fixes 256 trajectories per task and eight frames per
trajectory. At matched observation indices we probe successive visual-encoder
layers, $H=h_{\rm cls}$, projected $Z$, and predicted $\hat Z$, centering
within each task before comparison. Across 37 runs, the cross-run correlation
between effective ranks of $H$ and $Z$ is $.993$. In all 13 observed severe
low-rank runs, the earliest abnormal boundary is already in the visual
encoder; we find no case in which $H$ remains healthy while $Z$ or the
predictor alone first becomes low-rank.

\subsection{Controlled localization of the first numerical divergence}

\paragraph{Component swap.}
We clone each initialization, input sequence, random state, and optimizer,
changing only the attention backend of the visual encoder or latent
predictor. Keeping predictor Flash while switching the encoder to MATH gives
bitwise agreement for all three seeds over 100 steps; the reverse swap
diverges on step one in all three. In one seed, the first detected difference
is an encoder backward-gradient perturbation of $1.49\times10^{-8}$, which
is amplified over subsequent parameter updates. This locates the first
measurable numerical perturbation; long-run training separately establishes
the representation outcome.
\begin{table}[t]
\centering
\caption{Paired 100-step component swaps from identical initialization,
input order, random state, and optimizer. Seeds: 3072, 3073, 3074. The first
divergence is a numerical trajectory event, not itself a \relax \relax \relax paired rank-drop \relax event.}
\label{tab:app_collapse_component}
\begingroup\small\setlength{\tabcolsep}{4pt}\renewcommand{\arraystretch}{1.12}
\begin{tabular*}{\linewidth}{@{\extracolsep{\fill}}llccc@{}}
\toprule
Visual encoder & Latent predictor & Bitwise match & First divergence & Max. parameter gap at step 100 \\
\midrule
MATH & Flash & $3/3$ & Not observed & $0$ \\
Flash & MATH & $0/3$ & Step 1, $3/3$ & $.0556$--$.0671$ \\
\bottomrule
\end{tabular*}
\endgroup
\end{table}

\paragraph{Visual-encoder layer position.}
With predictor MATH and endpoint SIGReg weight $.03$, we vary which encoder
layers use MATH. In independent run A, protecting the final six layers
preserves substantially more variation in $H/Z/\hat Z$ than protecting the
first six. In run B, MATH in layers 10--12 approaches the all-MATH reference,
whereas changing only layers 7--9 does not recover it. Late attention is
thus a sensitive location under this architecture and protocol.
\begin{table}[t]
\centering
\caption{Encoder-layer backend ablation. Runs A and B are independently
trained four-task INTACT models with seed 3072, SIGReg$(Z)$ weight $.03$,
MATH predictor, and the same fixed PushT probe.}
\label{tab:app_collapse_layers}
\begingroup\small\setlength{\tabcolsep}{4pt}\renewcommand{\arraystretch}{1.11}
\begin{tabular*}{\linewidth}{@{\extracolsep{\fill}}llccc@{}}
\toprule
Run & Encoder MATH layers & \relax \relax \relax $\operatorname{erank}(H)$ \relax & \relax \relax \relax $\operatorname{erank}(Z)$ \relax & \relax \relax \relax $\operatorname{erank}(\hat Z)$ \relax \\
\midrule
A & All 12 & 21.13 & 50.18 & 49.14 \\
A & None (all Flash) & 4.65 & 7.65 & 7.74 \\
A & 1--6 & 10.58 & 20.72 & 23.00 \\
A & 7--12 & 20.62 & 51.22 & 51.32 \\
\midrule
B & All 12 & 26.84 & 66.63 & 64.78 \\
B & None (all Flash) & 13.80 & 14.86 & 4.73 \\
B & 7--9 & 7.53 & 12.08 & 2.64 \\
B & 10--12 & 27.09 & 64.61 & 64.39 \\
\bottomrule
\end{tabular*}
\endgroup
\end{table}

\paragraph{Constraint location.}
Taking the MATH encoder as a healthy reference, we compare \relax \relax \relax $\mathcal D$-only
\relax SIGReg, a fixed variance/spectral barrier on $H$ alone, and their direct
sum with Flash encoder and MATH predictor. The $H$-only barrier raises
\relax \relax \relax $\operatorname{erank}(H)$ \relax to $27.90$, but $Z/\hat Z$ reach only $9.67/2.96$;
SIGReg$(Z)$ alone gives $4.65/7.65/7.74$. Directly stacking the two
constraints does not automatically repair \relax \relax \relax $H$ and $Z$ together\relax . Neither
\emph{expanding $H$ alone} nor \emph{matching the marginal distribution of
$Z$ alone} guarantees a usable readout.
\begin{table}[t]
\centering
\caption{Fixed-threshold $H$ barrier and endpoint SIGReg placement. Four-task
INTACT, seed 3072, \relax \relax \relax epoch 2 \relax fixed PushT probe; predictor MATH throughout. The
$H$ barrier penalizes mean standard deviation below $.10$ or leading
eigenvalue share above $.35$.}
\label{tab:app_collapse_sites}
\begingroup\small\setlength{\tabcolsep}{3pt}\renewcommand{\arraystretch}{1.12}
\begin{tabular*}{\linewidth}{@{\extracolsep{\fill}}llccc@{}}
\toprule
Encoder & Constraint & \relax \relax \relax $\operatorname{erank}(H)$ \relax & \relax \relax \relax $\operatorname{erank}(Z)$ \relax & \relax \relax \relax $\operatorname{erank}(\hat Z)$ \relax \\
\midrule
MATH & SIGReg$(Z)$ $.03$ & 21.13 & 50.18 & 49.14 \\
Flash & SIGReg$(Z)$ $.03$ & 4.65 & 7.65 & 7.74 \\
Flash & $H$ barrier $.01$, no $Z$ constraint & 27.90 & 9.67 & 2.96 \\
Flash & $H$ barrier $.01$ + SIGReg$(Z)$ $.03$ & 2.12 & 6.14 & 3.33 \\
\bottomrule
\end{tabular*}
\endgroup
\end{table}

\subsection{Numerical trigger versus representational bottleneck}

The component swap identifies the first perturbation in visual-encoder
Flash-SDPA backward. Shared four-task optimization, BF16, and a relatively
large learning rate can amplify this small difference into distinct
non-convex training trajectories. Predictor Flash alone remains bitwise
reproducible. Conversely, enabling deterministic algorithms while retaining
Flash still yields a low-rank phase in a single-seed run (mean four-task
\relax \relax \relax epoch 1 \relax rank $7.36$, versus $25.42$ with MATH). Short-run divergence and
long-run rank loss together connect a numerical trigger to \relax \relax \relax the loss of
sample variation in the representation\relax .

Once the encoder has erased distinctions between samples, a deterministic
downstream map cannot restore them from identical inputs. Moreover, an
endpoint gradient reaches $H$ through the projector Jacobian,
\[
\nabla_H\mathcal D(Z)=J_P(H)^\top\nabla_Z\mathcal D(Z).
\]
\relax \relax \relax A Demand Regularizer \relax may therefore produce a gradient at $Z$ without
sufficiently exciting or transmitting useful differences in $H$. Conversely,
expanding $H$ alone does not ensure a useful endpoint or predictive readout.

\subsection{Operational definition of collapse and degradation}

For a fixed probe $U\in\mathbb R^{N\times d_U}$, define the centered sample
covariance \relax \relax \relax $\Sigma_U=\bar{U}^\top\bar{U}/(N-1)$\relax . An exactly constant
representation has $\operatorname{rank}(\Sigma_U)=0$. Training more often
produces a concentrated spectrum, nearly vanishing scale, or deteriorated
predictive readout, rather than exact rank zero. We therefore report
effective rank, mean inter-sample cosine, scale, and prediction error.
\relax \relax \relax For the earlier exploratory layer-wise audit, we flagged the \relax 13 \relax \relax \relax runs whose
effective rank fell below half of \relax the healthy same-epoch reference. \relax This
screening is separate from the four-way training outcome used throughout the
paper: at the end of training (Ep.\,2), a training seed is \emph{collapsed} if
$f_H<0.05$, \emph{degraded} if $0.05\leq f_H<0.5$, \emph{over-regularized} if
$f_H\geq0.5$ but official SR falls below $30\%$ at Ep.\,1 or Ep.\,2, and
\emph{normal} otherwise (Section~\ref{sec:experiments}; per-seed values in
Table~\ref{tab:app_per_seed_outcomes}).

\relax 

\relax \relax \relax On the fixed probe, let the smaller normalized effective rank of $H$ and $Z$ be
\relax \[
\relax \relax \relax \overline{\operatorname{erank}}_{\min}\relax =\min\!\left\{
\relax \relax \relax \frac{\operatorname{erank}(H)}{\min(N-1,d_H)}\relax ,
\relax \relax \relax \frac{\operatorname{erank}(Z)}{\min(N-1,d_Z)}\relax \right\}.
\]
This is an external diagnostic, distinct from the training-time supply $k$.
In the formal paired ten-seed audit, each candidate is compared with a
same-seed, same-epoch MATH reference. We bootstrap trajectories 10,000
times and record a \emph{\relax \relax \relax paired rank-drop \relax event} only when the
upper endpoint of the one-sided 95\% confidence interval of
\relax \relax \relax $\log(\overline{\operatorname{erank}}_{\min}^{\rm cand}/\overline{\operatorname{erank}}_{\min}^{\rm ref})$ \relax is below zero.
\relax \relax \relax The event does not require exact rank zero or a loss of task success\relax .

\subsection{Matching \relax \relax demand \relax capacity \relax to \relax \relax supply \relax capacity\relax }

\textbf{The central remedy is not to force latents onto a fixed manifold,
but to ensure that \relax \relax excitation in \relax \relax \relax $H$ \relax and transmission \relax \relax \relax from $H$ to $Z$
\relax can \relax \relax \relax meet \relax the \relax \relax demand \relax capacity of the endpoint\relax .} The \relax \relax \relax Demand Regularizer
\relax $\mathcal D(Z)$ \relax \relax \relax enters the demand capacity \relax $q$. \relax \relax \relax $H$ \relax must both preserve \relax \relax \relax sample
differences ($k_{\mathrm{exc}}$\relax ) and transmit them to $Z$
(\relax \relax \relax $k_{\mathrm{tr}}$\relax ), giving \relax \relax \relax the supply capacity $k=\min(k_{\mathrm{exc}},k_{\mathrm{tr}})$\relax .
If $q>k$, the request remains unsupported; if $k\ge q$, an auxiliary \relax \relax \relax Supply Regularizer
\relax should not continue forcing $H$ to expand. Excess supply is not itself
noise: task readout determines which directions are relevant.

SADL implements this principle with a \relax \relax \relax $\mathcal S$ \relax active only at a deficit,
\[
\mathcal S
=\bigl[\log(q+\epsilon)-\log(k+\epsilon)\bigr]_+^2.
\]
Unlike \relax \relax \relax $\mathcal D$-only training, it acts on $H$\relax ; unlike
a fixed $H$-expansion term, it switches off when supply suffices. In an
independent paired ten-seed four-task LeWM audit, Flash+SIGReg has $5/10$
\relax \relax \relax paired rank-drop \relax events, versus $0/10$ after adding \relax \relax \relax SADL.
$\overline{\operatorname{erank}}_{\min}$ \relax rises from $.032$ to $.373$, prediction loss falls from
$1.8586$ to $.0173$, and \relax \relax \relax official \relax Pure-CEM macro success rises from
$31.92\%$ to $64.97\%$.
\begin{table}[t]
\centering
\caption{Paired ten-training-seed \relax \relax \relax epoch 1 \relax four-task LeWM audit. A \relax \relax \relax paired
rank-drop \relax event is defined relative to the same-seed MATH reference by the fixed-probe
bootstrap criterion above; it is not an exact rank-zero event. Success is
\relax \relax \relax official \relax Pure-CEM macro SR (\%).}
\label{tab:app_collapse_formal}
\begingroup\small\setlength{\tabcolsep}{3pt}\renewcommand{\arraystretch}{1.12}
\begin{tabular*}{\linewidth}{@{\extracolsep{\fill}}lcccc@{}}
\toprule
Training condition & \relax \relax \relax Paired rank-drop \relax events & \relax \relax \relax $\overline{\operatorname{erank}}_{\min}$ \relax & Prediction loss & Official SR \\
\midrule
MATH + SIGReg (reference) & -- & $.031$ & $.7767$ & $32.02$ \\
Flash, no \relax \relax \relax Demand Regularizer \relax & $10/10$ & $.005$ & $27.0896$ & $24.11$ \\
Flash + SIGReg & $5/10$ & $.032$ & $1.8586$ & $31.92$ \\
Flash + SIGReg + \relax \relax \relax exc-only \relax SADL & $0/10$ & $.370$ & $.0174$ & $66.67$ \\
Flash + SIGReg + \relax \relax SADL & $0/10$ & $.373$ & $.0173$ & $64.97$ \\
\bottomrule
\end{tabular*}
\endgroup
\end{table}

\relax \relax \relax \begin{table}[!htbp]
\centering\scriptsize
\caption{Per-seed Ep.\,2 $f_H$ values for the collapse diagnostic cohorts.
Seeds are listed in the order $0,1,123,21,3072,3407,42,7,77,8888$ for
LeWM and $0,3072,42$ for INTACT. The N/O/D/C count uses the Ep.\,2
four-way definition. For LeWM, the archived ten-seed audit stores cohort
SR aggregates; for INTACT, the paired Ep.\,1/Ep.\,2 SR values are listed
by training seed below, in the seed order stated above.}
\label{tab:app_per_seed_outcomes}
\setlength{\tabcolsep}{3pt}\renewcommand{\arraystretch}{1.12}
\begin{tabularx}{\linewidth}{@{}>{\raggedright\arraybackslash}p{.2\linewidth}>{\raggedright\arraybackslash}p{.27\linewidth}>{\raggedright\arraybackslash}Xc@{}}
\toprule
Setting & Ep.\,1/Ep.\,2 SR & Ep.\,2 $f_H$ by seed & N/O/D/C \\
\midrule
LeWM, MATH, $\mathcal D$-only & 32.02/48.85 (cohort) & .168, .157, .0933, .168, .220, .177, .126, .219, .0949, .118 & $0/0/10/0$ \\
LeWM, Flash, no $\mathcal D$ & 24.11/23.03 (cohort) & $2.29\!\times\!10^{-7}$, $4.76\!\times\!10^{-8}$, $1.39\!\times\!10^{-7}$, .00138, $6.42\!\times\!10^{-8}$, $1.33\!\times\!10^{-7}$, $7.22\!\times\!10^{-8}$, $1.30\!\times\!10^{-7}$, $7.85\!\times\!10^{-8}$, $3.91\!\times\!10^{-5}$ & $0/0/0/10$ \\
LeWM, Flash, $\mathcal D$-only & 31.92/39.96 (cohort) & .000239, .149, $7.98\!\times\!10^{-5}$, .000395, .0177, .139, .0283, .287, .0984, .0928 & $0/0/5/5$ \\
LeWM, Flash, +\textbf{SADL} & 64.97/69.20 (cohort) & .990, .990, .989, .990, .990, .986, .989, .990, .989, .989 & $10/0/0/0$ \\
INTACT, MATH, $\mathcal D$-only & 49.50/76.92; 39.17/67.17; 32.75/65.83 & .074, .0767, .114 & $0/0/3/0$ \\
INTACT, Flash, $\mathcal D$-only & 43.42/92.33; 48.50/79.00; 45.33/27.00 & .0532, .0714, .00193 & $0/0/2/1$ \\
INTACT, MATH, +\textbf{SADL} & 94.75/96.83; 94.58/95.75; 95.33/96.42 & .988, .985, .986 & $3/0/0/0$ \\
INTACT, Flash, +\textbf{SADL} & 95.58/96.33; 95.42/96.17; 95.25/96.42 & .987, .984, .986 & $3/0/0/0$ \\
\bottomrule
\end{tabularx}
\end{table}

\begin{table}[!htbp]
\centering
\caption{Timing of the drop below $50\%$ (or $25\%$) of its own peak in the deep encoder layers relative to $Z$, in the ten-seed LeWM audit with SIGReg (40 run--task pairs per condition, one log every 569 steps). The measured quantity is the normalized effective rank of each layer on the fixed probe. A layer has dropped at the first log where it falls below the threshold. Columns give the run--task pairs in which $Z$ dropped, the encoder timing relative to $Z$ in those pairs, and the pairs in which the encoder dropped while $Z$ never did (encoder only).}
\label{tab:app_collapse_timing}
\begingroup\small\setlength{\tabcolsep}{3pt}\renewcommand{\arraystretch}{1.12}
\begin{tabularx}{\linewidth}{@{}Xcccc@{}}
\toprule
Condition & Threshold & $Z$ dropped & Encoder earlier / same / later / from start & Encoder only \\
\midrule
SIGReg ($\mathcal D$-only), Flash & $50\%$ & 16 & 5 / 7 / 0 / 4 & 5 \\
SIGReg ($\mathcal D$-only), Flash & $25\%$ & 4 & 0 / 3 / 0 / 1 & 9 \\
SIGReg with MATH; exc-only; \textbf{SADL} & both & 0 & -- & 0 \\
\bottomrule
\end{tabularx}
\endgroup
\end{table}

For C1, the nominal effective ranks of $H/Z$ are $69.4/65.3$ when
computed from a covariance at the numerical floor. Since $f_H<10^{-6}$,
these values reflect numerical noise rather than sample diversity.

exc-only \relax also reaches $0/10$ in this training regime\relax \relax \relax ; interventions in which
transmission limits supply are \relax given in Section~\ref{sec:app_channel_roles}.
The paired audit shows that SADL reduces \relax \relax \relax paired rank-drop events \relax under
Flash from $5/10$ to $0/10$, retaining the faster and lower-memory fused
attention implementation rather than reverting to MATH
(\href{https://arxiv.org/abs/2205.14135}{FlashAttention}).
The \relax \relax \relax Supply Regularizer \relax adds no model parameters;
implementation runtime and memory are reported in Table~\ref{tab:app_training_cost}.
\textbf{SADL also accelerates representation formation and improves \relax \relax \relax epoch 1
$\overline{\operatorname{erank}}_{\min}$\relax , predictive quality, and control success.}

\subsection{SADL Repairs Near-Collapsed Checkpoints}
\label{sec:app_rescue}

\begingroup
\linespread{1.05}\selectfont
\setlength{\parskip}{0.2em}

We next ask whether SADL can repair \relax \relax \relax a representation that has already lost its
sample variation\relax , rather than only prevent \relax \relax \relax this loss \relax from the beginning of training. Starting from fixed
near-collapsed checkpoints, we resume training with either the original
\relax \relax \relax Demand Regularizer \relax or the same objective augmented by SADL. Each reported
control value averages four tasks and evaluation seeds $\{0,1,42\}$, with
100 episodes per task--seed cell. LeWM uses \relax \relax \relax official \relax Pure-CEM ($300\times30$,
$T_H=R_H=5$), whereas INTACT uses \relax \relax \relax official \relax Direct control ($T_H=R_H=5$). Within
each row, the checkpoint, continuation protocol, training host, and evaluation
seeds are matched; rows from different training hosts are not pooled.

\relax %
\relax \relax \begin{table}[!htbp]
\relax \centering
\caption{Post-collapse rescue and \relax \relax \relax Supply Regularizer-strength \relax sensitivity. Panel (a)
describes the two seed-42 source checkpoints before continuation using
task-averaged fixed probes; $r_H,r_Z$ are effective ranks, $c_H,c_Z$ are mean
off-diagonal cosines, and $\mathcal L_{\rm pred}$ is latent prediction loss.
Both states retain nonzero numerical rank and are therefore termed
\emph{near-collapsed}, not exact rank-zero. Panel (b) reports matched \relax \relax \relax official\relax macro success rates. Native denotes $\lambda_S=.01$. Exploratory doses are
reported to expose non-monotone rescue rather than to select a best result. \relax The Pro6000D and H200 labels identify the training hosts for the corresponding continuation rows.\relax }
\label{tab:app_rescue}
\begingroup
\scriptsize
\setlength{\tabcolsep}{3.4pt}
\renewcommand{\arraystretch}{1.10}
\textbf{(a) Pre-rescue checkpoint diagnostics}\\[3pt]
\begin{tabular*}{\textwidth}{@{\extracolsep{\fill}}llcccccl@{}}
\toprule
Family & Seed / state & $r_H$ & $c_H$ & $r_Z$ & $c_Z$ &
$\mathcal L_{\rm pred}$ & Diagnosis \\
\midrule
LeWM & 42 / C2 \relax \relax \relax Ep.\,2 \relax & 15.01 & .972 & 33.65 & .128 & .346 &
low-rank, highly aligned $H$ \\
INTACT & 42 / \relax \relax \relax Ep.\,1 \relax & 18.14 & .945 & 23.62 & .599 & 1.126 &
low-rank \relax \relax \relax $H$ \relax and poor readout \\
\bottomrule
\end{tabular*}

\vspace{8pt}
\textbf{(b) Matched continuation and dose response}\\[3pt]
\begin{tabular*}{\textwidth}{@{\extracolsep{\fill}}lllrrrrl@{}}
\toprule
Family & Train seed / start & Continuation group & $\lambda_S$ &
Control SR & SADL SR & $\Delta$ (pp) & Outcome \\
\midrule
LeWM & 21 / \relax \relax \relax Ep.\,2\relax $\to$\relax \relax \relax Ep.\,4 \relax & matched native & .01 & 25.250 & 46.917 & +21.667 & clear rescue \\
LeWM & 7 / \relax \relax \relax Ep.\,2\relax $\to$\relax \relax \relax Ep.\,4 \relax & matched native & .01 & 61.000 & 72.167 & +11.167 & clear rescue \\
LeWM & 3072 / \relax \relax \relax Ep.\,3\relax $\to$\relax \relax \relax Ep.\,5 \relax & matched native & .01 & 21.500 & 23.083 & +1.583 & small positive \\
LeWM & 42 / \relax \relax \relax Ep.\,2\relax $\to$\relax \relax \relax Ep.\,4 \relax & matched native & .01 & 24.167 & 22.500 & -1.667 & not rescued \\
\cmidrule(lr){1-8}
LeWM & 42 / \relax \relax \relax Ep.\,2\relax $\to$\relax \relax \relax Ep.\,4 \relax & same source & .03 & 24.167 & 28.333 & +4.167 & rescued \\
LeWM & 42 / \relax \relax \relax Ep.\,2\relax $\to$\relax \relax \relax Ep.\,4 \relax & post-hoc dose & .05 & 24.167 & 23.167 & -1.000 & not rescued \\
LeWM & 42 / \relax \relax \relax Ep.\,2\relax $\to$\relax \relax \relax Ep.\,4 \relax & post-hoc dose & .07 & 24.167 & 25.667 & +1.500 & marginal \\
LeWM & 42 / \relax \relax \relax Ep.\,2\relax $\to$\relax \relax \relax Ep.\,4 \relax & same source & .10 & 24.167 & 29.083 & +4.917 & rescued \\
LeWM & 42 / \relax \relax \relax Ep.\,2\relax $\to$\relax \relax \relax Ep.\,4 \relax & same source & .15 & 24.167 & 21.917 & -2.250 & not rescued \\
LeWM & 42 / \relax \relax \relax Ep.\,2\relax $\to$\relax \relax \relax Ep.\,4 \relax & same source & .20 & 24.167 & 24.167 & +0.000 & neutral \\
LeWM & 42 / \relax \relax \relax Ep.\,2\relax $\to$\relax \relax \relax Ep.\,4 \relax & hardware replicate & .30 & 24.167 & 22.250 & -1.917 & not rescued \\
LeWM & 42 / \relax \relax \relax Ep.\,2\relax $\to$\relax \relax \relax Ep.\,4 \relax & hardware replicate & 1.00 & 24.167 & 20.250 & -3.917 & not rescued \\
\midrule
INTACT & 0 / \relax \relax \relax Ep.\,1\relax $\to$\relax \relax \relax Ep.\,2 \relax & matched native & .01 & 28.917 & 26.500 & -2.417 & not rescued \\
INTACT & 3072 / \relax \relax \relax Ep.\,1\relax $\to$\relax \relax \relax Ep.\,2 \relax & matched native & .01 & 24.417 & 53.917 & +29.500 & clear rescue \\
INTACT & 42 / \relax \relax \relax Ep.\,1\relax $\to$\relax \relax \relax Ep.\,2 \relax & Pro6000D native & .01 & 18.917 & 68.333 & +49.417 & clear rescue \\
\cmidrule(lr){1-8}
INTACT & 42 / \relax \relax \relax Ep.\,1\relax $\to$\relax \relax \relax Ep.\,2 \relax & Pro6000D matched & .03 & 18.917 & 39.500 & +20.583 & weakened \\
INTACT & 42 / \relax \relax \relax Ep.\,1\relax $\to$\relax \relax \relax Ep.\,2 \relax & Pro6000D matched & .10 & 18.917 & 29.750 & +10.833 & weakened \\
INTACT & 42 / \relax \relax \relax Ep.\,1\relax $\to$\relax \relax \relax Ep.\,2 \relax & Pro6000D matched & .15 & 18.917 & 27.000 & +8.083 & weakened \\
INTACT & 42 / \relax \relax \relax Ep.\,1\relax $\to$\relax \relax \relax Ep.\,2 \relax & Pro6000D matched & .20 & 18.917 & 27.667 & +8.750 & weakened \\
INTACT & 42 / \relax \relax \relax Ep.\,1\relax $\to$\relax \relax \relax Ep.\,2 \relax & H200 replicate & .01 & 23.083 & 69.083 & +46.000 & clear rescue \\
INTACT & 42 / \relax \relax \relax Ep.\,1\relax $\to$\relax \relax \relax Ep.\,2 \relax & H200 replicate & .03 & 23.083 & 55.833 & +32.750 & weakened \\
INTACT & 42 / \relax \relax \relax Ep.\,1\relax $\to$\relax \relax \relax Ep.\,2 \relax & H200 replicate & .10 & 23.083 & 28.917 & +5.834 & weakened \\
INTACT & 42 / \relax \relax \relax Ep.\,1\relax $\to$\relax \relax \relax Ep.\,2 \relax & H200 replicate & .15 & 23.083 & 74.583 & +51.500 & rebound \\
INTACT & 42 / \relax \relax \relax Ep.\,1\relax $\to$\relax \relax \relax Ep.\,2 \relax & H200 replicate & .20 & 23.083 & 27.583 & +4.500 & weakened \\
\bottomrule
\end{tabular*}
\endgroup
\relax %
\relax \relax \end{table}

\relax 

The native \relax \relax \relax Supply Regularizer \relax rescues three of four LeWM checkpoints only weakly or
strongly, but fails on seed 42. For that same LeWM state, moderate increases
to $.03$ or $.10$ make the macro improvement positive, whereas larger or
intermediate doses can again fail. INTACT is similarly checkpoint-dependent:
seed 0 is not robustly rescued, while seeds 3072 and 42 improve substantially.
For seed 42, the Pro6000D matched group is strongest at the native weight and
every larger tested dose weakens recovery. The H200 replication confirms a
non-monotone response---$.15$ rebounds above native, while $.10$ and $.20$
largely erase the gain---so absolute scores are not combined across hosts.

\textbf{SADL therefore has genuine post-collapse repair capability, but
recovery depends on the checkpoint state and \relax \relax \relax Supply Regularizer \relax strength; increasing
$\lambda_S$ is not monotonically beneficial.} This is more informative than
a binary ``can rescue'' claim: a deficit-triggered correction can reopen a
damaged representation\relax \relax , yet the intervention must remain commensurate
with the \relax \relax \relax attainable supply \relax capacity.

\endgroup

\relax %
\relax \section{Additional Control Results}
\label{app:ahead_control}

We separate \relax \relax \relax official \relax from CLEAR-LeWM v0.8 so that success rates are compared only within the same evaluation protocol. The tables report single-task and jointly trained four-task results for LeWM and INTACT under SIGReg, VISReg, and QQWorld. Each task value is the mean $\pm$ sample SD across three training seeds after averaging the complete three-evaluation-seed triplet; the macro is first averaged over the four tasks within each training seed and then summarized across training seeds. \relax \relax \relax Epoch 1 and epoch 5 \relax are equal-budget comparisons, while \relax \relax \relax $\mathcal D$-only epoch 10 \relax is a longer-budget reference. The additional tables here retain the original backend protocol: \relax \relax \relax $\mathcal D$-only \relax uses a MATH encoder, +SADL uses a Flash encoder, and both use MATH predictors. The joint \relax \relax \relax epoch 1 \relax Flash-matched comparison appears in Table~\ref{tab:joint}\relax \relax .

\relax \subsection{Joint four-task control and fixed probes}
Joint four-task training tests one shared encoder and projector across all tasks. Table~\ref{tab:joint_lowdiv} pairs six low-diversity Flash epoch 1 runs with same-seed +\textbf{SADL} runs, and Table~\ref{tab:joint} reports the three-seed comparison. Across the six Flash-matched model--Demand Regularizer pairs in Table~\ref{tab:joint}, \textbf{SADL} raises epoch 1 official SR by $23.9$--$67.4$ percentage points: LeWM reaches $63.0$--$71.6\%$ and INTACT $95.4$--$95.6\%$. Effective rank at $H/Z$ rises $3.4$--$6.4$/$1.6$--$5.5$ times, mean sample cosine falls $57$--$71$/$41$--$93$ times, and prediction loss falls $20$--$303$ times. \textbf{Both single-task and joint four-task training yield more useful control policies and more discriminative representations after only one epoch.} Per-task results are in Appendix~\ref{app:ahead_control}.

\begin{table}[H]
\centering
\caption{Six matched single-seed low-diversity Flash examples of joint four-task training at epoch 1, with same-seed +\textbf{SADL} runs. $\mathrm{SR}$/$\mathrm{SR}_{\rm SADL}$ are without/with \textbf{SADL}; arrows show the corresponding probe changes. The three-seed comparison is in Table~\ref{tab:joint}.}
\label{tab:joint_lowdiv}
\scriptsize
\setlength{\tabcolsep}{0.9pt}
\renewcommand{\arraystretch}{0.88}
\begin{tabular*}{\linewidth}{@{\extracolsep{\fill}}llccr@{${}\to{}$}lr@{${}\to{}$}lr@{${}\to{}$}lr@{${}\to{}$}lr@{${}\to{}$}l@{}}
\toprule
Model & $\mathcal D$ & $\mathrm{SR}$ & $\mathrm{SR}_{\rm SADL}$ & \multicolumn{2}{c}{$\mathrm{erank}_H$} & \multicolumn{2}{c}{$\mathrm{cos}_H$} & \multicolumn{2}{c}{$\mathrm{erank}_Z$} & \multicolumn{2}{c}{$\mathrm{cos}_Z$} & \multicolumn{2}{c}{$L_{\rm pred}$} \\
\midrule
LeWM & SIGReg s42 & $31.3$ & $\mathbf{61.8}$ & $17.4$ & $\mathbf{54.0}$ & $.926$ & $\mathbf{.012}$ & $23.5$ & $\mathbf{50.4}$ & $.601$ & $\mathbf{.007}$ & $.912$ & $\mathbf{.017}$ \\
 & SIGReg s3072 & $46.7$ & $\mathbf{62.9}$ & $12.3$ & $\mathbf{53.8}$ & $.960$ & $\mathbf{.014}$ & $24.9$ & $\mathbf{51.1}$ & $.103$ & $\mathbf{.008}$ & $.301$ & $\mathbf{.018}$ \\
 & QQWorld s3072 & $23.9$ & $\mathbf{64.0}$ & $14.1$ & $\mathbf{55.2}$ & $.992$ & $\mathbf{.014}$ & $13.0$ & $\mathbf{53.5}$ & $.864$ & $\mathbf{.009}$ & $3.752$ & $\mathbf{.018}$ \\
\midrule
INTACT & SIGReg s0 & $26.9$ & $\mathbf{95.8}$ & $5.4$ & $\mathbf{53.5}$ & $.995$ & $\mathbf{.013}$ & $9.4$ & $\mathbf{53.3}$ & $.795$ & $\mathbf{.007}$ & $4.430$ & $\mathbf{.017}$ \\
 & SIGReg s3072 & $22.5$ & $\mathbf{95.6}$ & $8.1$ & $\mathbf{52.8}$ & $.991$ & $\mathbf{.015}$ & $7.2$ & $\mathbf{53.4}$ & $.885$ & $\mathbf{.009}$ & $10.041$ & $\mathbf{.018}$ \\
 & QQWorld s3072 & $24.2$ & $\mathbf{96.0}$ & $4.6$ & $\mathbf{53.7}$ & $1.000$ & $\mathbf{.015}$ & $7.9$ & $\mathbf{54.9}$ & $.833$ & $\mathbf{.009}$ & $4.495$ & $\mathbf{.019}$ \\
\bottomrule
\end{tabular*}
\end{table}

\relax \subsection{Official}
\begin{table}[H]
\centering
\caption{Single-task LeWM \relax \relax \relax official \relax success (\%). All rows are evaluated after one epoch. LeWM uses Pure-CEM with $T_H=R_H=5$. \relax \relax \relax $\mathcal D$-only \relax uses a MATH encoder; +SADL uses a Flash encoder; both predictors use MATH. Values are mean $\pm$ sample SD over three training seeds.}
\label{tab:app_e_official_single_lewm}
\begingroup\relax %
\relax \relax \scriptsize\relax \setlength{\tabcolsep}{2pt}\relax %
\relax \relax \renewcommand{\arraystretch}{0.88}
\relax \begin{tabularx}{\linewidth}{@{}llcYYYYY@{}}
\toprule
\relax \relax \relax $\mathcal D$ \relax & Method & Epoch & PushT & Cube & Reacher & TwoRoom & Macro \\
\midrule
SIGReg & \relax \relax \relax $\mathcal D$-only \relax & \relax \relax \relax Ep.\,1 \relax & $67.00\pm13.02$ & $61.78\pm1.64$ & $26.44\pm12.70$ & $25.78\pm8.15$ & $45.25\pm6.50$ \\
SIGReg & +SADL & \relax \relax \relax Ep.\,1 \relax & $65.22\pm3.86$ & $58.89\pm1.84$ & $79.44\pm1.71$ & $85.22\pm4.82$ & $\boldsymbol{72.19\pm0.97}$ \\
\midrule
VISReg & \relax \relax \relax $\mathcal D$-only \relax & \relax \relax \relax Ep.\,1 \relax & $74.11\pm4.62$ & $63.22\pm6.45$ & $32.56\pm2.71$ & $52.78\pm23.40$ & $55.67\pm5.46$ \\
VISReg & +SADL & \relax \relax \relax Ep.\,1 \relax & $76.22\pm2.17$ & $58.22\pm0.51$ & $80.89\pm2.04$ & $89.89\pm2.55$ & $\boldsymbol{76.31\pm1.23}$ \\
\midrule
QQWorld & \relax \relax \relax $\mathcal D$-only \relax & \relax \relax \relax Ep.\,1 \relax & $72.22\pm5.32$ & $64.56\pm4.86$ & $34.33\pm16.50$ & $33.67\pm5.36$ & $51.19\pm6.98$ \\
QQWorld & +SADL & \relax \relax \relax Ep.\,1 \relax & $71.78\pm2.17$ & $61.00\pm4.63$ & $81.44\pm1.17$ & $86.89\pm1.26$ & $\boldsymbol{75.28\pm1.73}$ \\
\bottomrule
\end{tabularx}
\endgroup
\end{table}

\begin{table}[H]
\centering
\caption{Single-task INTACT \relax \relax \relax official \relax success (\%). All rows are evaluated after one epoch. INTACT uses search-free Direct control. \relax \relax \relax $\mathcal D$-only \relax uses a MATH encoder; +SADL uses a Flash encoder; both predictors use MATH. Values are mean $\pm$ sample SD over three training seeds.}
\label{tab:app_e_official_single_intact}
\begingroup\relax %
\relax \relax \scriptsize\relax \setlength{\tabcolsep}{2pt}\relax %
\relax \relax \renewcommand{\arraystretch}{0.88}
\relax \begin{tabularx}{\linewidth}{@{}llcYYYYY@{}}
\toprule
\relax \relax \relax $\mathcal D$ \relax & Method & Epoch & PushT & Cube & Reacher & TwoRoom & Macro \\
\midrule
SIGReg & \relax \relax \relax $\mathcal D$-only \relax & \relax \relax \relax Ep.\,1 \relax & $80.89\pm4.35$ & $97.67\pm2.03$ & $28.78\pm22.52$ & $27.22\pm8.98$ & $58.64\pm8.14$ \\
SIGReg & +SADL & \relax \relax \relax Ep.\,1 \relax & $85.67\pm1.45$ & $100.00\pm0.00$ & $96.89\pm2.14$ & $98.78\pm1.07$ & $\boldsymbol{95.33\pm0.44}$ \\
\midrule
VISReg & \relax \relax \relax $\mathcal D$-only \relax & \relax \relax \relax Ep.\,1 \relax & $84.22\pm2.50$ & $97.67\pm1.15$ & $51.89\pm40.50$ & $54.11\pm14.99$ & $71.97\pm12.79$ \\
VISReg & +SADL & \relax \relax \relax Ep.\,1 \relax & $86.89\pm0.19$ & $100.00\pm0.00$ & $97.00\pm0.58$ & $99.11\pm0.19$ & $\boldsymbol{95.75\pm0.08}$ \\
\midrule
QQWorld & \relax \relax \relax $\mathcal D$-only \relax & \relax \relax \relax Ep.\,1 \relax & $82.67\pm1.20$ & $97.22\pm3.10$ & $37.33\pm30.55$ & $29.00\pm6.17$ & $61.56\pm8.61$ \\
QQWorld & +SADL & \relax \relax \relax Ep.\,1 \relax & $87.11\pm1.39$ & $100.00\pm0.00$ & $95.22\pm1.58$ & $99.56\pm0.19$ & $\boldsymbol{95.47\pm0.21}$ \\
\bottomrule
\end{tabularx}
\endgroup
\end{table}

\begin{table}[H]
\centering
\caption{Joint four-task LeWM \relax \relax \relax official \relax success (\%). \relax \relax \relax Epoch 1 \relax and \relax \relax \relax epoch 5 \relax compare equal training budgets; \relax \relax \relax $\mathcal D$-only epoch 10 \relax is a longer-budget reference. LeWM uses Pure-CEM with $T_H=R_H=5$. \relax \relax \relax $\mathcal D$-only \relax uses a MATH encoder; +SADL uses a Flash encoder; both predictors use MATH. Values are mean $\pm$ sample SD over three training seeds.}
\label{tab:app_e_official_joint_lewm}
\begingroup\relax %
\relax \relax \scriptsize\relax \setlength{\tabcolsep}{2pt}\relax %
\relax \relax \renewcommand{\arraystretch}{0.88}
\relax \begin{tabularx}{\linewidth}{@{}llcYYYYY@{}}
\toprule
\relax \relax \relax $\mathcal D$ \relax & Method & Epoch & PushT & Cube & Reacher & TwoRoom & Macro \\
\midrule
SIGReg & \relax \relax \relax $\mathcal D$-only \relax & \relax \relax \relax Ep.\,1 \relax & $13.00\pm1.86$ & $53.11\pm2.22$ & $26.22\pm4.11$ & $35.67\pm7.42$ & $32.00\pm2.52$ \\
SIGReg & +SADL & \relax \relax \relax Ep.\,1 \relax & $28.78\pm5.32$ & $70.22\pm3.02$ & $79.89\pm0.77$ & $73.00\pm1.45$ & $\boldsymbol{62.97\pm1.25}$ \\
SIGReg & \relax \relax \relax $\mathcal D$-only \relax & \relax \relax \relax Ep.\,5 \relax & $75.67\pm3.71$ & $67.89\pm2.22$ & $82.00\pm1.53$ & $35.11\pm14.42$ & $65.17\pm2.21$ \\
SIGReg & +SADL & \relax \relax \relax Ep.\,5 \relax & $67.00\pm1.76$ & $67.56\pm2.50$ & $83.33\pm1.67$ & $64.67\pm1.76$ & $\boldsymbol{70.64\pm0.77}$ \\
SIGReg & \relax \relax \relax $\mathcal D$-only \relax & \relax \relax \relax Ep.\,10 \relax & $78.78\pm3.02$ & $67.44\pm1.95$ & $83.11\pm1.71$ & $44.78\pm8.19$ & $68.53\pm0.75$ \\
\midrule
VISReg & \relax \relax \relax $\mathcal D$-only \relax & \relax \relax \relax Ep.\,1 \relax & $53.00\pm4.84$ & $53.67\pm0.33$ & $31.56\pm8.69$ & $55.44\pm12.62$ & $48.42\pm3.76$ \\
VISReg & +SADL & \relax \relax \relax Ep.\,1 \relax & $53.56\pm11.64$ & $66.44\pm2.41$ & $78.44\pm0.84$ & $88.11\pm1.07$ & $\boldsymbol{71.64\pm2.96}$ \\
VISReg & \relax \relax \relax $\mathcal D$-only \relax & \relax \relax \relax Ep.\,5 \relax & $83.78\pm1.39$ & $60.00\pm1.20$ & $83.56\pm3.10$ & $61.44\pm7.20$ & $72.19\pm1.89$ \\
VISReg & +SADL & \relax \relax \relax Ep.\,5 \relax & $81.44\pm2.01$ & $66.11\pm2.55$ & $84.56\pm2.27$ & $82.22\pm1.64$ & $\boldsymbol{78.58\pm0.98}$ \\
VISReg & \relax \relax \relax $\mathcal D$-only \relax & \relax \relax \relax Ep.\,10 \relax & $86.33\pm2.08$ & $58.78\pm1.26$ & $85.89\pm0.51$ & $65.00\pm4.33$ & $74.00\pm1.08$ \\
\midrule
QQWorld & \relax \relax \relax $\mathcal D$-only \relax & \relax \relax \relax Ep.\,1 \relax & $16.67\pm5.13$ & $55.33\pm4.10$ & $28.44\pm3.79$ & $41.78\pm5.85$ & $35.56\pm3.97$ \\
QQWorld & +SADL & \relax \relax \relax Ep.\,1 \relax & $45.67\pm11.15$ & $67.44\pm2.87$ & $78.00\pm3.53$ & $82.56\pm0.69$ & $\boldsymbol{68.42\pm4.19}$ \\
QQWorld & \relax \relax \relax $\mathcal D$-only \relax & \relax \relax \relax Ep.\,5 \relax & $80.89\pm3.86$ & $67.11\pm4.50$ & $85.00\pm3.53$ & $52.22\pm12.29$ & $71.31\pm4.18$ \\
QQWorld & +SADL & \relax \relax \relax Ep.\,5 \relax & $76.89\pm2.46$ & $68.00\pm0.33$ & $83.67\pm0.88$ & $71.89\pm2.36$ & $\boldsymbol{75.11\pm1.34}$ \\
QQWorld & \relax \relax \relax $\mathcal D$-only \relax & \relax \relax \relax Ep.\,10 \relax & $86.11\pm1.17$ & $65.78\pm2.50$ & $82.78\pm1.07$ & $56.00\pm4.51$ & $72.67\pm0.65$ \\
\bottomrule
\end{tabularx}
\endgroup
\end{table}

\begin{table}[H]
\centering
\caption{Joint four-task INTACT \relax \relax \relax official \relax success (\%). \relax \relax \relax Epoch 1 \relax and \relax \relax \relax epoch 5 \relax compare equal training budgets; \relax \relax \relax $\mathcal D$-only epoch 10 \relax is a longer-budget reference. INTACT uses search-free Direct control. \relax \relax \relax $\mathcal D$-only \relax uses a MATH encoder; +SADL uses a Flash encoder; both predictors use MATH. Values are mean $\pm$ sample SD over three training seeds.}
\label{tab:app_e_official_joint_intact}
\begingroup\relax %
\relax \relax \scriptsize\relax \setlength{\tabcolsep}{2pt}\relax %
\relax \relax \renewcommand{\arraystretch}{0.88}
\relax \begin{tabularx}{\linewidth}{@{}llcYYYYY@{}}
\toprule
\relax \relax \relax $\mathcal D$ \relax & Method & Epoch & PushT & Cube & Reacher & TwoRoom & Macro \\
\midrule
SIGReg & \relax \relax \relax $\mathcal D$-only \relax & \relax \relax \relax Ep.\,1 \relax & $37.67\pm9.29$ & $71.67\pm6.51$ & $13.89\pm4.48$ & $55.00\pm8.99$ & $44.56\pm2.80$ \\
SIGReg & +SADL & \relax \relax \relax Ep.\,1 \relax & $85.22\pm0.77$ & $100.00\pm0.00$ & $97.11\pm0.38$ & $100.00\pm0.00$ & $\boldsymbol{95.58\pm0.17}$ \\
SIGReg & \relax \relax \relax $\mathcal D$-only \relax & \relax \relax \relax Ep.\,5 \relax & $85.78\pm3.83$ & $99.89\pm0.19$ & $95.33\pm0.58$ & $66.00\pm16.29$ & $86.75\pm3.20$ \\
SIGReg & +SADL & \relax \relax \relax Ep.\,5 \relax & $90.78\pm1.68$ & $100.00\pm0.00$ & $98.22\pm0.51$ & $99.78\pm0.19$ & $\boldsymbol{97.19\pm0.51}$ \\
SIGReg & \relax \relax \relax $\mathcal D$-only \relax & \relax \relax \relax Ep.\,10 \relax & $89.44\pm3.79$ & $100.00\pm0.00$ & $96.00\pm0.58$ & $80.89\pm11.19$ & $91.58\pm2.01$ \\
\midrule
VISReg & \relax \relax \relax $\mathcal D$-only \relax & \relax \relax \relax Ep.\,1 \relax & $54.67\pm3.28$ & $81.33\pm9.85$ & $34.78\pm11.65$ & $64.11\pm6.35$ & $58.72\pm4.79$ \\
VISReg & +SADL & \relax \relax \relax Ep.\,1 \relax & $83.78\pm2.04$ & $100.00\pm0.00$ & $97.78\pm1.17$ & $100.00\pm0.00$ & $\boldsymbol{95.39\pm0.65}$ \\
VISReg & \relax \relax \relax $\mathcal D$-only \relax & \relax \relax \relax Ep.\,5 \relax & $89.67\pm1.76$ & $99.33\pm0.33$ & $97.67\pm0.88$ & $76.44\pm1.95$ & $90.78\pm0.38$ \\
VISReg & +SADL & \relax \relax \relax Ep.\,5 \relax & $90.89\pm1.50$ & $100.00\pm0.00$ & $99.33\pm0.33$ & $99.89\pm0.19$ & $\boldsymbol{97.53\pm0.25}$ \\
VISReg & \relax \relax \relax $\mathcal D$-only \relax & \relax \relax \relax Ep.\,10 \relax & $91.00\pm0.58$ & $100.00\pm0.00$ & $97.11\pm0.19$ & $88.56\pm5.98$ & $94.17\pm1.45$ \\
\midrule
QQWorld & \relax \relax \relax $\mathcal D$-only \relax & \relax \relax \relax Ep.\,1 \relax & $36.89\pm11.65$ & $74.22\pm2.50$ & $21.00\pm6.77$ & $50.78\pm9.50$ & $45.72\pm4.33$ \\
QQWorld & +SADL & \relax \relax \relax Ep.\,1 \relax & $86.00\pm0.88$ & $100.00\pm0.00$ & $96.56\pm0.51$ & $100.00\pm0.00$ & $\boldsymbol{95.64\pm0.32}$ \\
QQWorld & \relax \relax \relax $\mathcal D$-only \relax & \relax \relax \relax Ep.\,5 \relax & $89.11\pm1.68$ & $99.89\pm0.19$ & $97.44\pm0.84$ & $82.56\pm14.20$ & $92.25\pm3.20$ \\
QQWorld & +SADL & \relax \relax \relax Ep.\,5 \relax & $91.67\pm0.88$ & $100.00\pm0.00$ & $98.56\pm0.19$ & $100.00\pm0.00$ & $\boldsymbol{97.56\pm0.17}$ \\
QQWorld & \relax \relax \relax $\mathcal D$-only \relax & \relax \relax \relax Ep.\,10 \relax & $89.56\pm2.22$ & $100.00\pm0.00$ & $97.22\pm1.54$ & $87.11\pm16.28$ & $93.47\pm4.69$ \\
\bottomrule
\end{tabularx}
\endgroup
\end{table}

\relax \FloatBarrier
\relax \subsection{CLEAR-LeWM v0.8}
\subsubsection{Single-task Moderate and Strict}
\begin{table}[H]
\centering
\caption{Single-task LeWM CLEAR-LeWM v0.8 success (\%) at \relax \relax \relax epoch 1 \relax under Moderate and Strict criteria. LeWM uses Pure-CEM with $T_H=R_H=5$; entries are mean $\pm$ sample SD over three training seeds.}
\label{tab:app_d_single_lewm}
\label{tab:app_e_strict_single_lewm}
\begingroup\scriptsize\setlength{\tabcolsep}{1.1pt}\relax %
\relax \relax \renewcommand{\arraystretch}{0.88}
\relax \begin{tabularx}{\linewidth}{@{}lllcYYYYY@{}}
\toprule
Criterion & \relax \relax \relax $\mathcal D$ \relax & Method & Epoch & PushT & Cube & Reacher & TwoRoom & Macro \\
\midrule
Moderate & SIGReg & \relax \relax \relax $\mathcal D$-only \relax & \relax \relax \relax Ep.\,1 \relax & $70.22\pm15.32$ & $41.33\pm3.84$ & $14.56\pm7.90$ & $10.44\pm5.80$ & $34.14\pm6.55$ \\
 & SIGReg & +SADL & \relax \relax \relax Ep.\,1 \relax & $67.89\pm4.22$ & $38.78\pm0.51$ & $76.89\pm6.36$ & $81.67\pm14.62$ & $\boldsymbol{66.31\pm2.38}$ \\
 & VISReg & \relax \relax \relax $\mathcal D$-only \relax & \relax \relax \relax Ep.\,1 \relax & $78.67\pm3.84$ & $43.33\pm7.26$ & $18.78\pm2.87$ & $45.67\pm32.30$ & $46.61\pm7.94$ \\
 & VISReg & +SADL & \relax \relax \relax Ep.\,1 \relax & $77.11\pm4.19$ & $37.67\pm1.20$ & $77.78\pm1.54$ & $83.00\pm5.20$ & $\boldsymbol{68.89\pm0.61}$ \\
 & QQWorld & \relax \relax \relax $\mathcal D$-only \relax & \relax \relax \relax Ep.\,1 \relax & $74.33\pm3.76$ & $42.33\pm3.93$ & $27.44\pm16.88$ & $18.33\pm13.17$ & $40.61\pm8.59$ \\
 & QQWorld & +SADL & \relax \relax \relax Ep.\,1 \relax & $71.44\pm3.24$ & $42.33\pm7.23$ & $75.78\pm0.69$ & $79.67\pm3.51$ & $\boldsymbol{67.31\pm2.56}$ \\
\midrule
Strict & SIGReg & \relax \relax \relax $\mathcal D$-only \relax & \relax \relax \relax Ep.\,1 \relax & $50.67\pm12.91$ & $18.44\pm2.01$ & $18.00\pm7.88$ & $3.89\pm2.36$ & $22.75\pm5.40$ \\
 & SIGReg & +SADL & \relax \relax \relax Ep.\,1 \relax & $51.22\pm0.19$ & $19.00\pm0.58$ & $82.33\pm1.15$ & $73.11\pm9.90$ & $\boldsymbol{56.42\pm2.43}$ \\
 & VISReg & \relax \relax \relax $\mathcal D$-only \relax & \relax \relax \relax Ep.\,1 \relax & $55.11\pm3.02$ & $22.22\pm5.01$ & $26.22\pm1.90$ & $30.56\pm26.34$ & $33.53\pm6.44$ \\
 & VISReg & +SADL & \relax \relax \relax Ep.\,1 \relax & $57.78\pm2.71$ & $17.67\pm2.73$ & $80.11\pm2.87$ & $73.78\pm2.91$ & $\boldsymbol{57.33\pm1.01}$ \\
 & QQWorld & \relax \relax \relax $\mathcal D$-only \relax & \relax \relax \relax Ep.\,1 \relax & $56.22\pm6.05$ & $19.33\pm6.08$ & $36.78\pm20.35$ & $11.22\pm12.28$ & $30.89\pm9.64$ \\
 & QQWorld & +SADL & \relax \relax \relax Ep.\,1 \relax & $58.44\pm1.02$ & $22.67\pm4.36$ & $78.78\pm3.89$ & $71.22\pm3.66$ & $\boldsymbol{57.78\pm3.08}$ \\
\bottomrule
\end{tabularx}
\endgroup
\end{table}

\begin{table}[H]
\centering
\caption{Single-task INTACT CLEAR-LeWM v0.8 success (\%) at \relax \relax \relax epoch 1 \relax under Moderate and Strict criteria with search-free Direct control; entries are mean $\pm$ sample SD over three training seeds.}
\label{tab:app_d_single_intact}
\label{tab:app_e_strict_single_intact}
\begingroup\scriptsize\setlength{\tabcolsep}{1.1pt}\relax %
\relax \relax \renewcommand{\arraystretch}{0.88}
\relax \begin{tabularx}{\linewidth}{@{}lllcYYYYY@{}}
\toprule
Criterion & \relax \relax \relax $\mathcal D$ \relax & Method & Epoch & PushT & Cube & Reacher & TwoRoom & Macro \\
\midrule
Moderate & SIGReg & \relax \relax \relax $\mathcal D$-only \relax & \relax \relax \relax Ep.\,1 \relax & $80.89\pm6.95$ & $95.00\pm5.03$ & $25.11\pm20.97$ & $10.78\pm4.67$ & $52.94\pm7.19$ \\
 & SIGReg & +SADL & \relax \relax \relax Ep.\,1 \relax & $86.67\pm2.91$ & $100.00\pm0.00$ & $97.00\pm0.88$ & $97.22\pm2.36$ & $\boldsymbol{95.22\pm1.02}$ \\
 & VISReg & \relax \relax \relax $\mathcal D$-only \relax & \relax \relax \relax Ep.\,1 \relax & $86.56\pm3.67$ & $96.22\pm2.83$ & $47.00\pm39.40$ & $37.00\pm41.01$ & $66.69\pm19.72$ \\
 & VISReg & +SADL & \relax \relax \relax Ep.\,1 \relax & $87.56\pm2.01$ & $100.00\pm0.00$ & $97.22\pm1.02$ & $98.00\pm1.20$ & $\boldsymbol{95.69\pm0.97}$ \\
 & QQWorld & \relax \relax \relax $\mathcal D$-only \relax & \relax \relax \relax Ep.\,1 \relax & $85.00\pm2.33$ & $96.67\pm4.36$ & $35.78\pm31.62$ & $9.11\pm5.32$ & $56.64\pm6.16$ \\
 & QQWorld & +SADL & \relax \relax \relax Ep.\,1 \relax & $89.00\pm0.58$ & $100.00\pm0.00$ & $97.11\pm1.50$ & $98.89\pm1.35$ & $\boldsymbol{96.25\pm0.55}$ \\
\midrule
Strict & SIGReg & \relax \relax \relax $\mathcal D$-only \relax & \relax \relax \relax Ep.\,1 \relax & $70.67\pm3.93$ & $80.78\pm6.38$ & $40.22\pm29.92$ & $3.00\pm1.73$ & $48.67\pm7.91$ \\
 & SIGReg & +SADL & \relax \relax \relax Ep.\,1 \relax & $76.00\pm2.67$ & $86.33\pm0.67$ & $98.67\pm0.88$ & $92.89\pm1.95$ & $\boldsymbol{88.47\pm0.84}$ \\
 & VISReg & \relax \relax \relax $\mathcal D$-only \relax & \relax \relax \relax Ep.\,1 \relax & $73.56\pm0.51$ & $80.11\pm5.32$ & $58.67\pm38.23$ & $23.67\pm33.50$ & $59.00\pm16.07$ \\
 & VISReg & +SADL & \relax \relax \relax Ep.\,1 \relax & $78.00\pm2.73$ & $86.56\pm0.19$ & $98.22\pm0.84$ & $93.33\pm2.89$ & $\boldsymbol{89.03\pm1.25}$ \\
 & QQWorld & \relax \relax \relax $\mathcal D$-only \relax & \relax \relax \relax Ep.\,1 \relax & $74.22\pm1.50$ & $82.33\pm6.66$ & $51.78\pm33.01$ & $3.67\pm4.93$ & $53.00\pm6.07$ \\
 & QQWorld & +SADL & \relax \relax \relax Ep.\,1 \relax & $79.89\pm1.26$ & $86.67\pm1.76$ & $98.11\pm0.19$ & $92.44\pm3.91$ & $\boldsymbol{89.28\pm1.64}$ \\
\bottomrule
\end{tabularx}
\endgroup
\end{table}

\subsubsection{Joint Moderate}
\begin{table}[H]
\centering
\caption{Joint four-task LeWM Moderate success (\%). \relax \relax \relax Epoch 1 \relax and \relax \relax \relax epoch 5 \relax compare equal training budgets; \relax \relax \relax $\mathcal D$-only epoch 10 \relax provides a longer-budget reference. LeWM uses Pure-CEM with $T_H=R_H=5$. \relax \relax \relax $\mathcal D$-only \relax uses a MATH encoder; +SADL uses a Flash encoder; both predictors use MATH. All entries are mean $\pm$ sample SD over three training seeds.}
\label{tab:app_d_joint_lewm}
\begingroup\relax %
\relax \relax \scriptsize\relax \setlength{\tabcolsep}{2pt}\relax %
\relax \relax \renewcommand{\arraystretch}{0.88}
\relax \begin{tabularx}{\linewidth}{@{}llcYYYYY@{}}
\toprule
\relax \relax \relax $\mathcal D$ \relax & Method & Epoch & PushT & Cube & Reacher & TwoRoom & Macro \\
\midrule
SIGReg & \relax \relax \relax $\mathcal D$-only \relax & \relax \relax \relax Ep.\,1 \relax & $12.89\pm3.34$ & $29.89\pm1.35$ & $14.67\pm5.78$ & $20.33\pm9.26$ & $19.44\pm3.59$ \\
SIGReg & +SADL & \relax \relax \relax Ep.\,1 \relax & $69.00\pm2.96$ & $49.00\pm2.00$ & $81.44\pm2.01$ & $58.11\pm1.58$ & $\boldsymbol{64.39\pm1.44}$ \\
SIGReg & \relax \relax \relax $\mathcal D$-only \relax & \relax \relax \relax Ep.\,5 \relax & $79.00\pm3.48$ & $50.44\pm4.50$ & $61.56\pm16.78$ & $22.33\pm18.05$ & $53.33\pm3.63$ \\
SIGReg & +SADL & \relax \relax \relax Ep.\,5 \relax & $68.33\pm2.73$ & $49.00\pm2.00$ & $67.67\pm13.86$ & $58.11\pm1.58$ & $\boldsymbol{60.78\pm3.63}$ \\
SIGReg & \relax \relax \relax $\mathcal D$-only \relax & \relax \relax \relax Ep.\,10 \relax & $76.67\pm3.76$ & $50.78\pm1.35$ & $70.44\pm18.03$ & $28.56\pm11.03$ & $56.61\pm6.49$ \\
\midrule
VISReg & \relax \relax \relax $\mathcal D$-only \relax & \relax \relax \relax Ep.\,1 \relax & $52.67\pm8.17$ & $32.56\pm3.10$ & $21.11\pm6.30$ & $32.22\pm15.87$ & $34.64\pm3.67$ \\
VISReg & +SADL & \relax \relax \relax Ep.\,1 \relax & $62.56\pm21.91$ & $45.89\pm7.34$ & $79.56\pm3.29$ & $80.00\pm8.97$ & $\boldsymbol{67.00\pm3.67}$ \\
VISReg & \relax \relax \relax $\mathcal D$-only \relax & \relax \relax \relax Ep.\,5 \relax & $84.78\pm0.51$ & $39.67\pm3.84$ & $79.89\pm0.69$ & $40.56\pm10.21$ & $61.22\pm2.94$ \\
VISReg & +SADL & \relax \relax \relax Ep.\,5 \relax & $83.22\pm2.27$ & $47.67\pm3.48$ & $82.56\pm1.35$ & $68.00\pm4.06$ & $\boldsymbol{70.36\pm0.53}$ \\
VISReg & \relax \relax \relax $\mathcal D$-only \relax & \relax \relax \relax Ep.\,10 \relax & $86.33\pm2.19$ & $39.33\pm1.53$ & $70.56\pm20.69$ & $47.44\pm6.16$ & $60.92\pm4.79$ \\
\midrule
QQWorld & \relax \relax \relax $\mathcal D$-only \relax & \relax \relax \relax Ep.\,1 \relax & $14.56\pm4.86$ & $31.89\pm6.55$ & $19.89\pm3.69$ & $26.22\pm10.87$ & $23.14\pm5.32$ \\
QQWorld & +SADL & \relax \relax \relax Ep.\,1 \relax & $71.44\pm9.31$ & $50.78\pm1.35$ & $79.44\pm3.53$ & $67.67\pm14.47$ & $\boldsymbol{67.33\pm1.04}$ \\
QQWorld & \relax \relax \relax $\mathcal D$-only \relax & \relax \relax \relax Ep.\,5 \relax & $80.67\pm5.49$ & $49.67\pm3.18$ & $68.78\pm21.17$ & $35.11\pm14.26$ & $58.56\pm10.60$ \\
QQWorld & +SADL & \relax \relax \relax Ep.\,5 \relax & $76.78\pm3.67$ & $52.11\pm1.17$ & $80.33\pm0.88$ & $58.56\pm1.68$ & $\boldsymbol{66.94\pm0.97}$ \\
QQWorld & \relax \relax \relax $\mathcal D$-only \relax & \relax \relax \relax Ep.\,10 \relax & $87.33\pm1.67$ & $49.00\pm5.03$ & $68.67\pm13.89$ & $38.44\pm7.12$ & $60.86\pm3.57$ \\
\bottomrule
\end{tabularx}
\endgroup
\end{table}

\begin{table}[H]
\centering
\caption{Joint four-task INTACT Moderate success (\%) with search-free Direct control. \relax \relax \relax Epoch 1 \relax and \relax \relax \relax epoch 5 \relax compare equal training budgets; \relax \relax \relax $\mathcal D$-only epoch 10 \relax provides a longer-budget reference. \relax \relax \relax $\mathcal D$-only \relax uses a MATH encoder; +SADL uses a Flash encoder; both predictors use MATH. All entries are mean $\pm$ sample SD over three training seeds.}
\label{tab:app_d_joint_intact}
\begingroup\relax %
\relax \relax \scriptsize\relax \setlength{\tabcolsep}{2pt}\relax %
\relax \relax \renewcommand{\arraystretch}{0.88}
\relax \begin{tabularx}{\linewidth}{@{}llcYYYYY@{}}
\toprule
\relax \relax \relax $\mathcal D$ \relax & Method & Epoch & PushT & Cube & Reacher & TwoRoom & Macro \\
\midrule
SIGReg & \relax \relax \relax $\mathcal D$-only \relax & \relax \relax \relax Ep.\,1 \relax & $39.11\pm9.14$ & $54.11\pm12.04$ & $9.89\pm4.17$ & $41.33\pm8.21$ & $36.11\pm5.28$ \\
SIGReg & +SADL & \relax \relax \relax Ep.\,1 \relax & $86.22\pm0.38$ & $100.00\pm0.00$ & $97.00\pm0.58$ & $99.67\pm0.00$ & $\boldsymbol{95.72\pm0.05}$ \\
SIGReg & \relax \relax \relax $\mathcal D$-only \relax & \relax \relax \relax Ep.\,5 \relax & $87.56\pm5.59$ & $99.67\pm0.58$ & $97.00\pm1.45$ & $58.89\pm20.42$ & $85.78\pm3.78$ \\
SIGReg & +SADL & \relax \relax \relax Ep.\,5 \relax & $92.89\pm0.38$ & $100.00\pm0.00$ & $99.44\pm0.69$ & $99.78\pm0.19$ & $\boldsymbol{98.03\pm0.13}$ \\
SIGReg & \relax \relax \relax $\mathcal D$-only \relax & \relax \relax \relax Ep.\,10 \relax & $91.00\pm3.18$ & $100.00\pm0.00$ & $96.89\pm0.69$ & $73.89\pm15.08$ & $90.44\pm3.14$ \\
\midrule
VISReg & \relax \relax \relax $\mathcal D$-only \relax & \relax \relax \relax Ep.\,1 \relax & $55.78\pm3.83$ & $69.78\pm13.62$ & $31.11\pm9.67$ & $55.00\pm11.32$ & $52.92\pm4.08$ \\
VISReg & +SADL & \relax \relax \relax Ep.\,1 \relax & $86.78\pm1.84$ & $100.00\pm0.00$ & $97.56\pm0.84$ & $99.67\pm0.00$ & $\boldsymbol{96.00\pm0.46}$ \\
VISReg & \relax \relax \relax $\mathcal D$-only \relax & \relax \relax \relax Ep.\,5 \relax & $91.89\pm1.84$ & $98.56\pm0.51$ & $97.67\pm0.58$ & $69.22\pm3.91$ & $89.33\pm0.92$ \\
VISReg & +SADL & \relax \relax \relax Ep.\,5 \relax & $93.22\pm0.38$ & $100.00\pm0.00$ & $98.89\pm0.77$ & $99.78\pm0.38$ & $\boldsymbol{97.97\pm0.10}$ \\
VISReg & \relax \relax \relax $\mathcal D$-only \relax & \relax \relax \relax Ep.\,10 \relax & $93.78\pm1.64$ & $100.00\pm0.00$ & $97.78\pm0.84$ & $89.11\pm5.30$ & $95.17\pm1.10$ \\
\midrule
QQWorld & \relax \relax \relax $\mathcal D$-only \relax & \relax \relax \relax Ep.\,1 \relax & $37.78\pm10.67$ & $60.78\pm3.15$ & $16.56\pm7.07$ & $30.89\pm15.12$ & $36.50\pm4.59$ \\
QQWorld & +SADL & \relax \relax \relax Ep.\,1 \relax & $88.78\pm0.84$ & $100.00\pm0.00$ & $97.44\pm0.38$ & $99.44\pm0.19$ & $\boldsymbol{96.42\pm0.22}$ \\
QQWorld & \relax \relax \relax $\mathcal D$-only \relax & \relax \relax \relax Ep.\,5 \relax & $90.11\pm1.26$ & $99.89\pm0.19$ & $97.67\pm0.58$ & $77.11\pm22.04$ & $91.19\pm5.22$ \\
QQWorld & +SADL & \relax \relax \relax Ep.\,5 \relax & $92.00\pm1.15$ & $100.00\pm0.00$ & $99.00\pm0.58$ & $99.67\pm0.33$ & $\boldsymbol{97.67\pm0.08}$ \\
QQWorld & \relax \relax \relax $\mathcal D$-only \relax & \relax \relax \relax Ep.\,10 \relax & $92.56\pm1.17$ & $100.00\pm0.00$ & $97.33\pm0.58$ & $85.22\pm22.13$ & $93.78\pm5.38$ \\
\bottomrule
\end{tabularx}
\endgroup
\end{table}

\subsubsection{Joint Strict}
\begin{table}[H]
\centering
\caption{Joint four-task LeWM CLEAR-LeWM v0.8 Strict success (\%). \relax \relax \relax Epoch 1 \relax and \relax \relax \relax epoch 5 \relax compare equal training budgets; \relax \relax \relax $\mathcal D$-only epoch 10 \relax is a longer-budget reference. LeWM uses Pure-CEM with $T_H=R_H=5$. \relax \relax \relax $\mathcal D$-only \relax uses a MATH encoder; +SADL uses a Flash encoder; both predictors use MATH. Values are mean $\pm$ sample SD over three training seeds.}
\label{tab:app_e_strict_joint_lewm}
\begingroup\relax %
\relax \relax \scriptsize\relax \setlength{\tabcolsep}{2pt}\relax %
\relax \relax \renewcommand{\arraystretch}{0.88}
\relax \begin{tabularx}{\linewidth}{@{}llcYYYYY@{}}
\toprule
\relax \relax \relax $\mathcal D$ \relax & Method & Epoch & PushT & Cube & Reacher & TwoRoom & Macro \\
\midrule
SIGReg & \relax \relax \relax $\mathcal D$-only \relax & \relax \relax \relax Ep.\,1 \relax & $16.00\pm4.93$ & $9.56\pm0.19$ & $21.11\pm3.56$ & $10.33\pm7.09$ & $14.25\pm3.44$ \\
SIGReg & +SADL & \relax \relax \relax Ep.\,1 \relax & $61.78\pm1.90$ & $24.56\pm2.52$ & $82.67\pm1.76$ & $40.00\pm1.45$ & $\boldsymbol{52.25\pm0.30}$ \\
SIGReg & \relax \relax \relax $\mathcal D$-only \relax & \relax \relax \relax Ep.\,5 \relax & $63.00\pm3.61$ & $26.67\pm2.96$ & $64.22\pm19.33$ & $11.78\pm11.21$ & $41.42\pm4.50$ \\
SIGReg & +SADL & \relax \relax \relax Ep.\,5 \relax & $62.22\pm2.14$ & $24.56\pm2.52$ & $72.44\pm7.71$ & $40.00\pm1.45$ & $\boldsymbol{49.81\pm2.14}$ \\
SIGReg & \relax \relax \relax $\mathcal D$-only \relax & \relax \relax \relax Ep.\,10 \relax & $62.44\pm7.32$ & $26.33\pm3.61$ & $69.00\pm24.29$ & $17.22\pm5.74$ & $43.75\pm6.94$ \\
\midrule
VISReg & \relax \relax \relax $\mathcal D$-only \relax & \relax \relax \relax Ep.\,1 \relax & $35.56\pm5.60$ & $12.22\pm2.59$ & $30.00\pm11.62$ & $18.00\pm6.66$ & $23.94\pm4.15$ \\
VISReg & +SADL & \relax \relax \relax Ep.\,1 \relax & $54.11\pm15.35$ & $25.22\pm6.36$ & $81.67\pm2.73$ & $64.56\pm15.91$ & $\boldsymbol{56.39\pm1.55}$ \\
VISReg & \relax \relax \relax $\mathcal D$-only \relax & \relax \relax \relax Ep.\,5 \relax & $73.78\pm4.91$ & $20.78\pm0.77$ & $82.22\pm0.19$ & $23.11\pm7.63$ & $49.97\pm1.63$ \\
VISReg & +SADL & \relax \relax \relax Ep.\,5 \relax & $71.78\pm1.17$ & $28.44\pm1.35$ & $83.22\pm2.22$ & $46.33\pm0.33$ & $\boldsymbol{57.44\pm1.04}$ \\
VISReg & \relax \relax \relax $\mathcal D$-only \relax & \relax \relax \relax Ep.\,10 \relax & $76.11\pm2.69$ & $22.11\pm0.84$ & $68.89\pm26.69$ & $24.11\pm1.50$ & $47.81\pm6.06$ \\
\midrule
QQWorld & \relax \relax \relax $\mathcal D$-only \relax & \relax \relax \relax Ep.\,1 \relax & $21.11\pm6.19$ & $13.00\pm3.18$ & $26.33\pm4.73$ & $14.56\pm8.44$ & $18.75\pm5.28$ \\
QQWorld & +SADL & \relax \relax \relax Ep.\,1 \relax & $61.78\pm11.48$ & $30.33\pm3.06$ & $84.11\pm2.46$ & $47.89\pm17.44$ & $\boldsymbol{56.03\pm1.74}$ \\
QQWorld & \relax \relax \relax $\mathcal D$-only \relax & \relax \relax \relax Ep.\,5 \relax & $66.22\pm6.27$ & $29.22\pm3.29$ & $79.56\pm6.58$ & $19.56\pm9.22$ & $48.64\pm5.98$ \\
QQWorld & +SADL & \relax \relax \relax Ep.\,5 \relax & $69.00\pm2.00$ & $29.44\pm1.68$ & $85.11\pm0.84$ & $37.78\pm0.84$ & $\boldsymbol{55.33\pm0.71}$ \\
QQWorld & \relax \relax \relax $\mathcal D$-only \relax & \relax \relax \relax Ep.\,10 \relax & $72.22\pm4.29$ & $28.33\pm2.40$ & $74.11\pm9.71$ & $21.89\pm3.95$ & $49.14\pm3.74$ \\
\bottomrule
\end{tabularx}
\endgroup
\end{table}

\begin{table}[H]
\centering
\caption{Joint four-task INTACT CLEAR-LeWM v0.8 Strict success (\%). \relax \relax \relax Epoch 1 \relax and \relax \relax \relax epoch 5 \relax compare equal training budgets; \relax \relax \relax $\mathcal D$-only epoch 10 \relax is a longer-budget reference. INTACT uses search-free Direct control. \relax \relax \relax $\mathcal D$-only \relax uses a MATH encoder; +SADL uses a Flash encoder; both predictors use MATH. Values are mean $\pm$ sample SD over three training seeds.}
\label{tab:app_e_strict_joint_intact}
\begingroup\relax %
\relax \relax \scriptsize\relax \setlength{\tabcolsep}{2pt}\relax %
\relax \relax \renewcommand{\arraystretch}{0.88}
\relax \begin{tabularx}{\linewidth}{@{}llcYYYYY@{}}
\toprule
\relax \relax \relax $\mathcal D$ \relax & Method & Epoch & PushT & Cube & Reacher & TwoRoom & Macro \\
\midrule
SIGReg & \relax \relax \relax $\mathcal D$-only \relax & \relax \relax \relax Ep.\,1 \relax & $36.00\pm7.23$ & $43.33\pm8.95$ & $20.67\pm7.54$ & $24.33\pm10.79$ & $31.08\pm2.04$ \\
SIGReg & +SADL & \relax \relax \relax Ep.\,1 \relax & $77.56\pm0.51$ & $88.89\pm1.17$ & $98.00\pm0.88$ & $97.33\pm0.67$ & $\boldsymbol{90.44\pm0.55}$ \\
SIGReg & \relax \relax \relax $\mathcal D$-only \relax & \relax \relax \relax Ep.\,5 \relax & $74.78\pm4.02$ & $89.44\pm1.07$ & $98.22\pm0.51$ & $37.33\pm20.80$ & $74.94\pm3.81$ \\
SIGReg & +SADL & \relax \relax \relax Ep.\,5 \relax & $84.33\pm1.20$ & $96.67\pm5.20$ & $98.56\pm0.38$ & $97.44\pm0.96$ & $\boldsymbol{94.25\pm1.42}$ \\
SIGReg & \relax \relax \relax $\mathcal D$-only \relax & \relax \relax \relax Ep.\,10 \relax & $81.00\pm2.19$ & $88.67\pm0.58$ & $98.44\pm0.19$ & $59.33\pm19.78$ & $81.86\pm4.32$ \\
\midrule
VISReg & \relax \relax \relax $\mathcal D$-only \relax & \relax \relax \relax Ep.\,1 \relax & $42.56\pm3.53$ & $55.33\pm12.55$ & $54.78\pm14.28$ & $32.11\pm14.05$ & $46.19\pm3.44$ \\
VISReg & +SADL & \relax \relax \relax Ep.\,1 \relax & $74.89\pm2.01$ & $87.67\pm1.20$ & $97.89\pm0.77$ & $95.22\pm1.02$ & $\boldsymbol{88.92\pm0.22}$ \\
VISReg & \relax \relax \relax $\mathcal D$-only \relax & \relax \relax \relax Ep.\,5 \relax & $78.11\pm1.64$ & $86.56\pm1.17$ & $99.44\pm0.38$ & $52.56\pm7.07$ & $79.17\pm1.83$ \\
VISReg & +SADL & \relax \relax \relax Ep.\,5 \relax & $83.56\pm0.38$ & $89.56\pm0.19$ & $98.44\pm0.69$ & $96.22\pm1.17$ & $\boldsymbol{91.94\pm0.29}$ \\
VISReg & \relax \relax \relax $\mathcal D$-only \relax & \relax \relax \relax Ep.\,10 \relax & $81.78\pm2.59$ & $88.56\pm0.51$ & $98.56\pm1.17$ & $74.78\pm6.06$ & $85.92\pm2.08$ \\
\midrule
QQWorld & \relax \relax \relax $\mathcal D$-only \relax & \relax \relax \relax Ep.\,1 \relax & $40.56\pm5.42$ & $47.11\pm6.11$ & $33.89\pm12.85$ & $11.00\pm4.10$ & $33.14\pm1.88$ \\
QQWorld & +SADL & \relax \relax \relax Ep.\,1 \relax & $76.67\pm2.03$ & $88.22\pm1.26$ & $97.89\pm0.69$ & $96.11\pm1.17$ & $\boldsymbol{89.72\pm0.96}$ \\
QQWorld & \relax \relax \relax $\mathcal D$-only \relax & \relax \relax \relax Ep.\,5 \relax & $79.11\pm2.91$ & $89.56\pm0.69$ & $99.00\pm0.33$ & $59.22\pm28.55$ & $81.72\pm7.39$ \\
QQWorld & +SADL & \relax \relax \relax Ep.\,5 \relax & $83.56\pm2.69$ & $96.89\pm5.10$ & $98.44\pm0.51$ & $96.22\pm0.19$ & $\boldsymbol{93.78\pm1.25}$ \\
QQWorld & \relax \relax \relax $\mathcal D$-only \relax & \relax \relax \relax Ep.\,10 \relax & $81.33\pm1.73$ & $92.00\pm0.58$ & $98.56\pm0.19$ & $73.89\pm27.35$ & $86.44\pm7.00$ \\
\bottomrule
\end{tabularx}
\endgroup
\end{table}

\relax %
\relax \section{Structural Ablations and Mechanism Diagnostics}
\label{app:ahead_mechanism}

\subsection{Supply branches and \relax \relax \relax Supply Regularizer \relax potentials}
All structural ablations use Flash, SIGReg weight $.03$,
\relax \relax $\lambda_S=.01$, and $\alpha=.5$. \relax \relax \relax exc-only and tr-only
\relax replace $k$ by \relax \relax \relax $k_{\mathrm{exc}}$ or $k_{\mathrm{tr}}$\relax , respectively. Potential
variants compare $[u]_+^2$, $u^2$, and $[u]_+$ for
$u=\log(q+\epsilon)-\log(k+\epsilon)$. \relax \relax \relax $\mathcal S$-only \relax still evaluates
detached demand but removes the \relax \relax \relax Demand Regularizer\relax 's direct training
gradient.
\relax %
\relax \relax \begin{table}[!htbp]
\relax \centering
\caption{\relax \relax \relax Epoch 2 \relax supply-branch and endpoint-gradient ablations under SIGReg. Entries are four-task CLEAR-LeWM v0.8 macro success, mean $\pm$ sample SD (\%) over three training seeds, with three evaluation seeds per task. All use Flash.}
\label{tab:app_structural_ablation}
\SADLTableStyle
\begin{tabularx}{\linewidth}{@{}LYYYY@{}}
\toprule
Configuration
& INTACT Mod.
& INTACT Strict
& LeWM Mod.
& LeWM Strict \\
\midrule
\relax \relax \relax exc-only\relax & $97.14\pm0.17$ & $91.03\pm0.61$
& $62.78\pm2.00$ & $50.94\pm3.26$ \\
\relax \relax \relax tr-only\relax & $42.94\pm45.45$ & $38.33\pm41.64$
& $35.28\pm27.30$ & $26.97\pm20.63$ \\
\relax \relax \relax $\mathcal S$-only\relax & $47.00\pm5.59$ & $35.33\pm1.86$
& $11.72\pm1.80$ & $7.81\pm0.49$ \\
\relax \relax SADL
& $\boldsymbol{97.14\pm0.17}$
& $\boldsymbol{92.06\pm1.04}$
& $\boldsymbol{65.50\pm0.85}$
& $\boldsymbol{53.03\pm1.42}$ \\
\bottomrule
\end{tabularx}
\end{table}

The \relax \relax \relax exc-only branch matches \relax SADL's INTACT \relax \relax \relax epoch 2 \relax Moderate
macro success ($97.14\%$) and trails it by $2.72$ percentage points
in LeWM ($62.78\%$ versus $65.50\%$). \relax \relax \relax tr-only \relax is markedly
weaker and more seed-variable in both models. \relax \relax \relax $\mathcal S$-only \relax falls to
$47.00\%$ in INTACT and $11.72\%$ in LeWM on Moderate, showing that the
\relax \relax \relax Supply Regularizer \relax complements rather than replaces the direct \relax \relax \relax gradient
of the Demand Regularizer\relax . In the independently
constructed transmission-deficit setting of
Table~\ref{tab:app_e_channel_roles}(B), \relax \relax \relax SADL raises $k_{\mathrm{tr}}$
\relax from $.180$ to $.746$ and held-out $R^2$ from $-.018$ to $.535$,
whereas exc\relax \relax -only leaves both essentially unchanged.

\relax %
\relax \relax \begin{table}[!htbp]
\relax \centering
\caption{Complete A2 \relax \relax \relax Supply Regularizer-potential \relax ablation under SIGReg on four-task CLEAR-LeWM v0.8. Within each model and epoch, the canonical one-sided square $[u]_+^2$, \relax \relax \relax two-sided \relax square $u^2$, and one-sided linear $[u]_+$ use matched training and evaluation protocols. Entries are macro success mean $\pm$ sample SD (\%) over three training seeds, with three evaluation seeds per task. LeWM uses Pure-CEM; INTACT uses Direct control.}
\relax %
\relax \relax \label{tab:app_potential}
\relax \SADLTableStyle
\begin{tabularx}{\linewidth}{@{}LcYY@{}}
\toprule
Potential & Epoch & Moderate & Strict \\
\midrule
\multicolumn{4}{@{}l}{\textit{LeWM}} \\
Canonical $[u]_+^2$ & \relax \relax \relax Ep.\,1 \relax & $\boldsymbol{64.39\pm1.44}$ & $\boldsymbol{52.25\pm0.30}$ \\
\relax \relax \relax Two-sided \relax $u^2$ & \relax \relax \relax Ep.\,1 \relax & $59.58\pm4.38$ & $52.11\pm2.66$ \\
Linear $[u]_+$ & \relax \relax \relax Ep.\,1 \relax & $47.17\pm5.55$ & $39.92\pm6.45$ \\
Canonical $[u]_+^2$ & \relax \relax \relax Ep.\,2 \relax & $\boldsymbol{65.50\pm0.85}$ & $\boldsymbol{53.03\pm1.42}$ \\
\relax \relax \relax Two-sided \relax $u^2$ & \relax \relax \relax Ep.\,2 \relax & $61.00\pm4.86$ & $47.44\pm6.22$ \\
Linear $[u]_+$ & \relax \relax \relax Ep.\,2 \relax & $53.53\pm4.19$ & $42.03\pm6.65$ \\
\midrule
\multicolumn{4}{@{}l}{\textit{INTACT}} \\
Canonical $[u]_+^2$ & \relax \relax \relax Ep.\,1 \relax & $95.72\pm0.05$ & $\boldsymbol{90.44\pm0.55}$ \\
\relax \relax \relax Two-sided \relax $u^2$ & \relax \relax \relax Ep.\,1 \relax & $95.61\pm0.47$ & $89.97\pm1.80$ \\
Linear $[u]_+$ & \relax \relax \relax Ep.\,1 \relax & $\boldsymbol{96.06\pm0.34}$ & $89.44\pm0.59$ \\
Canonical $[u]_+^2$ & \relax \relax \relax Ep.\,2 \relax & $\boldsymbol{97.14\pm0.17}$ & $\boldsymbol{92.06\pm1.04}$ \\
\relax \relax \relax Two-sided \relax $u^2$ & \relax \relax \relax Ep.\,2 \relax & $93.97\pm5.76$ & $88.03\pm7.90$ \\
Linear $[u]_+$ & \relax \relax \relax Ep.\,2 \relax & $\boldsymbol{97.14\pm0.43}$ & $90.86\pm0.67$ \\
\bottomrule
\end{tabularx}
\end{table}

Table~\relax \relax \relax \ref{tab:app_potential} \relax isolates potential shape. The
canonical and \relax \relax \relax two-sided \relax squares have identical values and gradients
for $u>0$; only the \relax \relax \relax two-sided \relax potential continues to penalize the
already-sufficient regime $u<0$, while the canonical \relax \relax \relax Supply Regularizer \relax is zero
there by construction. At equal \relax \relax \relax epochs 1/2 \relax budgets, canonical exceeds
\relax \relax \relax two-sided \relax in LeWM Moderate macro success by $4.81/4.50$ percentage
points; the \relax \relax \relax epoch 1 \relax gain is positive for all three paired training seeds.
INTACT is nearly tied at \relax \relax \relax epoch 1 \relax ($95.72\%$ versus $95.61\%$), but canonical
leads by $3.17$ points at \relax \relax \relax epoch 2 \relax ($97.14\%$ versus $93.97\%$). The large
\relax \relax \relax two-sided epoch 2 \relax seed SD in INTACT means this last mean difference is a
robustness signal, not a uniform gain across seeds. The one-sided
linear form remains competitive in INTACT but loses $17.22/11.97$
points to canonical in LeWM Moderate success at \relax \relax \relax epochs 1/2. \relax Together, the
matched A2 results support the practical value of a one-sided gate
and the squared potential across the two control models.

\relax \relax Persistent
excitation growth explains the early expansion and convergence improvement.
\relax \relax \relax tr-only \relax has high seed-level variance due to bimodal seed failures,
not episode noise. Its \relax \relax \relax epoch 2 \relax Moderate macro results by training seed are
\[
\begin{array}{ll}
\text{INTACT:}&95.42\%,\ 17.33\%,\ 16.08\%,\\
\text{LeWM:}&65.67\%,\ 12.83\%,\ 27.33\%.
\end{array}
\]
\relax \relax Section~\ref{sec:app_channel_roles} \relax gives independent transmission-limited tests\relax .

\subsection{Demand, supply, and gradients at fixed checkpoints}
The audit contains 168 observations: two model families, seven
configurations, three training seeds, and four tasks. Within each
model--seed--task group, configurations use the same deterministic
64-sample probe; statistics are computed task-locally. Define deficit,
active rate, and transmission-active rate by
\[
\Delta=\left[\log\frac{q+\epsilon}{k+\epsilon}\right]_+,
\quad p_{\rm act}=\Pr(k<q),
\quad p_{\rm tr,act}=\Pr(\relax \relax \relax k_{\mathrm{tr}}\relax <\relax \relax \relax k_{\mathrm{exc}}\relax ,\ k<q).
\]
The fixed-checkpoint diagnostics use \relax \relax \relax epochs 10/5/2; $\mathcal D$-only Supply Regularizer
\relax gradients are computed post hoc as counterfactuals. Tables
\ref{tab:app_fixed_snapshot_supply} and
\ref{tab:app_fixed_snapshot_gradients} report the observed states,
whereas Table~\ref{tab:app_structural_ablation} reports the equal-budget
\relax \relax \relax epoch 2 \relax control ablation. All \relax \relax 24 \relax \relax SADL
snapshots have an active \relax %
\relax \relax Supply Regularizer.\begin{table}[!htbp]
\relax \centering
\caption{Demand, supply, and control success at fixed checkpoints. $E$ is the audited epoch; capacity entries average 12 task--seed cells. $p_{\rm act}=\Pr(q>k)$. SR is independently evaluated four-task CLEAR-LeWM v0.8 Moderate macro success (\%; mean $\pm$ sample SD over three training seeds, with three evaluation seeds per task) at the listed epoch. The rows have different training budgets and are not an equal-budget SR comparison. $^\dagger$ denotes a counterfactual \relax \relax \relax Supply Regularizer \relax diagnostic on \relax \relax \relax $\mathcal D$-only \relax checkpoints; its SR is the actual \relax \relax \relax $\mathcal D$-only \relax result. Exc./Trans./\relax \relax \relax Two-sided\relax /Linear denote \relax \relax \relax exc-only\relax , \relax \relax \relax tr-only\relax , \relax \relax \relax two-sided \relax squared, and one-sided linear variants.}
\label{tab:app_fixed_snapshot_supply}
\begingroup\small\setlength{\tabcolsep}{3pt}\renewcommand{\arraystretch}{1.12}
\begin{tabularx}{\linewidth}{@{}lYYYYYYYr@{}}
\toprule
Setting & E & \relax \relax \relax $q_{\mathrm{reg}}$ \relax & \relax \relax \relax $q_{\mathrm{exc}}$ \relax & $q$
& \relax \relax \relax $k_{\mathrm{exc}}$ \relax & \relax \relax \relax $k_{\mathrm{tr}}$ \relax & $p_{\rm act}$ (\%) & SR (\%) \\
\midrule
\multicolumn{9}{l}{\textit{LeWM}} \\
Endpoint$^\dagger$ & 10
& .8049 & .6541 & .7174 & .3177 & .6948 & 100.0 & $56.61\pm6.49$ \\
\relax \relax \relax $\mathcal S$-only \relax & 2
& .0000 & .0781 & .0000 & .0439 & .1858 & 0.0 & $11.72\pm1.80$ \\
Trans. & 2
& .5416 & .3150 & .3886 & .2578 & .5062 & 41.7 & $35.28\pm27.30$ \\
\relax \relax \relax Two-sided \relax & 2
& .9496 & .5307 & .7098 & .5882 & .6944 & 100.0 & $61.00\pm4.86$ \\
Linear & 2
& .9479 & .5439 & .7179 & .6640 & .7287 & 100.0 & $53.53\pm4.19$ \\
Exc. & 5
& .9517 & .6127 & .7632 & .6426 & .7510 & 100.0 & $63.72\pm0.92$ \\
\relax \relax SADL & 5
& .9521 & .6133 & .7637 & .6430 & .7509 & 100.0 & $60.78\pm3.63$ \\
\midrule
\multicolumn{9}{l}{\textit{INTACT}} \\
Endpoint$^\dagger$ & 10
& .7381 & .6366 & .6789 & .2622 & .6918 & 100.0 & $90.44\pm3.14$ \\
\relax \relax \relax $\mathcal S$-only \relax & 2
& .0000 & .0791 & .0000 & .1196 & .3486 & 0.0 & $47.00\pm5.59$ \\
Trans. & 2
& .2947 & .2171 & .2225 & .2687 & .5673 & 25.0 & $42.94\pm45.45$ \\
\relax \relax \relax Two-sided \relax & 2
& .9343 & .5392 & .7096 & .5697 & .7246 & 100.0 & $93.97\pm5.76$ \\
Linear & 2
& .9497 & .5591 & .7286 & .6710 & .7539 & 100.0 & $97.14\pm0.43$ \\
Exc. & 5
& .9524 & .6411 & .7812 & .6384 & .7964 & 100.0 & $97.50\pm0.22$ \\
\relax \relax SADL & 5
& .9529 & .6362 & .7784 & .6331 & .7918 & 100.0 & $98.03\pm0.13$ \\
\bottomrule
\end{tabularx}
\endgroup
\end{table}

\relax %
\relax \relax \begin{table}[!htbp]
\relax \centering
\caption{\relax \relax \relax Supply Regularizer \relax deficit and encoder-parameter gradients on the same fixed probe. Epoch identities and independently evaluated CLEAR-LeWM v0.8 Moderate SR (\%; mean $\pm$ sample SD over three training seeds) match Table~\ref{tab:app_fixed_snapshot_supply}; different rows have different training budgets. $G_T,G_B$ are task and \relax \relax \relax Supply Regularizer \relax gradient norms; $G_B/G_T$ averages per-cell ratios. \relax \relax \relax $\mathcal S$-only \relax has zero \relax \relax \relax Supply Regularizer \relax gradient, so its cosine is undefined (n/a). $^\dagger$ denotes a counterfactual \relax \relax \relax Supply Regularizer \relax diagnostic.}
\label{tab:app_fixed_snapshot_gradients}
\begingroup\small\setlength{\tabcolsep}{3pt}\renewcommand{\arraystretch}{1.12}
\begin{tabular*}{\linewidth}{@{\extracolsep{\fill}}lcccccr@{}}
\toprule
Configuration & $\Delta$ & $G_T$ & $G_B$ & $G_B/G_T$ (\%) & $\cos(G_T,G_B)$ & SR (\%) \\
\midrule
\multicolumn{7}{l}{\textit{LeWM}} \\
Endpoint$^\dagger$
& .8180 & .25432 & .00756 & 3.843 & -.0251 & $56.61\pm6.49$ \\
\relax \relax \relax $\mathcal S$-only\relax & .0000 & .00092 & .00000 & .000 & n/a & $11.72\pm1.80$ \\
Trans.
& .0233 & 15.74698 & .00023 & .017 & .0987 & $35.28\pm27.30$ \\
\relax \relax \relax Two-sided\relax & .1886 & .16440 & .00196 & 1.763 & .0815 & $61.00\pm4.86$ \\
Linear
& .0790 & .14240 & .00408 & 3.697 & .1885 & $53.53\pm4.19$ \\
Exc.
& .1740 & .05307 & .00080 & 1.852 & .0588 & $63.72\pm0.92$ \\
\relax \relax SADL
& .1739 & .05342 & .00081 & 1.946 & .0592 & $60.78\pm3.63$ \\
\midrule
\multicolumn{7}{l}{\textit{INTACT}} \\
Endpoint$^\dagger$
& .9730 & .69619 & .01579 & 2.393 & .1254 & $90.44\pm3.14$ \\
\relax \relax \relax $\mathcal S$-only\relax & .0000 & 1.10121 & .00000 & .000 & n/a & $47.00\pm5.59$ \\
Trans.
& .0053 & 314.61248 & .00003 & .003 & .0966 & $42.94\pm45.45$ \\
\relax \relax \relax Two-sided\relax & .2223 & .97499 & .00455 & .948 & .2988 & $93.97\pm5.76$ \\
Linear
& .0829 & .24236 & .00472 & 2.123 & .1361 & $97.14\pm0.43$ \\
Exc.
& .2032 & .12964 & .00110 & .916 & .0648 & $97.50\pm0.22$ \\
\relax \relax SADL
& .2083 & .15412 & .00119 & .887 & .0416 & $98.03\pm0.13$ \\
\bottomrule
\end{tabular*}
\endgroup
\end{table}

The average \relax \relax \relax Supply Regularizer-to-task \relax encoder-gradient norm ratio is
$1.946\pm1.205\%$ for LeWM and $0.887\pm0.310\%$ for INTACT.
These are pre-update parameter gradients on the same probe,
$G_T=\|\nabla_\theta\mathcal L_{\rm task}\|_2$ and
$G_B=\|\nabla_\theta(\lambda_S\mathcal S)\|_2$; the INTACT task
term also contains action negative log-likelihood. For \relax \relax \relax SADL,
task--Supply Regularizer \relax cosine is $0.059\pm0.173$ in LeWM and
$0.042\pm0.131$ in INTACT; \relax \relax \relax Supply Regularizer--endpoint \relax cosine is
$0.338\pm0.197$ and $0.320\pm0.164$, respectively, with SD across
12 task--seed cells. A small \relax \relax \relax correction on $H$ thus changes representation
\relax geometry without overwhelming the task gradient. All \relax \relax \relax $\mathcal S$-only
snapshots have $q_{\mathrm{reg}}=0$ \relax and zero ideal demand; smoothed demand stays below
available supply despite nonzero endpoint excitation, so the \relax \relax \relax Supply Regularizer
\relax gradient is zero.

\subsection{Independent roles of the two supply capacities}
\label{sec:app_channel_roles}
Table~\ref{tab:app_e_channel_roles}(A) compares two initial capacity
deficits. In separate paired ten-seed, 200-step tests, panels B/C construct
\relax \relax \relax $k_{\mathrm{tr}}<q<k_{\mathrm{exc}}$ and $k_{\mathrm{exc}}<q<k_{\mathrm{tr}}$\relax , respectively.
Each runs \relax \relax \relax $\mathcal D$-only, exc-only, tr-only, and \relax SADL;
the table displays the ineffective single branch and \relax \relax \relax SADL\relax , since the
unshown paired arms coincide seed by seed in the corresponding controlled
\relax \relax \relax deficit\relax . The three setups respectively optimize $H$, encoder weights,
and a shared bias with fixed informative weights. B measures held-out
readout; C holds the information channel fixed and reports $H$ RMS.
\relax \relax \relax SADL uses $k=\min(k_{\mathrm{exc}},k_{\mathrm{tr}})$\relax ; each single-branch variant
uses one capacity alone.
\relax %
\relax \relax \begin{table}[!htbp]
\relax \centering
\caption{Paired mechanism tests of two capacity deficits with fixed demand $q=.70$. (A) Direct optimization of $H$ with 20 seeds. (B) Training encoder weights under a transmission deficit with held-out readout. (C) Training only a shared bias, with the informative weight channel fixed, under an excitation deficit. B/C use ten paired seeds each and display the ineffective single branch and \relax \relax SADL. Entries are initial $\to$ final mean $\pm$ seed SD.}
\label{tab:app_e_channel_roles}
\begingroup\small\setlength{\tabcolsep}{3pt}\renewcommand{\arraystretch}{1.25}
\begin{tabularx}{\linewidth}{@{}llYYY@{}}
\toprule
\multicolumn{5}{@{}l}{\textit{(A) Controlled capacity optimization, 20 seeds}} \\
\midrule
Initial deficit & \relax \relax \relax Supply Regularizer \relax target & \relax \relax \relax $k_{\mathrm{exc}}$\relax , 0$\to$200 & \relax \relax \relax $k_{\mathrm{tr}}$\relax , 0$\to$200 & \relax \relax \relax $\mathcal S$\relax , 0$\to$200 \\
\midrule
Excitation & \relax \relax \relax tr-only \relax & $.494\to.494\pm.002$ & $.999\to.999\pm.000$ & $.000\to.000$ \\
Excitation & \relax \relax SADL & $.494\to\boldsymbol{.713\pm.001}$ & $.999\to.975\pm.001$ & $.122\to.000$ \\
\midrule
Transmission & \relax \relax \relax exc-only \relax & $.976\to.976\pm.001$ & $.314\to.314\pm.009$ & $.000\to.000$ \\
Transmission & \relax \relax SADL & $.976\to.978\pm.001$ & $.314\to\boldsymbol{.789\pm.003}$ & $.646\to.000$ \\
\bottomrule
\end{tabularx}
\par\vspace{0.6\baselineskip}
\begin{tabularx}{\linewidth}{@{}LYYY@{}}
\toprule
\multicolumn{4}{@{}l}{\textit{(B) Transmission deficit, independent held-out readout, 10 seeds}} \\
\midrule
\relax \relax \relax Supply Regularizer \relax target & \relax \relax \relax $k_{\mathrm{exc}}$\relax , initial$\to$final & \relax \relax \relax $k_{\mathrm{tr}}$\relax , initial$\to$final & Held-out $R^2$, initial$\to$final \\
\midrule
\relax \relax \relax exc-only \relax & $.779\to.779\pm.003$ & $.180\to.180\pm.005$ & $-0.018\to-0.018\pm.002$ \\
\relax \relax SADL & $.779\to.861\pm.003$ & $.180\to\boldsymbol{.746\pm.011}$ & $-0.018\to.535\pm.018$ \\
\bottomrule
\end{tabularx}
\par\vspace{0.6\baselineskip}
\begin{tabularx}{\linewidth}{@{}LYYY@{}}
\toprule
\multicolumn{4}{@{}l}{\textit{(C) Excitation deficit, fixed weights/shared-bias training, 10 seeds}} \\
\midrule
\relax \relax \relax Supply Regularizer \relax target & \relax \relax \relax $k_{\mathrm{exc}}$\relax , initial$\to$final & \relax \relax \relax $k_{\mathrm{tr}}$\relax , initial$\to$final & $H$ RMS, initial$\to$final \\
\midrule
\relax \relax \relax tr-only \relax & $.442\to.442\pm.002$ & $.999\to.999\pm.000$ & $2.234\to2.234\pm.004$ \\
\relax \relax SADL & $.442\to\boldsymbol{.710\pm.002}$ & $.999\to.999\pm.000$ & $2.234\to1.459\pm.003$ \\
\bottomrule
\end{tabularx}
\endgroup
\end{table}

\begin{figure}[t]
\centering
\relax %
\relax \relax \includegraphics[width=\linewidth]{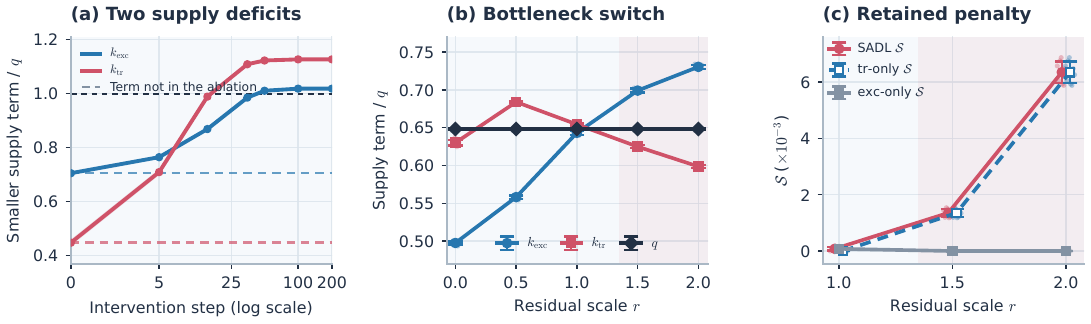}
\relax \caption{Independent capacity deficits and the transmission branch.
(a) Matched-initialization controlled optimization of $H$: blue denotes
excitation deficit, red transmission deficit; solid curves are \relax \relax SADL,
dashed curves lack the needed branch. Curves show mean $\pm$ seed SD over
20 seeds; the horizontal dashed line marks capacity equal to demand.
(b--c) Independent forward replay of an ImageNet-LT \relax \relax \relax epoch 100 \relax checkpoint:
endpoint and image mean are fixed while view-residual multiplier $r$ varies.
Pale dots show the same eight batches; filled dots and error bars show
mean $\pm$ batch SD. The hollow blue squares in (c) are same-state
\relax \relax \relax tr-only \relax counterfactual penalties, shifted horizontally only to
separate overlapping error bars. At $r\ge1.5$,
\relax \relax \relax $k_{\mathrm{tr}}<q<k_{\mathrm{exc}}$\relax : \relax \relax \relax SADL \relax and \relax \relax \relax tr-only \relax penalties coincide,
whereas \relax \relax \relax exc-only \relax is zero. Panels (b--c) are forward-replay mechanism
readouts, not retrained downstream success.}
\label{fig:app_e_roles}
\end{figure}

Figure~\ref{fig:app_e_roles} juxtaposes the controlled deficits with a
forward intervention on a real visual checkpoint. The former tests whether
each capacity can independently become limiting; the latter tests whether
the \relax \relax \relax Supply Regularizer \relax still responds when transmission falls below demand.
The same-state calculation shows why raising excitation alone cannot
guarantee that the \relax \relax \relax Supply Regularizer \relax has switched off: once transmission falls below
demand, \relax \relax \relax $k_{\mathrm{tr}}$ \relax still sets the minimum capacity $k$.

\subsection{Demand interpolation coefficient \texorpdfstring{$\alpha$}{alpha}}
$\alpha=0$ emphasizes \relax \relax \relax $q_{\mathrm{reg}}$ \relax and is generally more aggressive;
$\alpha=1$ emphasizes \relax \relax \relax $q_{\mathrm{exc}}$ \relax and is more conservative.
Table~\ref{tab:app_visual_alpha} gives \relax \relax \relax the resulting demand capacity \relax and outcome for
the same ImageNet-LT seed.
\relax %
\relax \relax \begin{table}[!htbp]
\relax \centering
\caption{Single-seed development ablation of demand interpolation $\alpha$ on ImageNet-LT (seed 42). $q$ is the \relax \relax \relax epoch 20 \relax demand \relax \relax \relax capacity\relax ; online accuracy is at \relax \relax \relax epoch 100. \relax $\alpha=0$ favors the more aggressive \relax \relax \relax $q_{\mathrm{reg}}$\relax , whereas $\alpha=1$ favors the \relax more \relax conservative \relax \relax \relax $q_{\mathrm{exc}}$\relax . In this setting, the highest observed result is at $\alpha=.5$.}
\label{tab:app_visual_alpha}
\begingroup\small\setlength{\tabcolsep}{8pt}\renewcommand{\arraystretch}{1.12}
\begin{tabularx}{\linewidth}{@{}LYYY@{}}
\toprule
Configuration & \relax \relax \relax Ep.\,20 \relax demand $q$ & \relax \relax \relax Ep.\,100 \relax Top-1 (\%) & vs. baseline (pp) \\
\midrule
VISReg baseline & n/a & 22.034 & 0.000 \\
SADL $\alpha=0$ & .772 & 22.904 & $+0.870$ \\
SADL $\alpha=.25$ & .772 & 22.348 & $+0.314$ \\
SADL $\alpha=.5$ & .667 & \textbf{23.026} & $\boldsymbol{+0.992}$ \\
SADL $\alpha=.75$ & .554 & 22.144 & $+0.110$ \\
SADL $\alpha=1$ & .460 & 22.064 & $+0.030$ \\
\bottomrule
\end{tabularx}
\endgroup
\end{table}

\relax %
\relax \subsection{Demand--Supply Geometry and \relax \relax \relax Dual Regularization\relax }
\label{sec:app_h_only_dual}

We compare a simple $H$-side barrier, \relax \relax \relax dual regularization (the same regularizer
on $H$ and $Z$)\relax , and SADL. For centered $H$ covariance $\Sigma_H$, the
barrier penalizes low average feature standard deviation or excessive
spectral concentration:
\begin{equation}
\mathcal R_{\rm bar}(H)=
[.10-\overline\sigma(H)]_+^2+
[\lambda_{\max}(\Sigma_H)/\operatorname{tr}(\Sigma_H)-.35]_+^2.
\label{eq:app_h_barrier}
\end{equation}
The tested barrier configuration retains SIGReg$(Z)$ at weight $.03$;
its $H$ weight is $.03$ in \relax \relax \relax epoch 1 \relax and $.003$ thereafter. It is therefore
an $H$-side regularizer added to \relax \relax \relax a Demand Regularizer\relax , not an
$H$-only training objective. The dual recipe instead applies the same
distributional penalty $\mathcal R$ at $H$ and $Z$, together with the
tested scale floors $F_\gamma$:
\begin{equation}
\mathcal L_{\rm dual}=\mathcal L_{\rm task}
+\lambda_R[\mathcal R(H)+\mathcal R(Z)]
+0.02[F_{0.5}(H)+F_{1.0}(Z)].
\label{eq:app_h_dual}
\end{equation}
Here $\mathcal R$ is SIGReg, VISReg, or QQWorld; $F_\gamma$ is the
fixed scale-floor term with threshold $\gamma$. The three tested
$\lambda_R$ values are $.02$, $1.5$, and $3.5$, respectively. This is a
\emph{dual-plus-floor} control, not a pure two-endpoint penalty. SADL
instead retains the selected \relax \relax \relax Demand Regularizer \relax $\lambda_D\mathcal D(Z)$
and applies the one-sided \relax \relax \relax Supply Regularizer
$\lambda_S\mathcal S(H,\operatorname{sg}(Z))$ \relax only for a
measured demand--supply deficit (Section~\ref{sec:theory}).

Table~\ref{tab:app_h_dual} reports four-task \relax \relax \relax official \relax control success
for three training seeds per recipe. LeWM uses Pure-CEM and INTACT uses Direct;
each task score is averaged over three evaluation seeds. The late
checkpoint is \relax \relax \relax epoch 2 \relax for dual LeWM and \relax \relax \relax epoch 5 \relax for SADL LeWM and both INTACT
recipes; LeWM late scores therefore are not equal-budget comparisons.
The barrier, dual-plus-floor, and SADL recipes were trained in separate
experimental matrices\relax \relax \relax . The table reports the observed recipes with their
training and evaluation protocols\relax .

\relax %
\relax \relax \begin{table}[!htbp]
\relax \centering
\caption{Simple $H$-side barrier, \relax \relax \relax dual \relax regularization with scale floors, and SADL in jointly trained four-task control. Official success rates are percentages; macro entries are mean $\pm$ sample SD over three training seeds. Superscripts in the late column give the checkpoint epoch.}
\label{tab:app_h_dual}
\begingroup
\scriptsize
\setlength{\tabcolsep}{1.7pt}
\renewcommand{\arraystretch}{1.13}
\begin{tabular*}{\linewidth}{@{\extracolsep{\fill}}lllcccccc@{}}
\toprule
Model & \relax \relax \relax $\mathcal D$ \relax & Recipe & \relax \relax \relax Ep.\,1 \relax PushT & \relax \relax \relax Ep.\,1 \relax Cube & \relax \relax \relax Ep.\,1 \relax Reacher & \relax \relax \relax Ep.\,1 \relax TwoRoom & \relax \relax \relax Ep.\,1 \relax Macro & Late Macro \\
\midrule
LeWM & SIGReg & Dual+floor & $7.56$ & $61.00$ & $80.67$ & $74.67$ & $55.97\pm.17$ & \relax \relax \relax $58.06\pm.89^{\rm Ep.\,2}$ \relax \\
 & & SADL & $28.78$ & $70.22$ & $79.89$ & $73.00$ & $\mathbf{62.97\pm1.25}$ & \relax \relax \relax $70.64\pm.77^{\rm Ep.\,5}$ \relax \\
 & VISReg & Dual+floor & $8.89$ & $65.33$ & $77.67$ & $81.44$ & $58.33\pm1.09$ & \relax \relax \relax $59.56\pm.35^{\rm Ep.\,2}$ \relax \\
 & & SADL & $53.56$ & $66.44$ & $78.44$ & $88.11$ & $\mathbf{71.64\pm2.96}$ & \relax \relax \relax $78.58\pm.98^{\rm Ep.\,5}$ \relax \\
 & QQWorld & Dual+floor & $7.22$ & $64.44$ & $78.89$ & $71.67$ & $55.56\pm1.65$ & \relax \relax \relax $56.31\pm2.12^{\rm Ep.\,2}$ \relax \\
 & & SADL & $45.67$ & $67.44$ & $78.00$ & $82.56$ & $\mathbf{68.42\pm4.19}$ & \relax \relax \relax $75.11\pm1.34^{\rm Ep.\,5}$ \relax \\
\midrule
INTACT & SIGReg & $H$ barrier & $26.56$ & $70.78$ & $11.22$ & $48.44$ & $39.25\pm12.74$ & \relax \relax \relax $48.00\pm41.68^{\rm Ep.\,5}$ \relax \\
\cmidrule(lr){2-9}
INTACT & SIGReg & Dual+floor & $42.11$ & $99.11$ & $94.78$ & $98.00$ & $83.50\pm1.99$ & \relax \relax \relax $95.81\pm.43^{\rm Ep.\,5}$ \relax \\
 & & SADL & $85.22$ & $100.00$ & $97.11$ & $100.00$ & $\mathbf{95.58\pm.17}$ & \relax \relax \relax $97.19\pm.51^{\rm Ep.\,5}$ \relax \\
 & VISReg & Dual+floor & $61.11$ & $99.44$ & $96.67$ & $97.56$ & $88.69\pm3.58$ & \relax \relax \relax $96.67\pm.36^{\rm Ep.\,5}$ \relax \\
 & & SADL & $83.78$ & $100.00$ & $97.78$ & $100.00$ & $\mathbf{95.39\pm.65}$ & \relax \relax \relax $97.53\pm.25^{\rm Ep.\,5}$ \relax \\
 & QQWorld & Dual+floor & $50.00$ & $99.67$ & $95.89$ & $96.45$ & $85.50\pm1.38$ & \relax \relax \relax $95.28\pm.46^{\rm Ep.\,5}$ \relax \\
 & & SADL & $86.00$ & $100.00$ & $96.56$ & $100.00$ & $\mathbf{95.64\pm.32}$ & \relax \relax \relax $97.56\pm.17^{\rm Ep.\,5}$ \relax \\
\bottomrule
\end{tabular*}
\endgroup
\end{table}

The simple barrier is not uniformly ineffective: its three \relax \relax \relax epoch 5
\relax training-seed macros are $28.67$, $19.50$, and $95.83\%$, but the
aggregate remains $48.00\pm41.68\%$. At \relax \relax \relax epoch 2 \relax its macro is
$62.00\pm28.10\%$, compared with $96.56\pm.17\%$ for three-seed
SIGReg+SADL \relax \relax \relax at epoch 2 of the mechanism audit behind the gate and
potential rows of Table~\ref{tab:robust}. Dual \relax regularization is also a
viable control: at \relax \relax \relax epoch 5 \relax it attains $95.28$--$96.67\%$ INTACT Direct
success. However, SADL is higher for all six model--endpoint pairs at
\relax \relax \relax epoch 1\relax , with the largest early gap on INTACT--SIGReg ($83.50\%$ versus
$95.58\%$). The gain is most visible on PushT. At the matched INTACT
\relax \relax \relax epoch 5 \relax checkpoint, SADL remains higher by $.86$--$2.28$ percentage
points; the margin has narrowed, but is not restricted to \relax \relax \relax epoch 1. \relax These
observations are consistent with adapting a
\relax \relax \relax correction on $H$ to the \relax measured demand and \relax \relax \relax supply capacities \relax rather than
enforcing a fixed geometry at both ends\relax \relax . Fixed-probe
effective-rank, cosine, and prediction-loss evidence for SADL is in
Table~\ref{tab:joint}; the dual-plus-floor runs do not have an audited
matched fixed-probe readout.

\relax %
\relax \relax 

\relax \section{Endpoint Distributions and Representation Geometry}
\label{app:ahead_geometry}

\subsection{Eight \relax \relax \relax Demand Regularizers \relax and designable two-dimensional geometry}

Let $Z\in\mathbb R^{T\times B\times d}$ and let $v_j$ be random unit projection directions. Define $Q_\nu(Z)$ by sorting each projected batch against the quantiles of a reference distribution $\nu$:
\[
Q_\nu(Z)=\frac{1}{TBK}\sum_{t=1}^T\sum_{j=1}^K\sum_{i=1}^B
\left((Z_tv_j)_{(i)}-F_\nu^{-1}\!\left(\frac{i-1/2}{B}\right)\right)^2.
\]
For independent reference samples $y,y'\sim\nu$, the empirical energy distance is
\[
E_\nu(Z)=\frac{1}{\sqrt d}\left[2\mathbb E\|z-y\|_2-
\mathbb E\|z-z'\|_2-\mathbb E\|y-y'\|_2\right]_+.
\]
At most 96 samples per temporal position are used to compute the energy distance.

\paragraph{Four Gaussian \relax \relax \relax Demand Regularizers\relax .}
Write $\nu_G=\mathcal N(0,I)$, let $\widehat\varphi_Z$ be the empirical characteristic function, and let $\mu_j,\sigma_j$ denote coordinate-wise means and standard deviations. Set $\widetilde Z=(Z-\mu)/\operatorname{sg}(\sigma)$. The four \relax \relax \relax Demand Regularizers \relax are
\begin{align*}
\mathcal D_{\rm SIG}(Z)&\propto
  \big\langle|\widehat\varphi_Z(\omega)-e^{-\|\omega\|^2/2}|^2\big\rangle_\omega,
&\mathcal D_{\rm QQ}(Z)&=Q_{\nu_G}(Z),\\
\mathcal D_{\rm VIS}(Z)&=
  \big\langle(\sigma_j-1)^2+\mu_j^2\big\rangle_j+Q_{\nu_G}(\widetilde Z),
&\mathcal D_{\rm EG}(Z)&=E_{\nu_G}(Z).
\end{align*}
All share a Gaussian reference, but respectively constrain the characteristic function, separated scale and shape, raw sorted projections, and sample distances; their gradients and finite-sample behavior differ.

\paragraph{Four non-Gaussian or partially specified \relax \relax \relax Demand Regularizers\relax .}
Let $C$ be the sample covariance of the $TB$ endpoint samples and $u=z/\|z\|_2$. VIC moments specifies no full probability density:
\[
\mathcal D_{\rm VIC}(Z)=
\left\langle\left[1-\sqrt{\operatorname{Var}(Z_j)+10^{-4}}\right]_+\right\rangle_j
+\left\langle C_{ij}^2\right\rangle_{i\ne j}
+\left\langle\mu_j^2\right\rangle_j.
\]
\noindent Sphere, Student-$t_5$, and GMM-Ring use, respectively,
\begin{align*}
\mathcal D_{\rm Sphere}(Z)&=
 \left\langle\bigl(\|z\|_2/\sqrt d-1\bigr)^2\right\rangle_z
 +\left\langle e^{-2\|u_i-u_j\|_2^2}\right\rangle_{t,i\ne j},\\
\mathcal D_{t_5}(Z)&=Q_{\nu_{t_5}}(Z),\qquad
 F_{\nu_{t_5}}^{-1}(p)=\sqrt{3/5}\,F_{t_5}^{-1}(p),\\
\mathcal D_{\rm Ring}(Z)&=E_{\nu_{\rm Ring}}(Z),\qquad
 \nu_{\rm Ring}=W_\#\!\left[
 \frac{1}{16}\sum_{\ell=0}^{15}
 \mathcal N\!\left(2.5m_\ell,\,0.35^2I\right)\right],\\
 m_\ell&=\cos\frac{2\pi\ell}{16}u_1+
 \sin\frac{2\pi\ell}{16}u_2.
\end{align*}
Here $W_\#$ analytically whitens the two-dimensional ring subspace and $u_1,u_2$ form an orthonormal basis. The two Sphere terms are added directly; Student-$t_5$ uses 128 projection directions. At $Z=0$, Sphere sets $u=0$ through its positive norm floor, and VISReg sets $\widetilde Z=0$ through its positive standard-deviation floor; the cached \relax \relax \relax $\mathcal D(0)$ \relax is evaluated with the same conventions.

\subsection{Learned endpoint geometry and designable glyphs}

The \relax \relax \relax control results for the eight Demand Regularizers \relax are reported in Table~\ref{tab:app_f_eight_official} in the main text. For SIGReg and QQWorld, the first-step encoder-gradient calibration was performed separately for LeWM and INTACT, with VISReg anchored at $\lambda_D=1.5$; the five additional \relax \relax \relax Demand Regularizers \relax were likewise calibrated at the first step of seed 0. Each model-specific endpoint weight was then held fixed during training.

\begin{figure}[!htbp]
\relax \centering
\relax %
\relax \relax \includegraphics[width=0.94\linewidth]{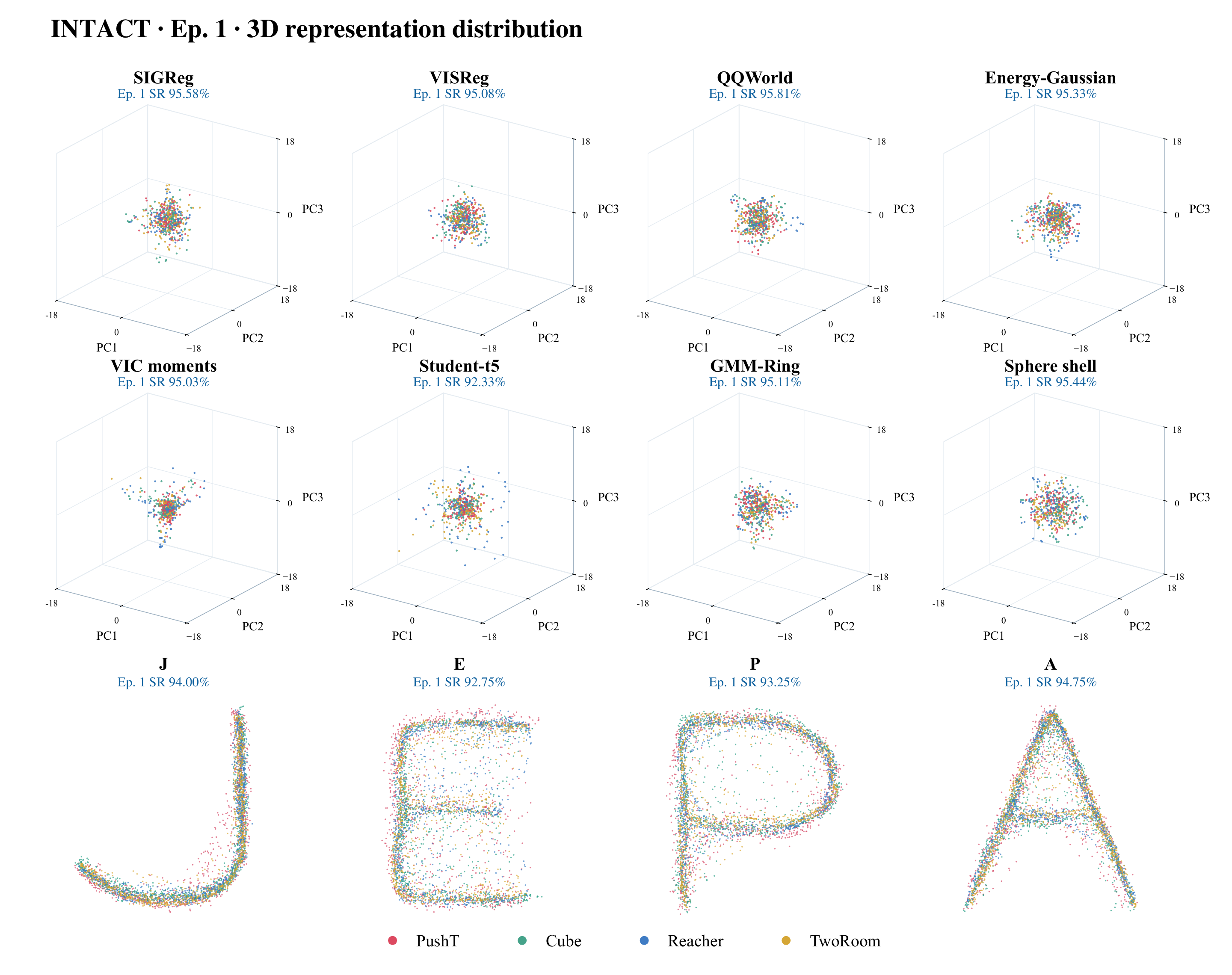}\relax \par
\vspace{0.7em}

\vspace{0pt}
\caption{Real \relax \relax \relax epoch 1 \relax representations with four-task color coding; all particles are vector objects. The first two rows show independent 3-D PCA projections of INTACT+SADL $z$ on a fixed probe; blue SR values are three-seed \relax \relax \relax official \relax Direct means. The third row shows independently trained raw 2-D glyph representations; blue SR values are single-seed Direct macro success, evaluated with 100 episodes per task, and are therefore not directly comparable to those above. The image illustrates learned local geometry, not the original high-dimensional density.}
\label{fig:geometry_gallery}
\end{figure}

Figure~\ref{fig:geometry_gallery} displays the corresponding \relax \relax \relax epoch 1 \relax representations on held-out clips. Its first two rows show learned endpoint geometries under eight \relax \relax \relax Demand Regularizers\relax ; the third row is a separate demonstration of designable two-dimensional glyph endpoints. The learned representations have no visually obvious collapse in these visualizations, although neither a low-dimensional projection nor control SR alone establishes a zero incidence of strict collapse.

\relax %
\relax \relax 

\subsection{\relax \relax \relax Latent-geometry \relax diagnostic \relax for the eight Demand Regularizers\relax }

The diagnostic uses \relax \relax \relax epoch 1 \relax INTACT checkpoints from three training seeds (0, 42, 3072); \relax \relax \relax official \relax success is the corresponding three-seed aggregate. For each task, 128 segments with eight positions yield 4,096 pooled observations. A reference subset of 1,024 fits whitening, and the remaining 3,072 observations are analyzed. With nonzero covariance spectrum, we adapt the entropy-based effective rank of \citet{roy2007effectiverank} to covariance eigenvalues:
\[
\relax \relax \relax \operatorname{erank}\relax (U)=\exp\!\left(-\sum_i p_i\log p_i\right),\qquad
p_i=\frac{\lambda_i(\Sigma_U)}{\operatorname{tr}(\Sigma_U)},
\]
where $0\log0=0$; raw effective rank is computed before whitening. We also report leading-eigenvalue share and average absolute excess kurtosis across random standardized projections:
\[
\eta_1=\frac{\lambda_{\max}(\Sigma_Z)}{\operatorname{tr}(\Sigma_Z)},\qquad
\bar\kappa=\langle|\operatorname{kurt}(Zv)|\rangle_v.
\]
Each checkpoint fits its own whitening transform, while observation indices, 32 random projection directions, and the split are shared. Four-task observations are pooled within each seed before reporting mean and sample standard deviation across seeds. Training uses within-task covariance, whereas pooled covariance obeys
\[
\operatorname{Cov}(Z)=\mathbb E_T[\operatorname{Cov}(Z\mid T)]
+\operatorname{Cov}_T(\mathbb E[Z\mid T]).
\]
Function readout uses 32 smooth random Fourier functions $f_j(x)=\sin(a_j\omega_j^\top x+b_j)$ in each checkpoint's whitened coordinates. Directions, frequencies, and phases are shared, while $x$ changes with the representation; these are isomorphic coordinate-defined function classes, not a shared physical-state target. For 128 fixed queries, $N=128$ or 1,024 training samples, and eight independent resamples, kNN with $n_{\rm nbr}=15$ and fixed-bandwidth RBF predictions are averaged before integrated squared bias is computed. Table~\ref{tab:app_f4_e1_unified} gathers the \relax \relax \relax epoch 1 geometry of the eight Demand Regularizers\relax , projected Fisher proxy, kNN bias, and control outcomes.

\subsection{Asymptotic endpoint optimality and finite-training attainability}

\href{https://arxiv.org/abs/2511.08544}{LeJEPA} analyzes linear and nonlinear probe risk under fixed covariance scale, smooth full-dimensional densities, and unknown downstream tasks. Its Fisher-information argument selects an \emph{isotropic Gaussian} endpoint, not an arbitrary anisotropic Gaussian. SADL does not dispute this endpoint choice; it asks whether the \relax \relax \relax encoder and projector \relax can produce and transmit enough sample variation to attain it within a training budget $T$. Conceptually, the evaluation levels can be separated as
\[
R_T(p;\mathcal F)\;\approx\;
R_{\mathrm{asymptotic\ probe}}(p)
+\Delta\relax \relax \relax _{\mathrm{optimization},T}\relax (p)
+\Delta_{\mathrm{task}}(p;\mathcal F),\qquad p\in\mathcal P_T.
\]
Here $\mathcal P_T$ is the set of endpoints attainable in $T$ steps and $\mathcal F$ is the task family. The expression distinguishes asymptotic probing, finite-training \relax \relax \relax optimization \relax effects, and task fit, with interactions assigned to the final term. Consequently, asymptotic, task-unknown Gaussian optimality does not automatically imply system-level optimality for a restricted \relax \relax \relax architecture\relax , finite budget, and specific control tasks.

We first calibrate synthetic distributions at matched total covariance, observation indices, smooth function family, and kNN-15 readout, then examine learned \relax \relax \relax epoch 1 \relax endpoints. For synthetic Student-$t_5$, the full-dimensional Fisher ratio has the closed-form benchmark
\[
\frac{J(t_\nu)}{J(\mathcal N(0,I_d))}
=\frac{\nu(\nu+d)}{(\nu-2)(\nu+d+2)}
=1.650\quad(d=192,\,\nu=5).
\]
Synthetic full Fisher ratios are $1.000$ for Gaussian, $1.650$ for Student-$t_5$, and $2.283$ for a two-scale full-dimensional Gaussian mixture, consistent with LeJEPA's theorem. Sphere lies on a singular lower-dimensional manifold outside its full-density assumptions.

The \relax \relax three-seed \relax \relax \relax epoch 1 experiment with eight Demand Regularizers \relax reveals a distinct finite-time perspective (Table~\ref{tab:app_f4_e1_unified}): the mean projected Fisher proxy is $1.220\pm.083$ for GMM-Ring versus $1.287\pm.053$ for SIGReg; finite-sample $b_{1024}$ is $.1590\pm.0025$ for Sphere versus $.1762\pm.0049$ for SIGReg, and is lower in all three paired seeds. Together with Table~\ref{tab:app_f_eight_official}, this shows that SADL's anti-collapse behavior and high control success do not depend on a Gaussian endpoint. LeJEPA's asymptotic Fisher conclusion remains applicable under its stated conditions. All geometry entries are \relax \relax \relax epoch 1 measurements; epoch 5 \relax appears only for control success.
\relax %
\relax \relax \begin{figure}[!htbp]
\noindent\begin{minipage}[c]{0.52\linewidth}
\relax \centering
\relax \includegraphics[width=\linewidth]{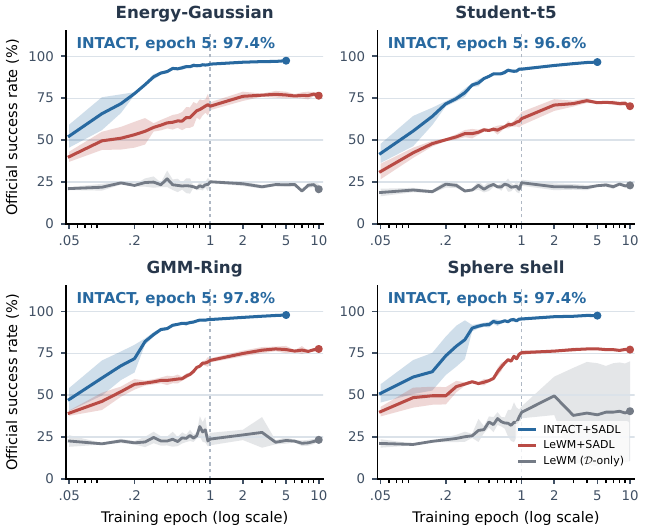}
\end{minipage}\hfill
\begin{minipage}[c]{0.45\linewidth}
\captionsetup{skip=0pt,justification=raggedright,singlelinecheck=false}
\relax \caption{\relax \relax \relax Four-task official success \relax across \relax \relax \relax four Demand Regularizers: \relax INTACT+\relax \relax \relax \textbf{SADL} (blue)\relax , \relax \relax \relax LeWM+\textbf{SADL} (red), \relax and \relax \relax \relax $\mathcal D$-only LeWM (gray). $\mathcal D$-only \relax training \relax is collapse-prone and unstable across \relax seeds\relax \relax \relax ; with \textbf{SADL}, diverse \relax endpoint \relax \relax \relax geometries yield stable\relax , \relax \relax \relax high-quality representations\relax .\relax \relax }
\relax \label{fig:geometry_sr}
\end{minipage}
\end{figure}

\begin{table}[!htbp]
\centering
\caption{Three-training-seed epoch 1 geometry, finite-sample readout, and official Direct control across eight Demand Regularizers. All models are jointly trained four-task INTACT+SADL; each seed pools 4,096 positions from the same fixed four-task observations. Geometry and bias are mean $\pm$ sample seed SD; epochs 1/5 SR also aggregates three training seeds. The two panels use identical row order. $10^3\mathrm{SW}^2$ denotes squared sliced-Wasserstein distance to a Gaussian reference, multiplied by $10^3$. Geometry and readout are measured at epoch 1; the epoch 5 column contains only control SR from epoch 5 checkpoints.}
\relax \label{tab:app_f4_e1_unified}
\begingroup\small\setlength{\tabcolsep}{2.3pt}\renewcommand{\arraystretch}{1.12}
\begin{tabular*}{\linewidth}{@{\extracolsep{\fill}}lccccc@{}}
\toprule
\multicolumn{6}{@{}l}{\relax \relax \relax \textit{A. Representation geometry: mean $\pm$ seed SD}\relax } \\
\midrule
\relax \relax \relax $\mathcal D$ \relax & \relax \relax \relax $\operatorname{erank}(H)$ \relax & \relax \relax \relax $\operatorname{erank}(Z)$ \relax & $\eta_1$ (\%) & $\bar\kappa$ & $\widehat J_{\rm proj}$ \\
\midrule
SIGReg & $118.85\pm1.19$ & $58.68\pm1.77$ & $3.45\pm0.22$ & $.573\pm.103$ & $1.287\pm.053$ \\
VISReg & $122.11\pm1.09$ & $60.44\pm1.26$ & $3.34\pm0.06$ & $.473\pm.074$ & $1.250\pm.077$ \\
QQWorld & $120.25\pm0.90$ & $58.82\pm0.76$ & $3.47\pm0.11$ & $.604\pm.063$ & $1.213\pm.004$ \\
Energy-Gaussian & $112.86\pm1.69$ & $48.10\pm0.56$ & $4.26\pm0.08$ & $.677\pm.098$ & $1.259\pm.077$ \\
\midrule
VIC moments & $106.17\pm1.53$ & $66.42\pm0.35$ & $3.01\pm0.22$ & $1.035\pm.158$ & $1.330\pm.066$ \\
Sphere & $102.61\pm1.29$ & $42.30\pm0.60$ & $4.96\pm0.15$ & $.920\pm.062$ & $1.251\pm.081$ \\
Student-$t_5$ & $89.91\pm3.03$ & $35.11\pm1.27$ & $6.38\pm0.47$ & $2.840\pm.319$ & $1.660\pm.135$ \\
GMM-Ring & $112.46\pm0.34$ & $48.29\pm0.57$ & $4.20\pm0.10$ & $.657\pm.066$ & $1.220\pm.083$ \\
\bottomrule
\end{tabular*}
\par\vspace{0.55\baselineskip}
\begin{tabular*}{\linewidth}{@{\extracolsep{\fill}}lccccc@{}}
\toprule
\multicolumn{6}{@{}l}{\textit{B. Distribution distance, function readout, and control: mean $\pm$ seed SD}} \\
\midrule
\relax \relax \relax $\mathcal D$ \relax & $10^3\mathrm{SW}^2$ & $b_{128}$ & $b_{1024}$ & \relax \relax \relax Ep.\,1 \relax SR (\%) & \relax \relax \relax Ep.\,5 \relax SR (\%) \\
\midrule
SIGReg & $5.19\pm0.97$ & $.2558\pm.0056$ & $.1762\pm.0049$ & $95.58\pm.00$ & $97.19\pm.51$ \\
VISReg & $4.02\pm0.46$ & $.2505\pm.0049$ & $.1716\pm.0044$ & $95.08\pm.17$ & $97.53\pm.25$ \\
QQWorld & $5.37\pm0.98$ & $.2552\pm.0045$ & $.1745\pm.0035$ & $95.81\pm.13$ & $97.56\pm.17$ \\
Energy-Gaussian & $5.88\pm0.58$ & $.2483\pm.0020$ & $.1683\pm.0012$ & $95.33\pm.63$ & $97.39\pm.42$ \\
\midrule
VIC moments & $9.43\pm1.58$ & $.2836\pm.0182$ & $.2287\pm.0167$ & $95.03\pm.99$ & $96.97\pm.19$ \\
Sphere & $8.35\pm0.77$ & $.2399\pm.0085$ & $.1590\pm.0025$ & $95.44\pm.82$ & $97.44\pm.10$ \\
Student-$t_5$ & $29.76\pm3.33$ & $.2804\pm.0052$ & $.2052\pm.0043$ & $92.33\pm1.04$ & $96.56\pm.55$ \\
GMM-Ring & $5.57\pm0.32$ & $.2403\pm.0061$ & $.1661\pm.0037$ & $95.11\pm.35$ & $97.81\pm.38$ \\
\bottomrule
\end{tabular*}
\endgroup
\end{table}

\relax %
\relax \relax \begin{table}[!htbp]
\relax \centering
\caption{LeWM+SADL \relax \relax \relax epoch 1 representation \relax diagnostics underlying the upper half of Table~\ref{tab:app_f_eight_official}. Each training seed pools 4,096 positions from the same fixed four-task observations as the INTACT probe in Table~\ref{tab:app_f4_e1_unified}. Entries are mean $\pm$ sample SD across three training seeds. $\widehat J_{\rm proj}$ is a projected Fisher proxy, not the full-dimensional asymptotic functional; $b_{1024}$ is finite-sample kNN integrated squared bias after per-checkpoint whitening.}
\label{tab:app_g_lewm_e1_geometry}
\begingroup\small\setlength{\tabcolsep}{2.3pt}\renewcommand{\arraystretch}{1.12}
\begin{tabular*}{\linewidth}{@{\extracolsep{\fill}}lcccc@{}}
\toprule
\relax \relax \relax $\mathcal D$ \relax & \relax \relax \relax $\operatorname{erank}(H)$ \relax & \relax \relax \relax $\operatorname{erank}(Z)$ \relax & $\widehat J_{\rm proj}$ & $b_{1024}$ \\
\midrule
SIGReg & $116.71\pm2.84$ & $48.84\pm2.23$ & $1.289\pm.049$ & $.1772\pm.0075$ \\
VISReg & $123.62\pm2.15$ & $55.24\pm2.65$ & $1.184\pm.020$ & $.1717\pm.0027$ \\
QQWorld & $114.11\pm9.39$ & $48.18\pm3.52$ & $1.179\pm.031$ & $.1699\pm.0020$ \\
Energy-Gaussian & $103.69\pm10.34$ & $40.40\pm2.31$ & $1.161\pm.037$ & $.1712\pm.0019$ \\
\midrule
VIC moments & $105.05\pm2.75$ & $64.58\pm2.00$ & $1.308\pm.117$ & $.2228\pm.0157$ \\
Sphere & $95.48\pm2.14$ & $33.57\pm.93$ & $1.189\pm.025$ & $.1660\pm.0022$ \\
Student-$t_5$ & $84.26\pm1.06$ & $30.48\pm1.29$ & $1.591\pm.093$ & $.2034\pm.0184$ \\
GMM-Ring & $106.73\pm2.29$ & $39.83\pm1.86$ & $1.232\pm.063$ & $.1756\pm.0042$ \\
\bottomrule
\end{tabular*}
\endgroup
\end{table}

\relax %
\relax \relax \begin{table}[!htbp]
\relax \centering
\caption{LeWM \relax \relax \relax epoch 1 $\mathcal D$-only \relax versus +SADL under the same \relax \relax \relax official \relax Pure-CEM protocol. Both configurations use a Flash encoder, MATH predictor, shared four-task training, and the same \relax \relax \relax $\lambda_D$ \relax for each \relax \relax \relax Demand Regularizer\relax . Pure-CEM uses 300 samples, 30 iterations, $T_H=5$, and $R_H=5$ without actor warm-start. Task columns show \relax \relax \relax $\mathcal D$-only \relax success rates (\%); macro columns are mean $\pm$ sample SD across three training seeds, each averaged over three evaluation seeds of 100 episodes per task. $\Delta$ is the difference of the two macro means in percentage points.}
\label{tab:app_g_lewm_e1_baseline}
\begingroup\footnotesize\setlength{\tabcolsep}{2.4pt}\renewcommand{\arraystretch}{1.10}
\begin{tabular*}{\linewidth}{@{\extracolsep{\fill}}lccccccc@{}}
\toprule
\relax \relax \relax $\mathcal D$ \relax & PushT$_0$ & Cube$_0$ & Reacher$_0$ & TwoRoom$_0$ & Macro$_0$ & Macro$_{\rm SADL}$ & $\Delta$ pp \\
\midrule
SIGReg & 3.44 & 49.89 & 20.00 & 36.44 & $27.44\pm3.44$ & $71.56\pm1.27$ & $+44.12$ \\
VISReg & 53.00 & 55.11 & 27.33 & 55.33 & $47.69\pm5.14$ & $71.58\pm4.67$ & $+23.89$ \\
QQWorld & 19.22 & 52.56 & 18.78 & 40.33 & $32.72\pm11.23$ & $74.36\pm2.90$ & $+41.64$ \\
Energy-Gaussian & 3.78 & 45.11 & 14.78 & 36.89 & $25.14\pm0.67$ & $70.25\pm2.35$ & $+45.11$ \\
\midrule
VIC moments & 5.67 & 47.44 & 15.56 & 29.56 & $24.56\pm4.21$ & $67.53\pm6.09$ & $+42.97$ \\
Sphere & 34.33 & 54.33 & 23.22 & 47.11 & $39.75\pm3.54$ & $75.33\pm1.36$ & $+35.58$ \\
Student-$t_5$ & 3.11 & 45.78 & 13.44 & 36.00 & $24.58\pm1.78$ & $62.81\pm3.99$ & $+38.23$ \\
GMM-Ring & 3.00 & 43.44 & 12.78 & 36.00 & $23.81\pm3.71$ & $70.75\pm1.23$ & $+46.94$ \\
\bottomrule
\end{tabular*}
\endgroup
\end{table}

\relax %
\relax \section{Image Classification Evaluation}
\label{app:ahead_visual}

We test whether the same \relax \relax \relax supply capacity \relax correction helps image representations across long-tailed natural images (ImageNet-LT), galaxies (Galaxy10), small natural images (CIFAR-10), and the Imagenette subset of ImageNet. Each comparison pairs the \relax \relax \relax $\mathcal D$-only \relax baseline with the same model plus SADL, holding the data, architecture, optimization, and evaluation protocol fixed. SADL \relax \relax \relax computes $k_{\mathrm{exc}}$ and $q_{\mathrm{exc}}$ \relax from the mean representation of each image's augmented views (the \emph{image-mean} variant) in all four datasets. Table~\ref{tab:app_visual_full} reports fixed-epoch classification accuracy: an online head for ImageNet-LT, Galaxy10, and Imagenette, and a held-out ridge readout for CIFAR-10. Each mean and standard deviation is over independent training seeds; the last column counts seeds improved by SADL.

\relax %
\relax \relax \begin{figure}[tb]
\relax \centering
\relax %
\relax \relax \includegraphics[width=\linewidth]{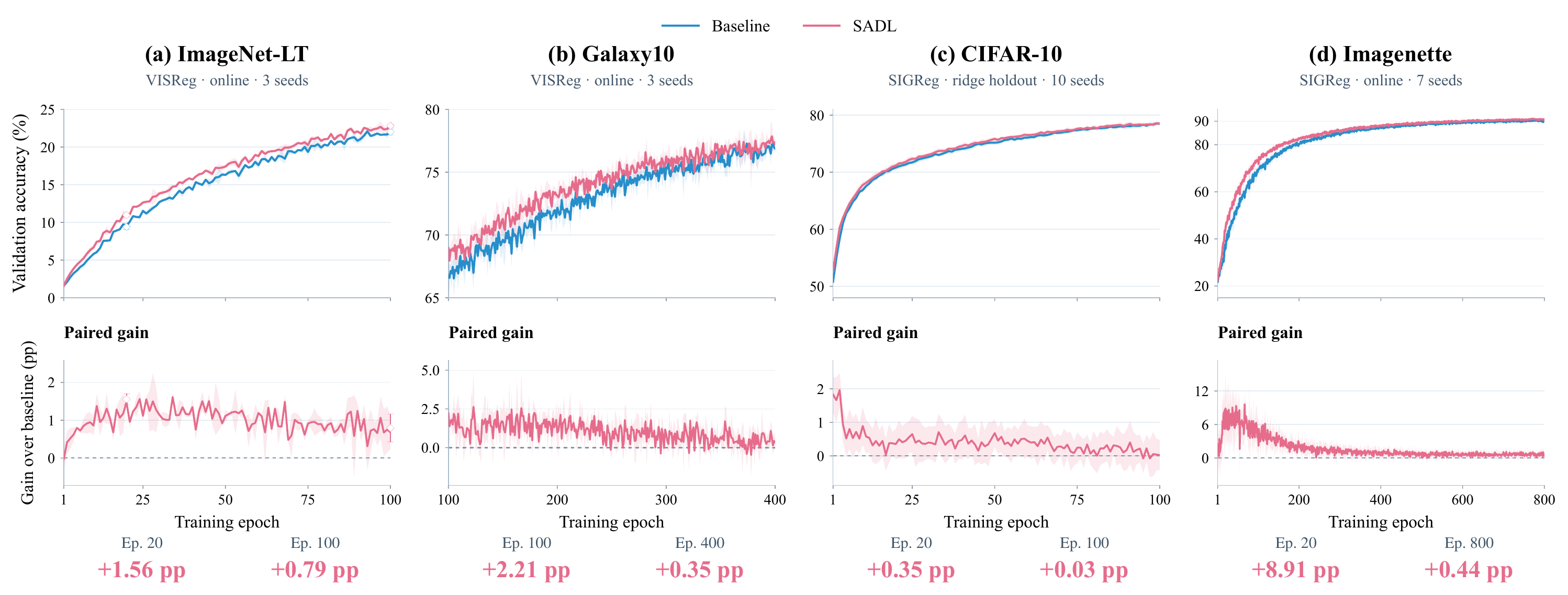}
\relax \caption{Training trajectories and paired accuracy gains across four image-classification datasets. Top: classification accuracy; bottom: SADL minus the paired baseline, in percentage points. Lines and bands show means and standard deviations over paired training seeds (3, 3, 10, and 7, respectively). ImageNet-LT, Galaxy10, and Imagenette use online heads; CIFAR-10 uses a held-out ridge readout.}
\label{fig:image_training_advantage}
\end{figure}

\relax %
\relax \relax \begin{table}[!htbp]
\relax \centering
\caption{Image-classification accuracy (\%). Baseline and SADL use matched training seeds and the same \relax \relax \relax Demand Regularizer\relax . $\Delta$ is the mean paired gain in percentage points; ``wins'' counts positive seed-wise differences.}
\label{tab:app_visual_full}
\SADLTableStyle
\setlength{\tabcolsep}{3pt}
\begin{tabularx}{\linewidth}{@{}p{0.16\linewidth}p{0.12\linewidth}cYYrc@{}}
\toprule
Dataset ($n$) & \relax \relax \relax $\mathcal D$ \relax & Epoch & Baseline & +SADL & $\Delta$ & Wins \\
\midrule
ImageNet-LT (3) & VISReg & 100 & $22.02\pm0.41$ & $\mathbf{22.81\pm0.20}$ & $+0.794$ & 3/3 \\
Galaxy10 (3) & VISReg & 400 & $76.89\pm0.45$ & $\mathbf{77.23\pm0.53}$ & $+0.348$ & 3/3 \\
\addlinespace[2pt]
CIFAR-10 (10) & SIGReg & 1 & $50.81\pm0.73$ & $\mathbf{52.64\pm0.80}$ & $+1.830$ & 10/10 \\
 & & 10 & $66.87\pm0.40$ & $\mathbf{67.76\pm0.67}$ & $+0.894$ & 9/10 \\
 & & 50 & $75.16\pm0.36$ & $\mathbf{75.84\pm0.36}$ & $+0.680$ & 9/10 \\
 & & 100 & $78.50\pm0.30$ & $78.52\pm0.32$ & $+0.026$ & 4/10 \\
\addlinespace[2pt]
Imagenette (7) & SIGReg & 5 & $24.23\pm1.49$ & $\mathbf{27.68\pm1.43}$ & $+3.450$ & 7/7 \\
 & & 20 & $35.60\pm4.48$ & $\mathbf{44.51\pm3.13}$ & $+8.906$ & 7/7 \\
 & & 100 & $69.34\pm1.99$ & $\mathbf{73.15\pm1.89}$ & $+3.807$ & 7/7 \\
 & & 800 & $90.25\pm0.17$ & $\mathbf{90.69\pm0.31}$ & $+0.444$ & 6/7 \\
\bottomrule
\end{tabularx}
\end{table}

\paragraph{ImageNet-LT and Galaxy10.}
On ImageNet-LT, ViT-S/8 with VISReg is trained for 100 epochs using a global batch of 256, four $128\times128$ global crops, and 4,096 projection directions. At \relax \relax \relax epoch 100\relax , SADL improves the matched three-seed online accuracy by $0.794$ points. Galaxy10 uses the same backbone and \relax \relax \relax Demand Regularizer\relax ; its three-seed gain is $2.207$ points at \relax \relax \relax epoch 100 \relax and remains positive at \relax \relax \relax epoch 400 \relax (Table~\ref{tab:app_visual_full}). The fixed \relax \relax \relax epoch 400 \relax comparison uses each arm's own \relax \relax \relax epoch 100 \relax model and optimizer state.

\paragraph{CIFAR-10 and Imagenette.}
CIFAR-10 pairs a ResNet-18 SIGReg baseline with image-mean SADL and the same held-out ridge readout over ten seeds. The \relax \relax \relax epoch 1 \relax gain is $1.830$ points, and nine of ten seeds still improve at \relax \relax \relax epoch 50; by epoch 100 \relax the mean accuracies are nearly equal. The Imagenette baseline follows the LeJEPA ViT-S/8 SIGReg recipe~\citep{balestriero2025lejepa}, with four global views and a global batch of 256. Seven paired seeds show the largest gain at \relax \relax \relax epoch 20 \relax ($+8.906$ points), and the gain remains positive at \relax \relax \relax epoch 800 \relax ($+0.444$ points). These results show the clearest and most consistent advantage in the early training regime, with improvements also present at the fixed final checkpoint on ImageNet-LT, Galaxy10, and Imagenette.

\paragraph{\relax \relax \relax Image-mean statistics\relax .}
\relax SADL computes $k_{\mathrm{exc}}$ and $q_{\mathrm{exc}}$ from the mean representation of each image's augmented views. \relax With $n$ image means and encoder width \relax \relax \relax $d_H$\relax , the covariance rank is at most \relax %

\relax \relax $m_H=\min(n-1,d_H)$, so the normalization of Equation~\ref{eq:ahead_excitation} is active: for \relax ImageNet-LT, $n=256$, \relax \relax \relax $d_H=384$\relax , and $\gamma=9$, giving \relax \relax \relax $m_H=255$. The demand capacity and the Supply Regularizer are otherwise exactly those of Section~\ref{sec:theory}, with $\epsilon=10^{-6}$ and the demand detached. \relax In the matched five-point interpolation sweep, $\alpha=0.5$ gives the highest \relax \relax \relax epoch 100 \relax online accuracy.\par\paragraph{Which \relax \relax \relax supply term \relax limits the image \relax \relax \relax representation\relax ?} \relax \relax \relax In the \relax ImageNet-LT \relax \relax \relax epoch 100 \relax view-residual replay\relax \relax \relax , transmission can become the smaller supply term ($k_{\mathrm{tr}}<q<k_{\mathrm{exc}}$\relax ; Figure~\ref{fig:app_e_roles}). This \relax \relax motivates retaining both \relax \relax \relax terms \relax in $k$; the classification comparisons above use the image-mean protocol.

\relax %
\relax \relax \section{Matched Backends, Demand, Direct Regularizers, and Sensitivity}
\label{app:rebuttal}

Control runs use four-task joint training with three training seeds
(0/42/3072); official evaluation uses seeds 0/1/42 and 100 episodes
per task and evaluation seed. Entries are mean $\pm$ sample SD over
training seeds unless stated otherwise. Original- and optimized-backend
comparisons are reported separately below. Encoder--endpoint
diagnostics are in Appendix~\ref{sec:app_collapse},
transmission ablations in Appendix~\ref{sec:app_channel_roles},
image time-to-target in Appendix~\ref{app:ahead_visual}, and
training cost in Appendix~\ref{app:ahead_cost}.

\begin{table}[!htbp]
\centering\small
\caption{Run batches used by the appendix comparisons. Rows from different
batches are not pooled; each comparison is interpreted within its stated batch.}
\label{tab:app_run_batches}
\begin{tabularx}{\linewidth}{@{}p{.19\linewidth}p{.24\linewidth}p{.10\linewidth}X@{}}
\toprule
Batch & Backend and path & $\lambda_D$ & Results using the batch \\
\midrule
Matched backend & MATH/Flash, original INTACT backend & .03 & Table~\ref{tab:reb_backend}; direct regularizer scan \\
Fast ablation & Flash, optimized attention backend & .03 & transmission sweep and sensitivity tables \\
Ten-seed audit & Flash/MATH LeWM diagnostic recipe & .03 & paired rank-drop, $H/Z$, and temporal diagnostics \\
Mechanism audit & Flash, mechanism-audit configuration & .03 & tr-only, gate and potential rows of Table~\ref{tab:robust}; Table~\ref{tab:reb_mechanism_scan} \\
Historical control & MATH $\mathcal D$-only, Flash +\textbf{SADL} & .03 & Appendix~\ref{app:ahead_control}; long-training reference \\
\bottomrule
\end{tabularx}
\end{table}

\subsection{Matched Attention Backend and Training Budget}
\label{sec:reb_backend}

We matched the attention backend and number of updates to check
whether the gains survive under the same training conditions.
Within each model, the four cells share the code path, data order
and evaluation, and only the encoder's attention backend and the
presence of SADL change. INTACT uses its original attention backend;
LeWM uses the ten-seed-audit recipe, with the three seeds listed
above. Both methods receive the same number of updates. INTACT
uses Direct control and LeWM uses Pure-CEM.

\begin{table}[H]
\centering\small
\caption{Four-task macro success (\%), SIGReg as the Demand
Regularizer, three training seeds. Official evaluation uses seeds
0, 1, and 42 and 100 episodes per task and evaluation seed.}
\label{tab:reb_backend}
\setlength{\tabcolsep}{3pt}
\begin{tabular*}{\linewidth}{@{\extracolsep{\fill}}llcccc@{}}
\toprule
Model & Backend & $\mathcal D$-only Ep.\,1 & \textbf{SADL} Ep.\,1 & $\mathcal D$-only Ep.\,2 & \textbf{SADL} Ep.\,2 \\
\midrule
INTACT & MATH & $40.5\pm8.5$ & $94.9\pm0.4$ & $70.0\pm6.1$ & $96.3\pm0.6$ \\
 & Flash & $45.8\pm2.6$ & $95.4\pm0.2$ & $66.1\pm34.5$ & $96.3\pm0.1$ \\
LeWM & MATH & $30.9\pm0.8$ & $63.9\pm1.2$ & $46.0\pm6.0$ & $70.4\pm1.6$ \\
 & Flash & $32.1\pm4.9$ & $63.4\pm5.7$ & $31.4\pm15.8$ & $69.1\pm2.9$ \\
\bottomrule
\end{tabular*}
\end{table}

\begin{table}[H]
\centering
\footnotesize
\caption{Supplemental LeWM MATH, +\textbf{SADL} fixed-probe record. This
run uses three training seeds (0, 42, 3072), whereas the other LeWM rows
in the collapse audit use ten seeds. The row is the MATH-encoder companion
to the Flash+SADL backend comparison.}
\label{tab:reb_lewm_math_sadl_probe}
\setlength{\tabcolsep}{3pt}
\begin{tabular*}{\linewidth}{@{\extracolsep{\fill}}lccccccccc@{}}
\toprule
Setting & SR (Ep.\,1) & N/O/D/C & $f_H$ & $f_Z$ & $\operatorname{erank}(H)$ & $\operatorname{erank}(Z)$ & $\overline{\cos}_H$ & $\overline{\cos}_Z$ & $L_{\rm pred}$ \\
\midrule
LeWM, MATH, +\textbf{SADL} & $63.9\pm1.2$ & $3/0/0/0$ & $.985$ & $.991$ & $51.7$ & $51.7$ & $.012$ & $.007$ & $.019$ \\
\bottomrule
\end{tabular*}
\end{table}

In every model and backend, SADL at epoch 1 already exceeds
$\mathcal D$-only training at epoch 2, reaching a higher success
with half the updates (5,695 vs. 11,390 for INTACT). At epoch 2
the gap is still $24.4$ to $37.7$ points. The historical control
results in Appendix~\ref{app:ahead_control} are separate runs.
For INTACT, SADL's epoch 1 success with either backend
($94.9$--$95.4\%$) also exceeds the historical $\mathcal D$-only
MATH result at epoch 10 ($91.58\%$;
Table~\ref{tab:app_e_official_joint_intact}), with one tenth of
the updates. On the same machine, the SADL+Flash run reaches
$95.4\%$ in about $4{,}400$ seconds for one epoch, whereas the
$\mathcal D$-only MATH run reaches only $70.0\%$ after $16{,}820$
seconds for two epochs; the historical epoch-10 result would take
about $84{,}100$ seconds by the same timing, roughly $19$ times as
long. These are checkpoint-budget estimates, not measured times to
first reach a common target. Appendix~\ref{app:ahead_cost} gives the
portable-release timing table, separate from these historical
checkpoint-budget estimates.

\begin{table}[H]
\centering
\footnotesize
\caption{Fixed-probe transmission margins $k_{\rm tr}-k_{\rm exc}$
for SADL and exc-only across Supply Regularizer weights. These runs use
INTACT with the optimized attention backend, SIGReg weight $0.03$, the default BatchNorm
projector, and three training seeds (0, 42, 3072). Entries are
SADL / exc-only; each margin is measured on the held-out probe.}
\label{tab:reb_margin}
\setlength{\tabcolsep}{4pt}
\begin{tabular*}{\linewidth}{@{\extracolsep{\fill}}lcccc@{}}
\toprule
$\lambda_S$ & Ep.\,1 Reacher & Ep.\,1 TwoRoom & Ep.\,2 Reacher & Ep.\,2 TwoRoom \\
\midrule
0.01 & $+.141 / +.142$ & $+.115 / +.115$ & $+.119 / +.120$ & $+.089 / +.093$ \\
0.1  & $+.064 / +.056$ & $+.032 / +.017$ & $+.060 / +.041$ & $+.032 / +.005$ \\
0.3  & $+.059 / +.044$ & $+.030 / +.010$ & $+.056 / +.031$ & $+.028 / -.003$ \\
1.0  & $+.057 / +.038$ & $+.029 / +.002$ & $+.055 / +.028$ & $+.028 / -.006$ \\
\bottomrule
\end{tabular*}
\end{table}

The transmission margin is positive for both methods at most weights.
At $\lambda_S=0.3$ and $1$, the exc-only TwoRoom margin becomes slightly
negative at epoch 2, while SADL remains positive. This is the same
transmission branch that is selected on a larger fraction of training
minibatches as the Supply Regularizer weight increases.

\paragraph{When transmission is the smaller supply term.} Table~\ref{tab:reb_tr} counts, on the training minibatches of SADL runs with $\lambda_S=0.3$, how often $k_{\rm tr}<k_{\rm exc}$. In the matched exc-only runs, $k_{\rm tr}-k_{\rm exc}$ on TwoRoom falls to $-0.048$ by the end of epoch 2, while SADL keeps it at $0.000$. Across seeds 0/42/3072 and two reruns with $\lambda_S\geq0.3$, SADL has the higher transmission margin in every run.

\begin{table}[!htbp]
\centering
\caption{Fraction of training minibatches on which SADL selects $k_{\rm tr}$ ($\lambda_S=0.3$; mean over seeds 0/42/3072).}
\label{tab:reb_tr}
\small
\begin{tabular*}{\linewidth}{@{\extracolsep{\fill}}lcccc@{}}
\toprule
Stage & PushT & Cube & Reacher & TwoRoom \\
\midrule
Ep.\,1, steps 0--2k & $0\%$ & $0\%$ & $0\%$ & $0\%$ \\
Ep.\,1, steps 4k--6k & $14\%$ & $4\%$ & $25\%$ & $45\%$ \\
Ep.\,2, steps 4k--6k & $27\%$ & $16\%$ & $44\%$ & $63\%$ \\
\bottomrule
\end{tabular*}
\end{table}

\subsection{Constant versus Adaptive Demand}
\label{sec:reb_constant}

To isolate the role of adaptive demand, everything except the
demand is identical to SADL: the same supply terms, the same
one-sided Supply Regularizer and the same weight; only the demand
capacity $q$ is replaced by a constant. For control, INTACT uses
Flash and the fast training path.
The control table is reported in Table~\ref{tab:reb_constq} in the
main text. The image results are reported separately below because
their evaluation protocols and training-seed counts differ.

\begin{table}[!htbp]
\centering\small
\caption{Image classification accuracy (\%), mean $\pm$ SD.
Imagenette uses the online probe at epoch 800 (3 seeds),
ImageNet-LT uses best online accuracy within 100 epochs (3 seeds),
and CIFAR-10 uses validation accuracy at epoch 100 (5 seeds).}
\label{tab:reb_constq_images}
\begin{tabular*}{\linewidth}{@{\extracolsep{\fill}}lccc@{}}
\toprule
Demand & Imagenette & ImageNet-LT & CIFAR-10 \\
\midrule
Baseline (no SADL) & $90.19\pm0.26$ & $22.51\pm0.46$ & $78.38\pm0.16$ \\
Constant $q=0.3$ & $90.32\pm0.03$ & $22.20\pm0.15$ & $78.91\pm0.35$ \\
Constant $q=0.5$ & $90.49\pm0.23$ & $22.60\pm0.19$ & $79.06\pm0.26$ \\
Constant $q=0.68$ & $90.61\pm0.43$ & $23.00\pm0.06$ & $78.77\pm0.42$ \\
Constant $q=0.7$ & $90.57\pm0.23$ & $22.74\pm0.09$ & $78.90\pm0.55$ \\
Constant $q=0.9$ & $90.66\pm0.46$ & $22.68\pm0.08$ & $78.75\pm0.38$ \\
Adaptive (default, not tuned) & $91.04\pm0.22$ & $22.92\pm0.07$ & $78.75\pm0.47$ \\
\bottomrule
\end{tabular*}
\end{table}

A tuned constant can match adaptive demand, but the best constant
ranges from $0.5$ to $0.9$ across settings: $0.7$ for SIGReg control,
$0.9$ for VISReg control, $0.9$ on Imagenette, $0.68$ on ImageNet-LT,
and $0.5$ on CIFAR-10. A poor choice gives no gain, as $q=0.3$ on ImageNet-LT ($22.20\%$ vs. $22.51\%$). Adaptive demand needs no separate
threshold search; we do not claim that it always exceeds the best
tuned constant.

\FloatBarrier
\subsection{Direct Regularizers on the Encoder Representation}
\label{sec:reb_direct}

We compared SADL with three regularizers applied directly to $H$,
sweeping the weight of each. The first is the MCR$^2$ expansion
term, maximized with a fixed weight on unit-normalized, uncentered
rows of $H$. The second is the always-on coding rate, which maximizes the
same $F_{\rm exc}(H)$ as SADL with a fixed weight, without demand
and without the one-sided gate. The third is VCReg applied to $H$.
All runs use INTACT with Flash, the original attention backend, SIGReg
with weight $0.03$, and three training seeds. Table~\ref{tab:reb_direct}
gives the complete epoch-1 sweep for this comparison. The separate
mechanism audit in Table~\ref{tab:reb_mechanism_scan} reports the
exc-only, tr-only, two-sided, one-sided-linear, and $\mathcal S$-only
variants with the four-way outcome of Section~\ref{sec:experiments}.

\begin{table}[!htbp]
\centering\small
\caption{Epoch 1 four-task macro success (\%) for regularizers on
$H$, by weight. Parentheses give the worst seed. Outcome counts for the complete scans are given in
Table~\ref{tab:reb_mechanism_scan}.}
\label{tab:reb_direct}
\begin{tabular*}{\linewidth}{@{\extracolsep{\fill}}llc@{}}
\toprule
Term on $H$ & Weight & Ep.\,1 SR \\
\midrule
SADL, $\lambda_S\mathcal S$ & 0.01 (default) & $95.42\pm0.17$ (95.25) \\
 & 0.03 & $94.72\pm0.34$ (94.33) \\
 & 0.1 & $93.64\pm0.46$ (93.17) \\
 & 0.3 & $93.75\pm0.37$ (93.50) \\
 & 1 & $93.31\pm1.93$ (91.17) \\
\addlinespace
MCR$^2$ expansion & 0.03 & $63.47\pm36.69$ (23.17) \\
 & 0.1 & $64.69\pm44.64$ (13.33) \\
 & 0.2 & $72.36\pm40.01$ (26.17) \\
 & 0.3 & $94.75\pm0.43$ (94.50) \\
 & 0.5 & $93.61\pm1.28$ (92.75) \\
 & 1 & $93.47\pm0.54$ (93.08) \\
 & 3 & $90.72\pm1.18$ (89.67) \\
 & 10 & $27.30\pm0.46$ (27.00) \\
\addlinespace
Always-on coding rate & 0.03 & $61.95\pm34.93$ (25.00) \\
 & 0.1 & $95.28\pm0.62$ (94.92) \\
 & 0.3 & $94.45\pm0.21$ (94.25) \\
 & 1 & $91.47\pm0.53$ (91.17) \\
 & 3 & $58.83\pm29.23$ (28.58) \\
\addlinespace
VCReg on $H$ & 0.01 & $92.33\pm4.28$ (87.42) \\
 & 0.03 & $94.56\pm0.49$ (94.00) \\
 & 0.1 & $94.75\pm1.09$ (93.58) \\
 & 0.3 & $94.19\pm0.73$ (93.58) \\
 & 1 & $91.72\pm1.22$ (90.58) \\
\bottomrule
\end{tabular*}
\end{table}

At their best epoch-1 weights, \textbf{SADL} reaches $95.42\%$ and the three direct terms $94.75\%$ to $95.28\%$.
Regularizing $H$ directly works when the weight is right, which
supports the diagnosis that the missing correction lies upstream of
$Z$. The difference is how much the result depends on the weight.
Across SADL's 100-fold range, the worst observed seed is $91.17\%$;
MCR$^2$ and the always-on coding rate show low-success seeds at the ends of
their ranges, while VCReg remains above $87.42\%$ in this scan.

The default $\lambda_S=0.01$ used throughout the control experiments
is also the best of the five SADL weights in this sweep. On the same
machine, SADL adds $0.8\%$ to step time with MATH and $0.5\%$ with
Flash ($2\%$ in the faster ablation path), without a measurable
peak-memory change.

The always-on coding rate uses exactly SADL's excitation function, so
this pair isolates the demand and the one-sided gate. The correction
strength in Equation~\ref{eq:ahead_adaptive_gradient} is large when
supply is far below demand and falls to zero once supply reaches
it. A fixed-weight term pushes with the same weight in both states.
The matched one-sided versus two-sided comparison is in
Table~\ref{tab:app_potential}: at Ep.\,2, CLEAR-LeWM Moderate success is
$65.50\pm0.85\%$ versus $61.00\pm4.86\%$ for LeWM and
$97.14\pm0.17\%$ versus $93.97\pm5.76\%$ for INTACT.

\begin{table}[!htbp]
\centering\small
\caption{Additional direct-regularizer and mechanism scans. As in Table~\ref{tab:robust}, best is the best per-weight mean and
worst is the lowest single seed. N/O/D/C
counts are aggregated over the seeds and weights in each scan and use
the Ep.\,2 four-way outcome definition.}
\label{tab:reb_mechanism_scan}
\begin{tabular*}{\linewidth}{@{\extracolsep{\fill}}lccc@{}}
\toprule
Variant & Weight or audit setting & Ep.\,1 SR (best / worst) & N/O/D/C \\
\midrule
SADL & $\lambda_S=0.01$--$1$ & $95.42/91.17$ & $15/0/0/0$ \\
exc-only & $\lambda_S=0.3,1$ & $93.08/91.17$ & $6/0/0/0$ \\
tr-only & mechanism audit, $\lambda_S=0.01$ & $47.94/23.33$ & $1/0/0/2$ \\
Two-sided $u^2$ (no gate) & mechanism audit & $95.64/88.42$ & $3/0/0/0$ \\
One-sided linear $[u]_+$ & mechanism audit & $95.31/94.75$ & $3/0/0/0$ \\
$\mathcal S$-only & mechanism audit & $45.69/33.33$ & $1/0/2/0$ \\
Always-on coding rate & full scan & -- & $11/2/1/1$ \\
MCR$^2$ expansion & full scan & -- & $18/5/1/0$ \\
VCReg on $H$ & full scan & -- & $15/0/0/0$ \\
\bottomrule
\end{tabular*}
\end{table}

\subsection{Sensitivity of Performance and Bottleneck Diagnosis}
\label{sec:reb_sensitivity}

We varied the ridge coefficient $\rho$ over two orders of
magnitude and the SNR coefficient $\gamma$ over one. We did this
at the default weight and at $\lambda_S=0.3$, where the
transmission term is active (INTACT, SIGReg, optimized attention backend, three
training seeds). For each setting we report success, how often
$k_{\rm tr}$ is selected, and the demand capacity reached.

\begin{table}[!htbp]
\centering\footnotesize
\caption{Sensitivity at the default weight $\lambda_S=0.01$.
Selection entries are fractions, not percentages. At
$\rho=10^{-2}$, TwoRoom selection fractions are 0.76/0.81 at
epochs 1/2. The last column gives the range across four tasks.}
\label{tab:reb_sensitivity_default}
\setlength{\tabcolsep}{3pt}
\begin{tabular*}{\linewidth}{@{\extracolsep{\fill}}lcccc@{}}
\toprule
Setting & Ep.\,1 SR & Ep.\,2 SR & $k_{\rm tr}$ selected, Ep.\,1/2 & $q$, Ep.\,2 \\
\midrule
Default ($\rho=10^{-3},\gamma=9$) & $94.9\pm0.8$ & $96.5\pm0.2$ & 0 / 0 & .736--.752 \\
$\rho=10^{-4}$ & $95.0\pm0.2$ & $96.5\pm0.1$ & 0 / 0 & .735--.751 \\
$\rho=10^{-2}$ & $95.1\pm1.1$ & $96.1\pm0.5$ & .37 / .43 & .733--.753 \\
$\gamma=3$ & $94.9\pm0.2$ & $96.3\pm0.4$ & 0 / 0 & .771--.781 \\
$\gamma=27$ & $95.3\pm0.4$ & $96.6\pm0.6$ & 0 / 0 & .718--.740 \\
\bottomrule
\end{tabular*}
\end{table}

\begin{table}[!htbp]
\centering\small
\caption{Sensitivity to $\rho$ at $\lambda_S=0.3$, where the
transmission term is active. Selection entries are fractions.}
\label{tab:reb_sensitivity_active}
\begin{tabular*}{\linewidth}{@{\extracolsep{\fill}}lccc@{}}
\toprule
Ridge $\rho$ & Ep.\,1 SR & Ep.\,2 SR & $k_{\rm tr}$ selected, Ep.\,1/2 \\
\midrule
$10^{-4}$ & $93.56\pm0.61$ & $95.67\pm0.63$ & 0 / .01 \\
$10^{-3}$ (default) & $93.11\pm0.67$ & $95.81\pm0.32$ & .33 / .35 \\
$10^{-2}$ & $93.47\pm1.01$ & $95.72\pm0.41$ & 1.00 / .97 \\
\bottomrule
\end{tabular*}
\end{table}

Final success changes little with either coefficient: across these
eight settings, epoch 2 success stays between $95.67\%$ and
$96.6\%$. The bottleneck diagnosis does depend on $\rho$.
At $\lambda_S=0.3$, $k_{\rm tr}$ is selected on almost no
minibatches, on a third of them, or on almost all of them for
$\rho=10^{-4}$, $10^{-3}$, and $10^{-2}$. The ridge sets how
small the variance of a direction can be before it stops counting
as transmitted, so a larger $\rho$ makes $F_{\rm tr}$ stricter.
The direction of the update therefore changes with $\rho$,
while final success changes little. These are retrained
configurations.

\subsection{Finite-Batch Capacity Normalization}
\label{sec:reb_normalization}

The capacity normalization matters when $N-1$ is smaller than the
feature dimension. The original implementation divided the excitation function
by $d_U\log(1+\gamma)$. At most $N-1$ directions can vary in a
batch of $N$ samples, so we use
$m_U\log(1+\gamma d_U/m_U)$ with $m_U=\min(d_U,N-1)$, and the
analogous dimension $m=\min(d_H,d_Z,N-1)$ for $F_{\rm tr}$.
In control, $N-1$ exceeds the feature dimensions, so the two
normalizations are identical and none of the control results
change. In the image-mean variant they differ: on ImageNet-LT,
for example, a batch has 256 image means and $d_H=384$.
Table~\ref{tab:reb_normalization} compares the paired gains over
the baseline under both normalizations.

\begin{table}[!htbp]
\centering\footnotesize
\caption{Paired gain of SADL over the baseline (accuracy points)
under the two normalizations. Parentheses count seeds with a
positive gain where available. ImageNet-LT uses the online probe.}
\label{tab:reb_normalization}
\setlength{\tabcolsep}{3pt}
\begin{tabularx}{\linewidth}{@{}p{.32\linewidth}YYYY@{}}
\toprule
Normalization & ImageNet-LT & Galaxy10, epoch 400 & CIFAR-10 & Imagenette, epoch 800 \\
\midrule
Original: $d_U\log(1+\gamma)$ & +.794 (3/3) & +.348 (3/3) & +.026 (4/10) & +.444 (6/7) \\
Finite batch: $m_U\log(1+\gamma d_U/m_U)$ & +.694 & +.470 (2/3) & +.186 (7/10) & +.401 (6/7) \\
\bottomrule
\end{tabularx}
\end{table}

The mean gain keeps its sign on all four datasets, and its size
changes by at most $0.16$ points. The fixed-epoch image table
(Table~\ref{tab:app_visual_full}) retains the original
normalization; Table~\ref{tab:reb_normalization} reports the
normalization comparison separately. For CIFAR-10, nine of ten seeds
are ahead by epoch 50 ($+1.830$, $+0.894$, and $+0.680$ points at
epochs 1, 10, and 50; Table~\ref{tab:app_visual_full}), and the
finite-batch normalization leaves a $+0.19$-point gain at epoch 100
on seven of ten seeds.

\paragraph{Time to target accuracy.} Table~\ref{tab:reb_ttt} gives the number of updates needed to reach a fraction of the final baseline accuracy (median over seeds). On CIFAR-10, \textbf{SADL} reaches the $80\%$ target with about $17\%$ fewer updates but does not reach the final baseline accuracy sooner in these five-seed runs.

\begin{table}[!htbp]
\centering
\caption{Updates (thousands) to reach a given fraction of the final baseline accuracy (median over seeds: 3 on Imagenette and ImageNet-LT, 5 on CIFAR-10), and the speed-up of \textbf{SADL} (baseline updates / \textbf{SADL} updates). One epoch is 36 updates on Imagenette, 452 on ImageNet-LT, and 195 on CIFAR-10.}
\label{tab:reb_ttt}
\scriptsize
\setlength{\tabcolsep}{4pt}
\begin{tabular*}{\linewidth}{@{\extracolsep{\fill}}llcccc@{}}
\toprule
Dataset & & $50\%$ & $80\%$ & $90\%$ & $100\%$ \\
\midrule
Imagenette & baseline & $0.9$ & $3.8$ & $7.0$ & $23.2$ \\
 & \textbf{SADL} & $\mathbf{0.8}$ & $\mathbf{3.2}$ & $\mathbf{5.8}$ & $\mathbf{17.8}$ \\
 & speed-up & $1.1\times$ & $1.2\times$ & $1.2\times$ & $1.3\times$ \\
\midrule
ImageNet-LT & baseline & $11.3$ & $27.1$ & $34.8$ & reached by 2 of 3 seeds \\
 & \textbf{SADL} & $\mathbf{9.9}$ & $\mathbf{23.5}$ & $\mathbf{31.2}$ & $\mathbf{42.9}$ (3 of 3 seeds) \\
 & speed-up & $1.1\times$ & $1.2\times$ & $1.1\times$ & only \textbf{SADL} reaches it on every seed \\
\midrule
CIFAR-10 & baseline & $0.2$ & $1.2$ & $3.7$ & $16.6$ \\
 & \textbf{SADL} & $0.2$ & $\mathbf{1.0}$ & $\mathbf{3.5}$ & $17.0$ \\
 & speed-up & $1.0\times$ & $1.2\times$ & $1.1\times$ & $1.0\times$ \\
\bottomrule
\end{tabular*}
\end{table}

\subsection{Encoder and Endpoint Ranks after Ten Epochs}
\label{sec:reb_long_ranks}

We also checked whether the encoder--endpoint gap persists after
longer training. With joint training and SIGReg, $\mathcal D$-only LeWM reaches
$\operatorname{erank}(Z)=83.8$ at epoch 10 while
$\operatorname{erank}(H)=31.8$; with SADL they are $86.2$ and
$81.3$. For INTACT the values are $79.3$ and $30.0$ without and
$87.7$ and $74.8$ with SADL. These historical joint runs use a
MATH encoder for $\mathcal D$-only and Flash for SADL.
Table~\ref{tab:reb_backend} provides
the matched-backend test.

\relax \section{Computational Cost and Inference Workload}
\label{app:ahead_cost}

\subsection{Training Runtime and Memory}
\label{app:ahead_training_cost}

We measure native and portable implementations of LeWM and INTACT on H20-3e and H200 GPUs. Native implementations use MATH attention for both encoder and predictor. All arms use SIGReg $0.03$; SADL uses weight $0.01$ when enabled. Each reported cell uses three independent 40-update runs, BF16, batch size 256 per task, history 3, eight frames, seed 42, and 10 workers per task. PushT uses one GPU and warmup/cosine scheduling; four-task training uses one GPU per task, Gloo, and a constant learning rate of $5\times10^{-4}$. Both implementations use the original data without a decoded-image dataset cache. Software is PyTorch 2.7.1+cu128, Triton 3.3.1, and Lightning 2.6.4; driver versions are 550.127.08 on H20/H200.

\begin{table}[H]
\centering
\caption{Training time and peak device memory. B: native SIGReg-only; N: native SIGReg+SADL; P: portable SIGReg+SADL. Step time is mean $\pm$ sample SD in ms; peak memory is the mean in GiB/card. Each entry uses three independent launches.}
\label{tab:app_training_cost}
\label{tab:app_training_cost_matched_raw}
\begingroup\small\setlength{\tabcolsep}{2pt}\renewcommand{\arraystretch}{1.12}
\begin{tabular*}{\linewidth}{@{\extracolsep{\fill}}lrrrr@{}}
\toprule
Model / workload & Step B & Step N & Step P & Peak B/N/P \\
\midrule
\multicolumn{5}{l}{\textbf{H20 (H20-3e)}} \\
LeWM / PushT & $1351.01\!\pm\!48.56$ & $1415.10\!\pm\!16.56$ & $350.79\!\pm\!11.95$ & 21.49/21.65/12.10 \\
LeWM / 4 tasks & $1751.89\!\pm\!297.75$ & $1569.79\!\pm\!64.75$ & $417.44\!\pm\!16.90$ & 22.16/22.17/12.35 \\
INTACT / PushT & $1487.99\!\pm\!203.11$ & $1455.16\!\pm\!130.03$ & $355.99\!\pm\!18.88$ & 21.51/21.67/12.73 \\
INTACT / 4 tasks & $1620.28\!\pm\!259.33$ & $1468.83\!\pm\!89.54$ & $625.43\!\pm\!358.90$ & 22.57/22.58/12.62 \\
\midrule
\multicolumn{5}{l}{\textbf{H200}} \\
LeWM / PushT & $931.77\!\pm\!18.03$ & $934.92\!\pm\!42.26$ & $149.01\!\pm\!0.09$ & 21.72/21.88/12.42 \\
LeWM / 4 tasks & $1154.90\!\pm\!68.80$ & $1108.45\!\pm\!114.64$ & $205.58\!\pm\!0.80$ & 22.39/22.40/12.58 \\
INTACT / PushT & $929.10\!\pm\!23.07$ & $933.92\!\pm\!4.36$ & $198.10\!\pm\!43.64$ & 21.74/21.90/12.42 \\
INTACT / 4 tasks & $1211.28\!\pm\!45.01$ & $1048.06\!\pm\!50.77$ & $207.34\!\pm\!5.43$ & 22.80/22.80/12.85 \\
\bottomrule
\end{tabular*}
\endgroup
\end{table}

Step time is $(t_{30}-t_{10})/20$, using the slowest rank for four tasks. Device memory is sampled every 500 ms; we take each run's maximum over participating GPUs, then average across launches. Comparisons use the same GPU model and host; concurrent blocks have separate GPUs but share CPU/storage resources. One H20 single-task INTACT repetition changed GPU index between arms. Incomplete or GPU-contaminated H20/H200 runs are excluded.

B/N comparisons keep the attention backend fixed. N/P comparisons hold the objective fixed and include attention-backend and execution changes; the portable path omits native diagnostic probes. B/P comparisons measure the overall implementation difference. Startup, compilation before step 10, and checkpoint writing outside the interval are excluded. These short runs do not measure full-epoch time or establish equal final task performance.

\subsection{Search Workload of the Control Interfaces}
\label{app:ahead_inference_cost}

INTACT's Direct interface performs no candidate-action search.
The \relax \relax \relax official \relax records for the alternative endpoint
experiments contain zero candidate-action evaluations
and zero planning-cost calls.
LeWM Pure-CEM uses 300 candidates per iteration and
30 iterations, corresponding to 9,000 candidate evaluations
per planning invocation.

SADL adds a training-only objective and retains these
existing control interfaces.
It introduces no additional inference network or
candidate-search procedure.

\end{document}